\documentclass[11pt]{article}

\usepackage[]{acl}

\usepackage{times}
\usepackage{latexsym}

\usepackage[T1]{fontenc}

\usepackage[utf8]{inputenc}

\usepackage{microtype}

\usepackage{inconsolata}

\usepackage{graphicx}

\usepackage{microtype}
\usepackage{times}
\usepackage{amsmath}
\usepackage{amssymb}
\usepackage{textcomp}       
\usepackage{latexsym}
\usepackage{booktabs}
\usepackage{multirow}
\usepackage{xcolor}
\usepackage{colortbl}
\usepackage{array}
\usepackage{subcaption}
\usepackage{inconsolata}
\usepackage{setspace}
\usepackage{longtable}

\definecolor{rankone}{RGB}{180,230,180}
\definecolor{ranktwo}{RGB}{180,210,240}
\definecolor{rankthree}{RGB}{255,220,180}

\title{%
  \raisebox{-0.9ex}{\protect\includegraphics[height=2.0\fontcharht\font`\B]{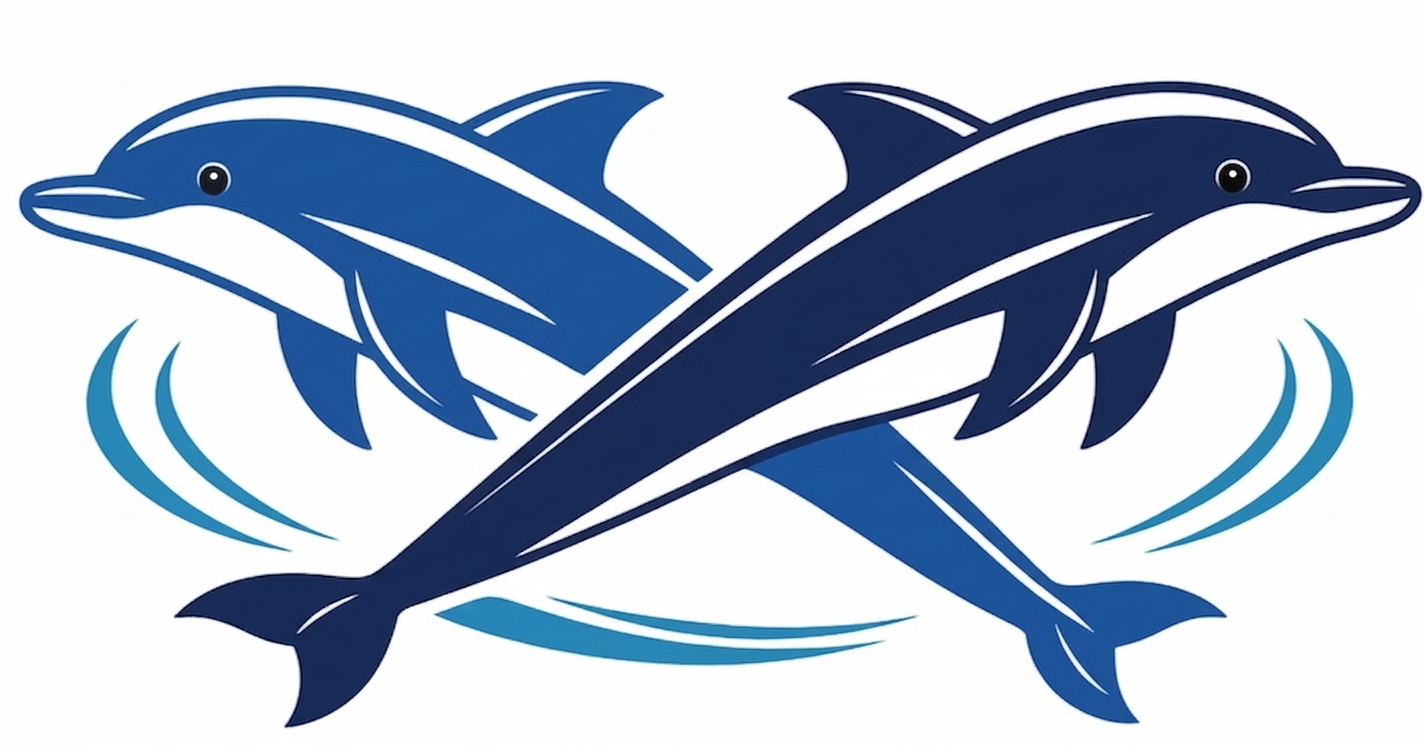}}%
  CrossHallu: Do Hallucination Signals Generalize Across Languages
  and Domains in Large Language Model's Internals?%
}

\author{Aisha Alansari\textsuperscript{1} \and Malak Alkhorasani\textsuperscript{2} \and Hamzah Luqman\textsuperscript{1} \\
         King Fahd University of Petroleum and Minerals \\ Imam Abdulrahman bin Faisal University}

\begin{document}
\maketitle
\begin{abstract}
Recent hallucination detection techniques in large language models (LLMs) focus on directly extracting features from a model's internal representations and training a classifier on these features to detect hallucinations, demonstrating promising results. Notwithstanding this advancement, most internal-state hallucination detection techniques have been explored predominantly in English, raising the question of whether such internal signals generalize across different languages and domains. To address this gap, we present \textit{CrossHallu}, the first study to evaluate the cross-lingual and cross-domain generalization of hallucination detection using internal representations from six LLMs on the generative question-answering task. We conduct a systematic Arabic $\leftrightarrow$ English evaluation using TruthfulQA, its Arabic-translated version, and HalluScore. This evaluation encompasses monolingual training and testing, cross-lingual transfer, cross-domain transfer, and combined cross-lingual and cross-domain transfer. The results reveal that internal-state hallucination signals in LLMs transfer across languages and domains for most models, with cross-lingual performance highly dependent on both class separability and language alignment in the feature space, whereas cross-domain transfer within Arabic varies depending on the training and testing datasets used for the hallucination detector. The code is publicly available at \url{https://github.com/aishaalansari57/CrossHal}.
\end{abstract}

\section{Introduction}

Large language models (LLMs) have demonstrated remarkable success across a wide range of natural language generation tasks, achieving state-of-the-art performance in benchmarks spanning question answering (QA), summarization, and machine translation \cite{li2024leveraging}. Despite these advances, LLMs continue to be susceptible to hallucination, which occurs when they confidently generate outputs that are factually incorrect or unsupported \cite{maynez2020faithfulness,ji2023survey,huang2025survey}. This issue is particularly critical in high-stakes domains such as finance, medicine, and law, where hallucinated information can lead to harmful decisions, misinformation, or loss of trust \cite{alansari2025large}. 

\begin{figure}[t!]
    \centering
    \includegraphics[width=\columnwidth]{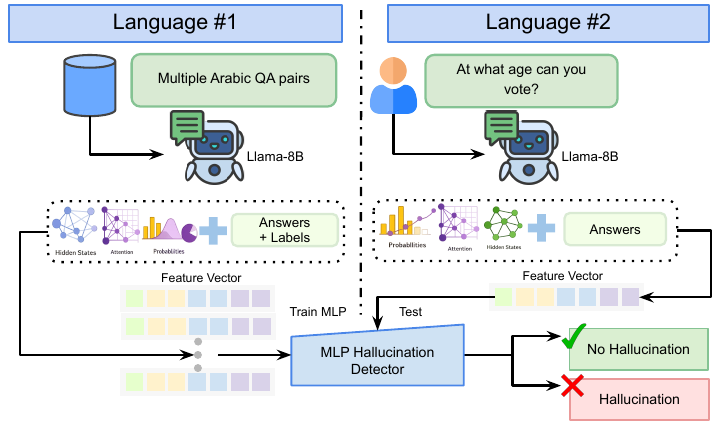}
    \caption{Illustration of cross-lingual transfer of hallucination signals derived from LLM internal representations. A detector trained in a source language is evaluated on a target language.}
    \label{fig:abstract}
\end{figure}
An emerging direction in hallucination detection for LLMs focuses on directly extracting features from an LLM's internal representations, using activations, attention patterns, or hidden states as probes of truthfulness \cite{liu2025attention,ogasa2025hallucinated}. These detection methods are motivated by the hypothesis that hallucinations may correspond to detectable distributional shifts or uncertainty patterns within the model's internal states \cite{chuang2024lookback}. Recent studies have demonstrated that training lightweight detectors that predict hallucinations directly from internal activation often outperforms surface‑level confidence heuristics based on probabilities or entropy \cite{chuang2024lookback,shelmanov2025head,dasgupta2025hallushift,samaga2026halluzig}.

Despite this progress, most internal-state hallucination detection methods have been investigated primarily in English, leaving open the question of \textit{whether such internal signals generalize across languages and domains.} This is a critical gap, as LLMs are increasingly deployed in multilingual settings, and hallucination rates vary substantially across languages, often being higher for lower-resource languages \cite{islam2025much,alansari2025arahallueval}. A key hypothesis motivating cross-lingual internal state analysis is that multilingual models, trained jointly across many languages, may develop language-agnostic representations at deeper layers of the network. This phenomenon is supported by findings in cross-lingual transfer learning, which show that multilingual models tend to learn shared semantic representations across languages, particularly in deeper layers, enabling knowledge transfer even between linguistically distant languages. \cite{riemenschneider2025cross,tashu2025cross}. If hallucination-related signals exist within these shared representational subspaces, it may be feasible to train a detector in one language and then use it in another. It can also be used to enhance detection in low-resource languages by aligning their internal states with those of higher-resource languages.

In this work, we present \textit{CrossHallu}, which, to the best of our knowledge, is the first study to systematically evaluate the cross-lingual and cross-domain generalization of hallucination detection using internal representations from six LLMs on generative QA. We conduct a systematic Arabic $\leftrightarrow$ English evaluation using three datasets: TruthfulQA \cite{lin2022truthfulqa}, an Arabic translated version of TruthfulQA, and HalluScore \cite{alansari2026halluscorelargelanguagemodel}. The cross-evaluation covers monolingual training and testing, cross-lingual transfer (English $\leftrightarrow$ Arabic), cross-domain transfer across Arabic datasets, and combined cross-lingual and cross-domain transfer. Figure \ref{fig:abstract} illustrates cross-lingual transfer, in which a multi-layer perceptron (MLP) is trained on hidden states from an LLM in one language and evaluated on another language. The main contributions of this study are summarized as follows:


\begin{itemize}
    \item To the best of our knowledge, we present the first study evaluating the cross-lingual and cross-domain transferability of hallucination signals derived from six LLMs' internal representations.
    
    \item We propose an evaluation framework that examines how hallucination signals extracted from LLM internal representations generalize across different languages and datasets across cross-lingual, cross-domain, and combined cross-lingual cross-domain transfer settings.

    \item We provide a detailed analysis of hallucination-related representation shifts across transformer layers and languages.
    
    
    \item We provide an empirical analysis of which LLMs exhibit transferable hallucination signals across languages and domains, offering insights into the language-agnostic nature of internal hallucination representations.
\end{itemize}

\section{Related Work}

Recent studies suggest that internal representations provide reliable and generalizable signals for hallucination detection \cite{du2024haloscope,dasgupta2025hallushift,binkowski2025hallucination, samaga2026halluzig, bazarova2025hallucination}. In parallel, prior work has shown that multilingual LLMs learn partially shared internal representations that support cross-lingual transfer \cite{wang2024probing, huang2025neurons}. However, despite these capabilities, existing hallucination detection approaches largely focus on single-language settings, leaving the transferability of internal hallucination signals across languages underexplored.

This gap raises an important question: \textit{do hallucination signals derived from internal representations remain reliable when transferred across languages and domains?} To answer this question, we introduce \textit{CrossHallu}, a framework that systematically evaluates the consistency and transferability of such signals across monolingual, cross-lingual (English $\leftrightarrow$ Arabic), cross-domain (across Arabic datasets), and combined cross-lingual cross-domain settings.  A detailed review of hallucination detection, internal representation analysis for hallucination detection, and cross-lingual representation alignment is provided in Appendix~\ref{related work}.
\section{CrossHallu Methodology}
We extend the work of \textit{Hallushift} \cite{dasgupta2025hallushift} to evaluate the cross-lingual and cross-domain generalization of hallucination detection from six LLM internal representations for generative QA.
The core intuition behind \textit{HalluShift} is that when a model generates a hallucinated response, the dynamics of information flow through its transformer layers exhibit measurably different patterns than in grounded, factual generation. These differences can be observed through hidden state transitions, attention patterns, and token probability distributions. Internal signals can be passively extracted during a single inference pass without incurring additional computational cost. Following \textit{HalluShift}, we aim to evaluate whether these shifts remain consistent across languages and domains. Our goal is therefore to empirically evaluate the transferability of internal hallucination signals rather than assume their invariance.

\begin{figure*}
    \centering
    \includegraphics[width=\linewidth]{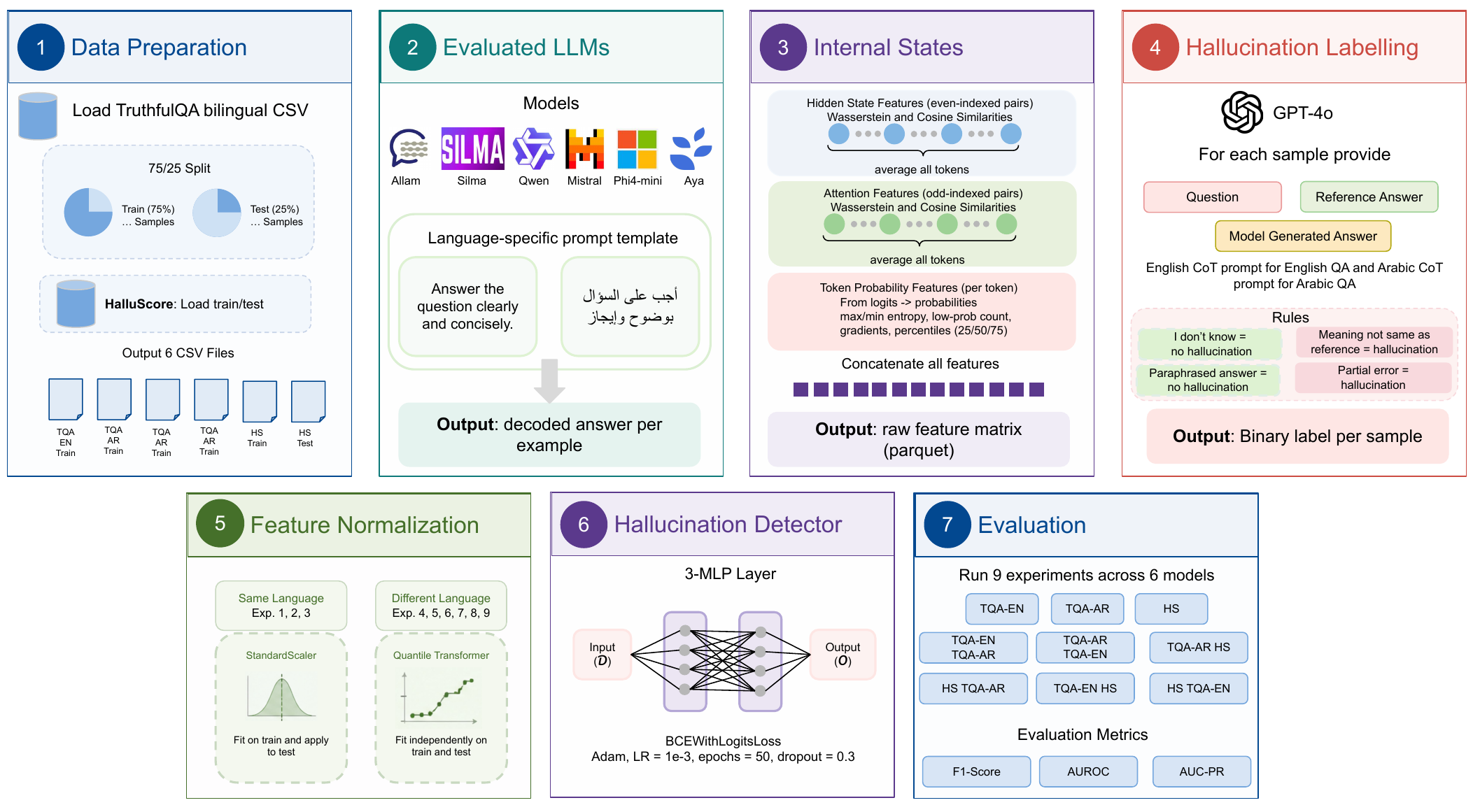}
    \caption{Overview of the proposed framework for evaluating the transferability of hallucination signals from LLM internal representations across languages and domains.}
    \label{fig:pipeline}
\end{figure*}

As illustrated in Figure \ref{fig:pipeline}, we evaluate six LLMs across three datasets: TruthfulQA \cite{lin2022truthfulqa}, an Arabic translation of TruthfulQA, and HalluScore \cite{alansari2026halluscorelargelanguagemodel}. We perform several experiments designed to systematically isolate the effects of language shift, domain shift, and their combination. Three monolingual baseline experiments are conducted, in which training and testing occur within the same language and dataset (TruthfulQA EN, TruthfulQA AR, and HalluScore). In the other experiments, a hallucination detector is trained on internal representations from a source setting and evaluated on a target setting to measure transferability. For each transfer scenario, we consider both directions (source$\rightarrow$target) and (target$\rightarrow$source). Therefore, two cross-lingual experiments are conducted in which the domain is kept constant while the language changes (TruthfulQA EN$\leftrightarrow$TruthfulQA AR). Two cross-domain experiments are conducted, with the language held constant while the dataset changes (TruthfulQA AR$\leftrightarrow$HalluScore). Finally, two cross-lingual, cross-domain experiments are conducted in which both language and domain change simultaneously, representing the most challenging transfer setting (TruthfulQA EN$\leftrightarrow$HalluScore).

\subsection{Internal State Feature Extraction}
We extract three groups of features from an LLM's internal representations during a single inference pass. Let $L$ denote the total number of transformer layers in an LLM, and let $T$ denote the number of tokens generated in response to a given question. For each generated token $t \in \{1, \ldots, T\}$, we extract features from the hidden states, attention tensors, and output logits simultaneously.

\noindent \vspace{0mm}
\textbf{Hidden State Features}
Let $\mathbf{h}^{(l)}_t \in \mathbb{R}^{d}$ denote the hidden state at 
layer $l$ for token $t$, where $d$ is the LLM's hidden dimension. We 
select every second layer starting from layer 2, yielding the index set 
$\mathcal{L} = \{2, 4, 6, \ldots, L\}$, producing $|\mathcal{L}| = L/2$ 
selected layers. This layer selection balances highly correlated adjacent layers with larger windows that smooth over fine-grained cross-layer shifts. This choice is also empirically supported by HalluShift's ablation. For each pair of consecutive selected layers 
$(l_i, l_{i+1})$, we first normalize each hidden state as a probability 
distribution using softmax. We then compute two similarity measures between consecutive layer 
representations, which are the Wasserstein distance and cosine similarity. The Wasserstein distance measures the distributional divergence between hidden states at adjacent layers, whereas the cosine similarity measures the geometric alignment between consecutive layer representations. Both measures are computed for all $|\mathcal{L}| - 1$ consecutive layer pairs and averaged across all $T$ generated tokens.

\noindent \vspace{0mm}
\textbf{Attention Features}
Let $\mathbf{a}^{(l)}_t \in \mathbb{R}^{H \times S}$ denote the attention 
tensor in layer $l$ for token $t$, where $H$ is the number of attention 
heads and $S$ is the sequence length. We select every second layer starting 
from layer 1, producing the index set 
$\mathcal{A} = \{1, 3, 5, \ldots, L-1\}$. The same normalization, 
Wasserstein distance and cosine similarity computed for the hidden state features
are applied to the attention tensors. These tensors are averaged across tokens in the same manner as the hidden state features, producing an additional feature vector of length $L - 2$.

\noindent \vspace{0mm}
\textbf{Token Probability Features}
For each generated token $t$, we apply Softmax to the output logits 
$\mathbf{z}_t \in \mathbb{R}^{V}$, where $V$ is the vocabulary size, to 
obtain the token probability distribution. We record the maximum and minimum vocabulary probabilities at each step $\mathbf{p}^{\max} = \{p_1^{\max}, \ldots, p_T^{\max}\}$ and 
$\mathbf{p}^{\min} = \{p_1^{\min}, \ldots, p_T^{\min}\}$. These values are essential as they reflect the LLM's generation confidence 
at each step. Specifically, $p_t^{\max}$ captures the LLM's peak confidence in its most likely next token. A high $p_t^{\max}$ indicates that the model is certain about what to generate next, while a consistently low $p_t^{\max}$ across the sequence suggests that the model is uncertain or struggling to commit to a token, which may be an indication of  
hallucination. Conversely, $p_t^{\min}$ captures the lowest probability 
assigned to any token in the vocabulary at step $t$. When 
the distribution is highly concentrated on a small number of tokens, 
$p_t^{\min}$ will be extremely small, whereas when the distribution is more uniform, it indicates greater uncertainty across the vocabulary and $p_t^{\min}$ will 
be comparatively larger. 

From these two sequences: $\mathbf{p}^{\max} = \{p_1^{\max}, \ldots, p_T^{\max}\}$ and 
$\mathbf{p}^{\min} = \{p_1^{\min}, \ldots, p_T^{\min}\}$, we derive eight scalar features. The \textit{minimum of maximum probabilities} captures the most uncertain generation step across the entire response:

\begin{equation}
\label{eq1}
    f_{\min\text{-max}} = \min_t\ p_t^{\max}
\end{equation}

The \textit{maximum probability spread} ($f_{\text{Mps}}$) measures the largest per-step gap between the most and least likely tokens, reflecting the sharpest concentration of probability mass observed during generation:

\begin{equation}
\label{eq2}
    f_{\text{Mps}} = \max_t \left(p_t^{\max} - p_t^{\min}\right)
\end{equation}

The \textit{normalized entropy} of $ p^{\max}$, 
bounded in $[0, 1]$, treats the sequence of per-step peak probabilities as a distribution and measures how uniformly confidence is spread across generation steps:

\begin{equation}
\label{eq3}
    f_{\text{ent}} = \frac{
        -\sum_{t} p_t^{\max} \log p_t^{\max}
    }{
        \log T
    }
\end{equation}

The \textit{count of low-probability tokens},counts how often the model's top token probability falls below a threshold $\tau$ = 0.1, indicating steps where the model fails to commit to any token with reasonable confidence:

\begin{equation}
\label{eq4}
    f_{\text{low}} = \sum_{t=1}^{T} 
        \mathbf{1}\left[p_t^{\max} < \tau\right], 
    \qquad \tau = 0.1
\end{equation}

The \textit{mean probability gradient},  measures the average absolute change in  $\mathbf{p}^{\max}$ between consecutive generation steps, capturing how erratically the model's certainty varies across the generated sequence

\begin{equation}
\label{eq5}
    f_{\text{grad}} = \frac{1}{T-1} \sum_{t=1}^{T-1} 
        \left| p_{t+1}^{\max} - p_t^{\max} \right|
\end{equation}

The \textit{percentiles} $f_{p25}$, $f_{p50}$, and $f_{p75}$ of the maximum 
probability sequence characterize the overall confidence distribution. All 
probability features are computed separately for both $\mathbf{p}^{\max}$ 
and $\mathbf{p}^{\min}$ sequences. The complete feature vector for each example is formed by concatenating all hidden states, attention, and token probabilities into a fixed-length representation that is passed to the classifier.

\subsection{Hallucination Labelling}

Unlike \textit{HalluShift} \cite{dasgupta2025hallushift}, which relies 
on BLEURT \cite{sellam2020bleurt} to assess the semantic similarity between the generated and reference answers for hallucination labelling, we use GPT-4o-as-a-Judge, which evaluates factual correctness rather than surface-level wording. This choice is motivated by two limitations of BLEURT in our setting. First, BLEURT is trained 
predominantly on English data and has been shown to perform poorly on Arabic 
text, making it unsuitable for evaluating Arabic-language responses. Second, 
BLEURT operates as a surface-level similarity metric and does not account 
for the nuances of factual equivalence, such as valid paraphrases, partial 
answers, or the common misconception traps that are characteristic of 
\textit{TruthfulQA} \cite{lin2022truthfulqa} and \textit{HalluScore} \cite{alansari2026halluscorelargelanguagemodel}. The details of prompting GPT-4o and evaluating its labeling are presented in Appendix \ref{gpt-4o}.

\subsection{Hallucination Detector}
We train a fully connected three-layer MLP to detect hallucinations. We intentionally adopt a single lightweight probing classifier and keep its architecture and hyperparameters fixed across all experiments. This controlled design ensures that the differences in performance are primarily attributable to the underlying internal representations and the transfer setting, rather than changes in the detector architecture. The details of training the MLP network are discussed in Appendix \ref{implementation}.

\section{Experiments}

\subsection{Datasets}
We evaluate the cross-lingual and cross-domain generalization of LLMs internal representations to detect hallucination using three datasets: TruthfulQA \cite{lin2022truthfulqa}, the Arabic translated version of TruthfulQA, and HalluScore \cite{alansari2026halluscorelargelanguagemodel}. For English, we used the TruthfulQA dataset, which is the most widely used benchmark for evaluating hallucinations in generative QA. To enable cross-lingual evaluation, we translated TruthfulQA into Arabic. After manual verification, we retained 772 of the original 817 questions, excluding samples that did not translate clearly or that lost their intended meaning in Arabic. More details about translating TruthfulQA are presented in Appendix \ref{datasets}. These 772 aligned QA pairs were used for both English and Arabic to evaluate cross-lingual transfer while keeping the domain constant. We split the dataset into 75\% for training and 25\% for testing, including 597 samples in the training set and 193 in the testing set. For cross-lingual and cross-domain evaluation, we additionally used HalluScore, which, to the best of our knowledge, is currently the only available Arabic dataset specifically designed for hallucination detection in Arabic generative QA. The training set for HalluScore consists of 621 samples, whereas the testing set comprises 206 samples.

\subsection{Evaluated Models}
We evaluate six Arabic-capable LLMs with varying sizes, architectures, and levels of Arabic support, selected for their Arabic–English capabilities and computational feasibility. \textit{ALLaM-7B} \cite{bari2024allam} is an Arabic–English bilingual model developed by SDAIA, while \textit{SILMA-9B} \cite{silma-9b-2024} is an Arabic-optimized model based on Gemma-2-9B. We further evaluate the compact multilingual \textit{Phi-4-mini} \cite{phi4mini2025}, \textit{Qwen2.5-14B} \cite{qwen25_14b2024}, pretrained on over 18 trillion tokens, and the efficient multilingual \textit{Ministral-8B} \cite{ministral8b2024}. Finally, \textit{Aya-23-8B} \cite{aryabumi2024aya} is a multilingual model trained on instruction data spanning 101 languages, including Arabic.

\subsection{Evaluation Metrics}
Consistent with prior work \cite{alansari2025large}, we evaluate cross-lingual and cross-domain generalization using F1-score, AUROC, and AUC-PR. F1-score and AUC-PR account for class imbalance, with AUC-PR emphasizing precision and recall for hallucinated outputs, while AUROC measures discrimination between hallucinated and grounded outputs across thresholds.

\subsection{Implementation Details}

\textbf{Response generation.} We used the HuggingFace Transformers library to load all six evaluated models. We used the English prompt to generate English responses \textit{"Answer the question concisely. Q: \{question\}, A:} and the Arabic prompt to generate Arabic responses \raisebox{-0.3\height}{\includegraphics[height=1.3em]{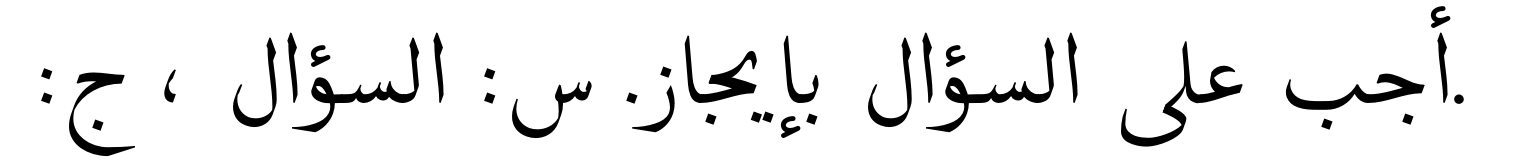}} More details are presented in Appendix \ref{implementation}.  

\noindent
\textbf{Hallucination labelling.} We used GPT-4o with temperature zero and \texttt{seed=42} for deterministic and reproducible labelling. Each call is allocated a maximum of 80 output tokens to accommodate the 
chain-of-thought reasoning sentence, followed by the binary label. The prompt used for hallucination labelling is illustrated in Figure \ref{fig:gpt-4o}. More details about hallucination labeling are discussed in Appendix \ref{gpt-4o}.

\noindent
\textbf{Feature normalization.} For monolingual experiments, a StandardScaler is fit on the training features and applied to the test 
features. For cross-lingual experiments, a QuantileTransformer is fit 
independently on the training and test feature sets. This normalization is motivated by the observed scale mismatch between English and Arabic internal state features. Arabic token probabilities are 
systematically lower and more spread than English due to the morphological complexity of Arabic, causing a standard scaler fit on English features to 
produce distorted representations when applied to Arabic. By mapping each language's feature distribution independently to a standard normal distribution, the classifier receives comparable representations regardless of the source language.

\section{Results}
\begin{figure*}[ht!]
    \centering
    \includegraphics[width=\textwidth]{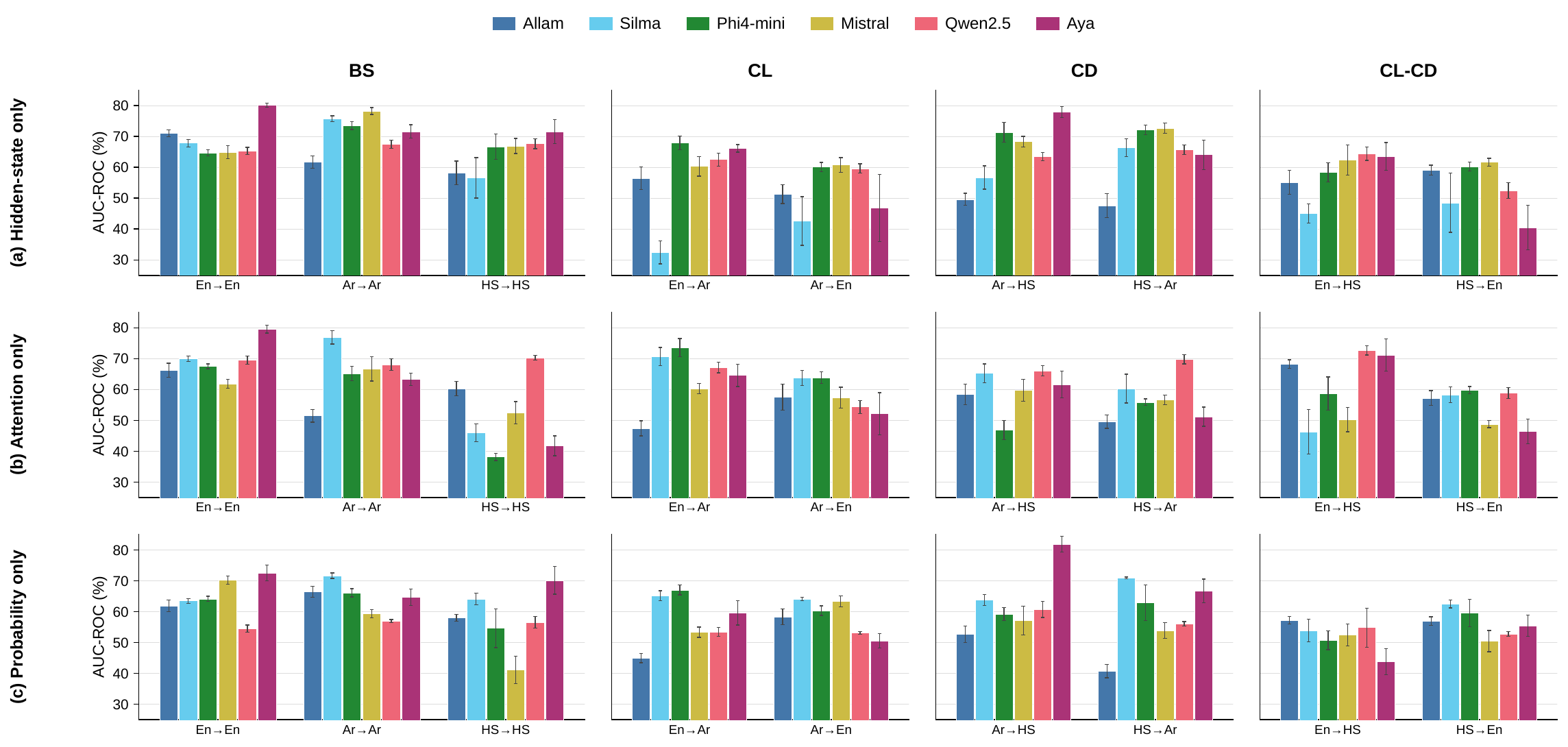}
    \caption{Comparison of training the MLP on hidden-state, attention, and probability features across nine transfer settings and six LLMs. Each subplot reports the AUC-ROC achieved using a single feature group.}
    \label{fig:transfer_results}
\end{figure*}

Table \ref{tab:results} presents the \textit{CrossHallu} evaluation results, while Figure \ref{fig:transfer_results} compares feature effects across the nine settings. We evaluate six LLMs (Allam, Silma, Phi-4-mini, Ministral, Qwen2.5, and Aya) under four experimental regimes. The baseline trains and tests the hallucination detector on the same dataset. Cross-lingual evaluation transfers between TruthfulQA EN and AR while keeping the domain fixed, whereas cross-domain evaluation transfers between the Arabic TruthfulQA and HalluScore datasets. Finally, the combined cross-lingual-cross-domain setting transfers between TruthfulQA EN and HalluScore, changing both language and domain. The high AUC-PR scores are largely influenced by the hallucination-majority class imbalance; therefore, we use AUROC as our primary metric. Additional ablations, insights, and visualizations are provided in Appendices \ref{ablation} and \ref{insights}.

\definecolor{rankone}{RGB}{180,230,180}
\definecolor{ranktwo}{RGB}{180,210,240}
\definecolor{rankthree}{RGB}{255,220,180}

\begin{table*}[t!]
\centering
\caption{Hallucination detection performance (F1-Score, AUC-ROC, and PR-AUC in \%, mean over 6 seeds with $\pm$std shown below each value) across models and transfer settings. En = TruthfulQA EN, Ar = TruthfulQA AR, HS.\ = HalluScore. Groups: (1)~Same-language baseline (BS), (2)~Cross-lingual (CL), (3)~Cross-domain (CD), (4)~Cross-lingual + cross-domain (CL-CD). \colorbox{rankone}{\phantom{X}} 1st, \colorbox{ranktwo}{\phantom{X}} 2nd, \colorbox{rankthree}{\phantom{X}} 3rd best AUC-ROC per row. Within each model column, \textbf{bold} and \underline{underline} mark the best and second-best score for that metric across the transfer settings (column-wise).}
\label{tab:results}
\resizebox{\linewidth}{!}{%
\begin{tabular}{@{}p{1.1cm} l ccc ccc ccc ccc ccc ccc}
\toprule
\multirow{2}{*}{\textbf{Exp.}} & \multirow{2}{*}{\textbf{Train $\to$ Test}} &
\multicolumn{3}{c}{\textbf{Allam}} &
\multicolumn{3}{c}{\textbf{Silma}} &
\multicolumn{3}{c}{\textbf{Phi4-mini}} &
\multicolumn{3}{c}{\textbf{Ministral}} &
\multicolumn{3}{c}{\textbf{Qwen2.5}} &
\multicolumn{3}{c}{\textbf{Aya}} \\
\cmidrule(lr){3-5} \cmidrule(lr){6-8} \cmidrule(lr){9-11} \cmidrule(lr){12-14} \cmidrule(lr){15-17} \cmidrule(lr){18-20}
& &
\textbf{F1} & \textbf{AUC} & \textbf{PR} &
\textbf{F1} & \textbf{AUC} & \textbf{PR} &
\textbf{F1} & \textbf{AUC} & \textbf{PR} &
\textbf{F1} & \textbf{AUC} & \textbf{PR} &
\textbf{F1} & \textbf{AUC} & \textbf{PR} &
\textbf{F1} & \textbf{AUC} & \textbf{PR} \\
\midrule
\multirow{3}{*}{BS}
& En $\to$ En
  & \shortstack{85.12\\ {\tiny$\pm$0.47}}
  & \shortstack{\textbf{69.86}\\ {\tiny$\pm$2.31}}
  & \shortstack{81.57\\ {\tiny$\pm$1.63}}
  & \shortstack{76.79\\ {\tiny$\pm$0.85}}
  & \cellcolor{rankthree}\shortstack{69.99\\ {\tiny$\pm$0.64}}
  & \shortstack{75.34\\ {\tiny$\pm$1.01}}
  & \shortstack{73.56\\ {\tiny$\pm$1.60}}
  & \shortstack{67.29\\ {\tiny$\pm$1.21}}
  & \shortstack{79.57\\ {\tiny$\pm$1.51}}
  & \shortstack{78.84\\ {\tiny$\pm$2.45}}
  & \cellcolor{ranktwo}\shortstack{71.50\\ {\tiny$\pm$0.72}}
  & \shortstack{84.64\\ {\tiny$\pm$0.64}}
  & \shortstack{57.48\\ {\tiny$\pm$1.34}}
  & \shortstack{69.81\\ {\tiny$\pm$1.05}}
  & \shortstack{67.17\\ {\tiny$\pm$1.24}}
  & \shortstack{86.77\\ {\tiny$\pm$0.87}}
  & \cellcolor{rankone}\shortstack{\underline{82.86}\\ {\tiny$\pm$0.71}}
  & \shortstack{88.64\\ {\tiny$\pm$0.96}} \\
\addlinespace[2pt]

& Ar $\to$ Ar
  & \shortstack{84.12\\ {\tiny$\pm$0.61}}
  & \shortstack{\underline{64.67}\\ {\tiny$\pm$2.45}}
  & \shortstack{83.17\\ {\tiny$\pm$1.30}}
  & \shortstack{82.11\\ {\tiny$\pm$0.80}}
  & \cellcolor{ranktwo}\shortstack{\textbf{78.93}\\ {\tiny$\pm$1.35}}
  & \shortstack{85.82\\ {\tiny$\pm$1.66}}
  & \shortstack{86.92\\ {\tiny$\pm$0.16}}
  & \cellcolor{rankthree}\shortstack{\underline{77.58}\\ {\tiny$\pm$1.07}}
  & \shortstack{91.64\\ {\tiny$\pm$0.44}}
  & \shortstack{80.04\\ {\tiny$\pm$1.38}}
  & \cellcolor{rankone}\shortstack{\textbf{79.38}\\ {\tiny$\pm$1.36}}
  & \shortstack{89.10\\ {\tiny$\pm$1.13}}
  & \shortstack{34.66\\ {\tiny$\pm$2.76}}
  & \shortstack{\underline{70.62}\\ {\tiny$\pm$1.69}}
  & \shortstack{53.55\\ {\tiny$\pm$1.66}}
  & \shortstack{87.74\\ {\tiny$\pm$0.32}}
  & \shortstack{75.97\\ {\tiny$\pm$0.89}}
  & \shortstack{90.30\\ {\tiny$\pm$0.38}} \\
\addlinespace[2pt]

& HS $\to$ HS
  & \shortstack{\underline{88.18}\\ {\tiny$\pm$0.88}}
  & \shortstack{63.57\\ {\tiny$\pm$2.62}}
  & \shortstack{\textbf{87.19}\\ {\tiny$\pm$0.87}}
  & \shortstack{\textbf{95.96}\\ {\tiny$\pm$0.00}}
  & \shortstack{63.23\\ {\tiny$\pm$3.91}}
  & \shortstack{\underline{95.35}\\ {\tiny$\pm$0.56}}
  & \shortstack{\textbf{97.77}\\ {\tiny$\pm$0.00}}
  & \shortstack{57.00\\ {\tiny$\pm$1.75}}
  & \shortstack{97.02\\ {\tiny$\pm$0.41}}
  & \shortstack{\textbf{95.80}\\ {\tiny$\pm$0.38}}
  & \cellcolor{rankthree}\shortstack{69.09\\ {\tiny$\pm$2.23}}
  & \shortstack{\underline{95.67}\\ {\tiny$\pm$0.32}}
  & \shortstack{\textbf{84.23}\\ {\tiny$\pm$1.90}}
  & \cellcolor{ranktwo}\shortstack{\textbf{76.19}\\ {\tiny$\pm$2.21}}
  & \shortstack{\textbf{89.34}\\ {\tiny$\pm$1.25}}
  & \shortstack{\textbf{97.00}\\ {\tiny$\pm$0.00}}
  & \cellcolor{rankone}\shortstack{79.81\\ {\tiny$\pm$3.69}}
  & \shortstack{97.88\\ {\tiny$\pm$0.50}} \\
\addlinespace[2pt]
\midrule
\multirow{2}{*}{CL}
& En $\to$ Ar
  & \shortstack{81.02\\ {\tiny$\pm$1.76}}
  & \shortstack{50.69\\ {\tiny$\pm$2.55}}
  & \shortstack{73.27\\ {\tiny$\pm$1.68}}
  & \shortstack{76.94\\ {\tiny$\pm$1.32}}
  & \shortstack{43.28\\ {\tiny$\pm$4.99}}
  & \shortstack{61.35\\ {\tiny$\pm$2.89}}
  & \shortstack{84.22\\ {\tiny$\pm$1.29}}
  & \cellcolor{rankone}\shortstack{\textbf{78.03}\\ {\tiny$\pm$0.98}}
  & \shortstack{92.75\\ {\tiny$\pm$0.44}}
  & \shortstack{75.80\\ {\tiny$\pm$2.17}}
  & \shortstack{63.42\\ {\tiny$\pm$3.35}}
  & \shortstack{76.46\\ {\tiny$\pm$3.39}}
  & \shortstack{51.66\\ {\tiny$\pm$2.86}}
  & \cellcolor{rankthree}\shortstack{66.74\\ {\tiny$\pm$2.39}}
  & \shortstack{53.41\\ {\tiny$\pm$3.40}}
  & \shortstack{86.46\\ {\tiny$\pm$0.47}}
  & \cellcolor{ranktwo}\shortstack{73.36\\ {\tiny$\pm$1.38}}
  & \shortstack{89.72\\ {\tiny$\pm$0.57}} \\
\addlinespace[2pt]

& Ar $\to$ En
  & \shortstack{83.41\\ {\tiny$\pm$0.63}}
  & \shortstack{57.34\\ {\tiny$\pm$1.72}}
  & \shortstack{75.51\\ {\tiny$\pm$1.53}}
  & \shortstack{76.13\\ {\tiny$\pm$0.87}}
  & \shortstack{43.92\\ {\tiny$\pm$4.53}}
  & \shortstack{57.84\\ {\tiny$\pm$3.42}}
  & \shortstack{79.00\\ {\tiny$\pm$0.00}}
  & \cellcolor{ranktwo}\shortstack{66.04\\ {\tiny$\pm$1.74}}
  & \shortstack{78.84\\ {\tiny$\pm$1.89}}
  & \shortstack{74.04\\ {\tiny$\pm$2.50}}
  & \cellcolor{rankone}\shortstack{66.37\\ {\tiny$\pm$0.83}}
  & \shortstack{81.27\\ {\tiny$\pm$0.55}}
  & \shortstack{13.94\\ {\tiny$\pm$8.13}}
  & \cellcolor{rankthree}\shortstack{59.71\\ {\tiny$\pm$1.65}}
  & \shortstack{55.23\\ {\tiny$\pm$0.63}}
  & \shortstack{79.08\\ {\tiny$\pm$0.36}}
  & \shortstack{57.26\\ {\tiny$\pm$3.06}}
  & \shortstack{75.35\\ {\tiny$\pm$1.79}} \\
\addlinespace[2pt]
\midrule
\multirow{2}{*}{CD}
& Ar $\to$ HS
  & \shortstack{\textbf{88.27}\\ {\tiny$\pm$1.65}}
  & \shortstack{54.52\\ {\tiny$\pm$1.63}}
  & \shortstack{83.96\\ {\tiny$\pm$0.76}}
  & \shortstack{89.76\\ {\tiny$\pm$1.62}}
  & \shortstack{65.68\\ {\tiny$\pm$1.13}}
  & \shortstack{\textbf{95.51}\\ {\tiny$\pm$0.22}}
  & \shortstack{\underline{97.72}\\ {\tiny$\pm$0.10}}
  & \cellcolor{ranktwo}\shortstack{71.77\\ {\tiny$\pm$2.68}}
  & \shortstack{\textbf{98.39}\\ {\tiny$\pm$0.19}}
  & \shortstack{\underline{85.22}\\ {\tiny$\pm$1.78}}
  & \shortstack{70.46\\ {\tiny$\pm$1.86}}
  & \shortstack{95.31\\ {\tiny$\pm$0.44}}
  & \shortstack{34.26\\ {\tiny$\pm$5.83}}
  & \cellcolor{rankthree}\shortstack{70.59\\ {\tiny$\pm$1.41}}
  & \shortstack{\underline{85.99}\\ {\tiny$\pm$0.70}}
  & \shortstack{\underline{96.77}\\ {\tiny$\pm$0.88}}
  & \cellcolor{rankone}\shortstack{\textbf{86.49}\\ {\tiny$\pm$1.69}}
  & \shortstack{\textbf{98.64}\\ {\tiny$\pm$0.14}} \\
\addlinespace[2pt]

& HS $\to$ Ar
  & \shortstack{81.36\\ {\tiny$\pm$1.38}}
  & \shortstack{44.44\\ {\tiny$\pm$1.76}}
  & \shortstack{71.45\\ {\tiny$\pm$0.86}}
  & \shortstack{78.62\\ {\tiny$\pm$0.00}}
  & \cellcolor{rankone}\shortstack{\underline{74.39}\\ {\tiny$\pm$1.69}}
  & \shortstack{83.00\\ {\tiny$\pm$1.85}}
  & \shortstack{86.80\\ {\tiny$\pm$0.00}}
  & \shortstack{70.13\\ {\tiny$\pm$1.62}}
  & \shortstack{88.45\\ {\tiny$\pm$1.36}}
  & \shortstack{80.62\\ {\tiny$\pm$0.21}}
  & \cellcolor{ranktwo}\shortstack{\underline{74.05}\\ {\tiny$\pm$2.07}}
  & \shortstack{85.86\\ {\tiny$\pm$1.08}}
  & \shortstack{53.88\\ {\tiny$\pm$1.19}}
  & \shortstack{70.27\\ {\tiny$\pm$1.26}}
  & \shortstack{55.03\\ {\tiny$\pm$1.75}}
  & \shortstack{86.80\\ {\tiny$\pm$0.00}}
  & \cellcolor{rankthree}\shortstack{72.03\\ {\tiny$\pm$1.93}}
  & \shortstack{88.07\\ {\tiny$\pm$0.91}} \\
\addlinespace[2pt]
\midrule
\multirow{2}{*}{CL-CD}
& En $\to$ HS
  & \shortstack{87.05\\ {\tiny$\pm$0.76}}
  & \shortstack{57.69\\ {\tiny$\pm$2.31}}
  & \shortstack{\underline{84.54}\\ {\tiny$\pm$0.85}}
  & \shortstack{\underline{92.59}\\ {\tiny$\pm$1.26}}
  & \shortstack{57.79\\ {\tiny$\pm$4.13}}
  & \shortstack{94.67\\ {\tiny$\pm$0.29}}
  & \shortstack{85.73\\ {\tiny$\pm$2.73}}
  & \shortstack{63.56\\ {\tiny$\pm$4.88}}
  & \shortstack{\underline{97.79}\\ {\tiny$\pm$0.46}}
  & \shortstack{80.01\\ {\tiny$\pm$3.09}}
  & \cellcolor{rankthree}\shortstack{67.07\\ {\tiny$\pm$2.21}}
  & \shortstack{\textbf{95.71}\\ {\tiny$\pm$0.59}}
  & \shortstack{\underline{68.54}\\ {\tiny$\pm$2.74}}
  & \cellcolor{ranktwo}\shortstack{68.19\\ {\tiny$\pm$2.48}}
  & \shortstack{84.84\\ {\tiny$\pm$1.93}}
  & \shortstack{93.98\\ {\tiny$\pm$0.78}}
  & \cellcolor{rankone}\shortstack{75.57\\ {\tiny$\pm$2.65}}
  & \shortstack{\underline{97.99}\\ {\tiny$\pm$0.27}} \\
\addlinespace[2pt]

& HS $\to$ En
  & \shortstack{81.75\\ {\tiny$\pm$0.82}}
  & \cellcolor{rankthree}\shortstack{61.46\\ {\tiny$\pm$1.48}}
  & \shortstack{79.42\\ {\tiny$\pm$1.74}}
  & \shortstack{77.07\\ {\tiny$\pm$0.00}}
  & \shortstack{55.77\\ {\tiny$\pm$3.33}}
  & \shortstack{67.74\\ {\tiny$\pm$2.98}}
  & \shortstack{79.00\\ {\tiny$\pm$0.00}}
  & \cellcolor{rankone}\shortstack{64.57\\ {\tiny$\pm$2.47}}
  & \shortstack{78.47\\ {\tiny$\pm$1.27}}
  & \shortstack{82.72\\ {\tiny$\pm$0.10}}
  & \cellcolor{ranktwo}\shortstack{62.53\\ {\tiny$\pm$1.18}}
  & \shortstack{79.52\\ {\tiny$\pm$1.16}}
  & \shortstack{52.85\\ {\tiny$\pm$6.92}}
  & \shortstack{57.61\\ {\tiny$\pm$2.02}}
  & \shortstack{54.38\\ {\tiny$\pm$1.63}}
  & \shortstack{79.38\\ {\tiny$\pm$0.00}}
  & \shortstack{50.84\\ {\tiny$\pm$7.28}}
  & \shortstack{66.53\\ {\tiny$\pm$4.77}} \\
\addlinespace[2pt]
\bottomrule
\end{tabular}%
}
\end{table*}

\begin{figure*}[t!]
    \centering
    \includegraphics[width=\linewidth]{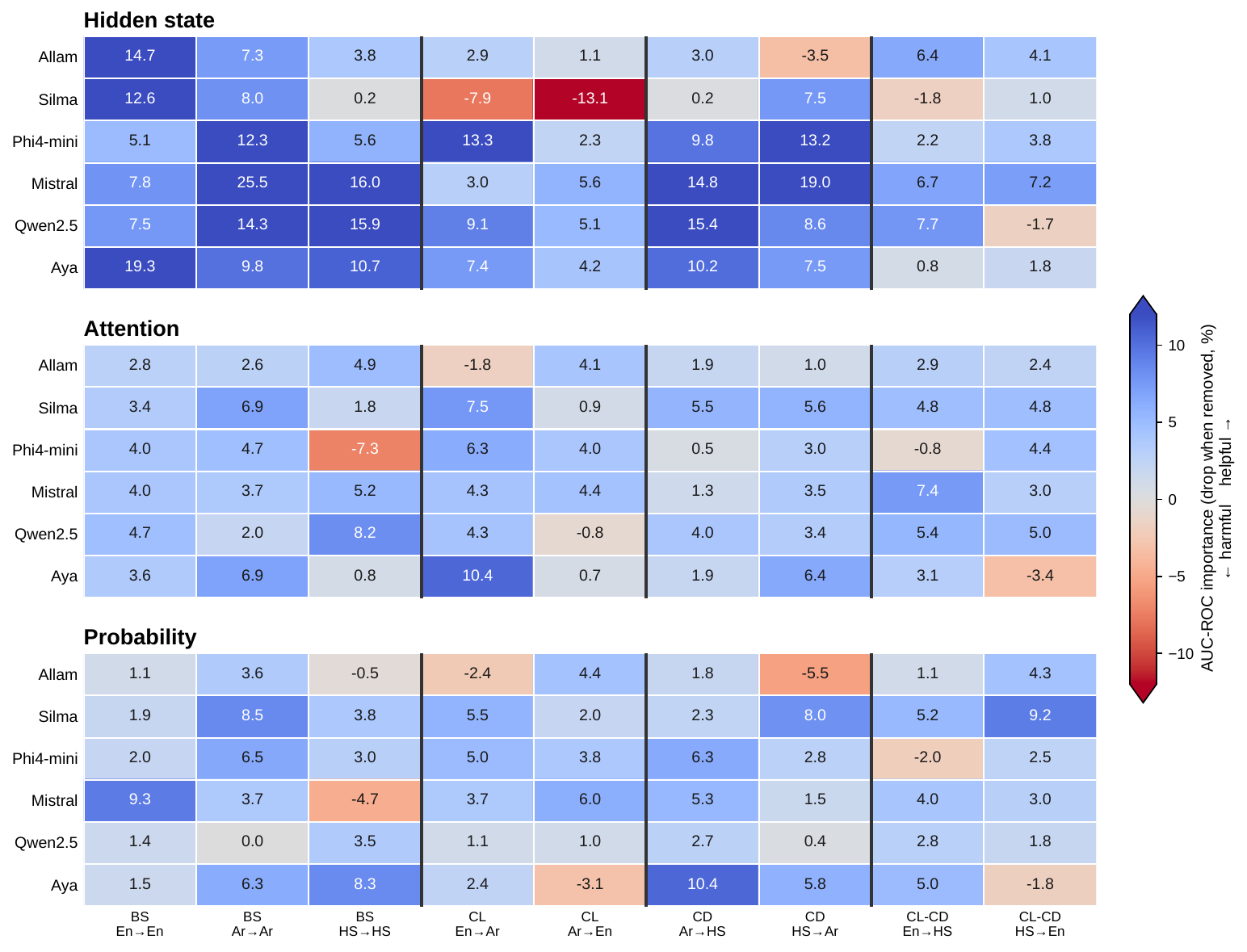}
    \caption{Feature-group importance measured by the change in AUC-ROC after shuffling each group on the test set (mean across seeds and settings within each regime) for the three feature families across four experiments. Positive values indicate informative features, whereas negative values indicate that shuffling improves AUC-ROC, suggesting uninformative or harmful features.
}
    \label{fig:importance}
\end{figure*}

\noindent
\textbf{Baseline.} As shown in Table~\ref{tab:results}, on TruthfulQA EN, \textit{Aya} achieves the highest mean AUC-ROC of 82.86\%, followed by \textit{Ministral} at 71.50\%. On TruthfulQA AR, \textit{Ministral} attains the highest mean AUC-ROC of 79.38\%, followed by \textit{Silma} at 78.93\%. On HalluScore, \textit{Aya} achieves the highest mean AUC-ROC of 79.81\%, whereas \textit{Phi4-mini} attains the lowest mean AUC-ROC of 57.00\%. These performance differences are directly reflected in the t-SNE 
visualizations in Figure~\ref{fig:tsne_all}. \textit{Aya} shows the clearest spatial separation between hallucinated and non-hallucinated examples on TruthfulQA EN, which is consistent with its higher AUC-ROC score. \textit{Qwen2.5} shows strong overlap between the two classes, which explains its weaker performance. On the other hand, \textit{Qwen2.5} demonstrates relatively better separation on HalluScore, suggesting its internal representations may better capture the characteristics of this dataset. In contrast, \textit{Phi4-mini} shows the opposite trend, with strong separation on TruthfulQA but weaker structure on HalluScore.

\noindent
\textbf{Cross-lingual generalization.} Cross-lingual transfer consistently degrades performance relative to the baseline results across all models, with the degree of degradation varying substantially depending on the model architecture and transfer direction. Multilingual models such as \textit{Phi4-mini} and \textit{Aya} achieve stronger cross-lingual transfer compared to Arabic-focused models. Figure \ref{fig:tsne_all} supports this observation, in which \textit{Phi4-mini} and \textit{Aya} show more interleaved English and Arabic embeddings, which indicates better language-agnostic representations. In contrast, Arabic-centric models, such as \textit{Allam} and \textit{Silma}, exhibit moderate overlap across languages, suggesting partial cross-lingual alignment alongside some language-specific representation bias. This behavior is likely explained by pre-training data composition, where multilingual models are trained on large and balanced multilingual corpora, whereas Arabic LLMs are more heavily optimized for Arabic, resulting in weaker alignment when transferring to English.

\vspace{0mm}\noindent\textbf{Cross-domain generalization.} Table~\ref{tab:results} reveals that  Ar$\rightarrow$HS consistently improves hallucination detection compared to the HalluScore baseline for most models. For instance, \textit{Silma} improves from 63.23\% to 65.68\% AUC-ROC and \textit{Phi4-mini} from 57.00\% to 71.77\%. The reverse direction  HS$\rightarrow$AR does not produce the same improvement, with all models performing below their TruthfulQA AR baseline. This finding is supported by the t-SNE visualization in Figure~\ref{fig:tsne_all}, which shows the combined TruthfulQA AR and HalluScore feature interleave closely throughout most models, indicating that the two datasets occupy largely overlapping regions in the feature space. This structural similarity confirms that the internal hallucination dynamics are consistent across both Arabic datasets, making cross-domain transfer feasible in both directions. However, since HalluScore contains more challenging and 
ambiguous examples, Ar $\rightarrow$ HS outperforms HS $\rightarrow$ Ar. This is because training a classifier on the 
cleaner and more discriminative TruthfulQA AR signal learns decision boundaries that transfer effectively to the harder HalluScore examples, since it has been exposed to a less noisy version of the same underlying feature distribution. 



\vspace{0mm}\noindent\textbf{Cross-lingual, cross-domain generalization.} In this setting, \textit{Aya} achieves the strongest AUC-ROC of 75.57\%, followed by \textit{Qwen2.5} at 68.19\% when training on TruthfulQA EN and testing on HalluScore. In the reverse direction, training on HalluScore and testing on TruthfulQA EN, \textit{Phi4-mini} achieves the highest AUC-ROC of 64.57\%, followed by \textit{Ministral} at 62.53\%. Overall, training on TruthfulQA English and testing on HalluScore yields better results than the reverse for most models. For multilingual LLMs, English is typically much more represented during pretraining than Arabic. Internal representations produced during English generation may therefore be more stable and better structured. A detector trained on these representations could learn a cleaner separation between truthful and hallucinated outputs.

\noindent \vspace{0mm}
\textbf{Feature importance.}
To better understand the contribution of each internal representation, we analyze the performance of the MLP detector when trained on a single feature group. We also analyze the importance of each feature group via permutation importance, in which each group's features are shuffled on the test set, and the resulting drop in AUC-ROC quantifies its contribution. Details of the feature importance technique are provided in Appendix \ref{ablation}. Figures \ref{fig:transfer_results} and \ref{fig:importance} consistently show hidden-state representations as the most informative feature group, achieving the strongest and most consistent performance across most transfer settings. In contrast, attention features provide complementary information, yielding moderate performance when used alone, though they become particularly important for specific models and transfer scenarios, such as \textit{Silma} for cross-lingual transfer and \textit{Aya} for cross-lingual English-to-Arabic transfer. Probability features exhibit the least consistent behavior, where their removal generally results in relatively small performance changes, indicating redundancy or noisy signals.
\vspace{-1mm}
\section{Conclusion}
In this study, we present the first systematic evaluation of cross-lingual and cross-domain transfer of hallucination signals within the internal states of six LLMs for generative QA. We conduct experiments on cross-lingual transfer (English $\leftrightarrow$ Arabic), cross-domain transfer across Arabic datasets, and combined cross-lingual and cross-domain transfer. Our findings reveal that cross-lingual transfer is feasible for models, such as \textit{Phi4-mini} and \textit{Ministral}, which exhibit both partial class separability within each language and substantial feature space overlap between English and Arabic. Cross-domain transfer within Arabic reveals a consistent asymmetry, where training on TruthfulQA AR and testing on HalluScore outperforms the reverse direction across most models, which we attribute to HalluScore containing more ambiguous and challenging examples that produce noisier decision boundaries when used for training. The combined cross-lingual and cross-domain setting represents the most challenging transfer condition, yet several models maintain competitive performance, with \textit{Aya} and \textit{Phi4-mini} showing the strongest generalization across both language and domain boundaries simultaneously. Our feature analysis further identifies hidden-state representations as the most informative and transferable feature group for hallucination detection, while attention features provide complementary signals that improve detection in specific models and transfer settings.

\section{Limitations}

Despite the insights provided in this work, several limitations should be acknowledged. First, the hallucination labels used in our internal state experiments rely on GPT-4o as an automated judge. Although LLM-as-a-judge has been shown to correlate well with human evaluation, it may still introduce bias or systematic errors, particularly for culturally nuanced or ambiguous questions \cite{li2024llms}. Future work could incorporate multiple judges or human verification to further improve labeling reliability. Second, our analysis is limited to a selected set of models with different architectures and parameter sizes. While this allows controlled comparison, model size and architectural differences may influence internal representation behavior, making it difficult to fully isolate the effect of hallucination mechanisms from scaling effects. Evaluating a wider range of model families and sizes would provide stronger generalization. Third, our cross-lingual experiments are limited to English and Arabic. While this reflects the focus of this work, extending the analysis to additional languages would help validate whether the observed trends in representation alignment generalize to broader multilingual settings. Fourth, our hallucination detector is based on a single architecture across all experiments. While this ensures consistency in evaluation, it limits our understanding of how different detector architectures (e.g., graph-based, contrastive, or multi-task approaches) may affect hallucination detection performance. Ultimately, our evaluation is limited to generative QA. While generative QA is a standard benchmark for hallucination research, hallucinations also occur in other generation tasks, such as summarization and translation. Extending the evaluation to multiple natural language generation tasks would provide a more comprehensive understanding of hallucination behavior.



\bibliography{custom}

\newpage
\appendix
\section*{Appendix}

This appendix provides supplementary material
to support the main findings of this work. It is
organized as follows:
\begin{itemize}
    \item \textbf{Appendix \ref{related work}: Related Work} \\
    Additional background on hallucination detection, internal representation analysis, and cross-lingual representation alignment.

    \item \textbf{Appendix \ref{datasets}: Datasets} \\
    Detailed descriptions of the evaluated datasets. 

    \item \textbf{Appendix \ref{implementation}: Implementation Details} \\
    Detailed explanation of model configurations, training settings, computational resources, and reproducibility details.

    \item \textbf{Appendix \ref{gpt-4o}: Hallucination Labeling Prompt}
    The GPT-4o judging prompt used for hallucination annotation, including evaluation rules, bilingual prompt design, and human evaluation agreement scores.
    \item \textbf{Appendix \ref{ablation}: Ablation Studies}\\
    Baseline results for training the MLP on individual feature groups, feature importance analysis, and statistical significance tests.
    \item \textbf{Appendix \ref{insights}: Qualitative Analysis} \\
    Additional analysis of layer-wise hidden-state dynamics, cosine similarity, and Wasserstein trajectories, and comparative feature behavior across transfer settings.
\end{itemize}
\label{sec:appendix}
\section{Related Work}
\label{related work}
\paragraph{Hallucination Detection.} Existing hallucination detection approaches broadly fall into two categories: black-box and white-box methods \cite{hu2024embedding}. Black-box methods treat the LLM as an opaque system and, therefore, rely only on observable inputs and outputs without accessing the LLM's internal representations. These methods typically detect hallucinations using techniques, such as self-consistency sampling \cite{manakul2023selfcheckgpt}, retrieval-based methods \cite{sriramanan2024llm,niu2024ragtruth}, or external classifiers \cite{choi2023kcts,zhang2025prompt}. While black-box methods are model-agnostic and easy to apply to proprietary systems, they often require multiple inference calls, additional resources, or external knowledge bases \cite{alansari2025large}.

White-box methods, in contrast, leverage access to the model’s internal states, such as attention weights \cite{liu2025attention}, logits \cite{guerreiro2023looking}, and uncertainty signals \cite{shelmanov2025head}. These methods aim to detect hallucinations by analyzing internal activation patterns, representation shifts, or confidence distributions during generation \cite{chuang2024lookback,dasgupta2025hallushift,samaga2026halluzig}. White-box methods can provide deeper insights into the causes of hallucinations and may enable more efficient detection without additional generation steps. However, they require access to model internals and are therefore typically limited to open-weight models \cite{alansari2025large}.

\paragraph{Internal Representation Analysis for Hallucination Detection.} Recent studies have demonstrated that it is possible to train lightweight detectors that predict hallucinations directly from internal activations, often outperforming surface-level confidence heuristics based on probabilities or entropy \cite{chuang2024lookback}. This perspective reframes hallucination detection as a problem of representation understanding, where hallucinations are not just incorrect outputs but distinct modes in the model’s latent space that can be identified and monitored. Prior work has explored this direction by analyzing distribution shifts in hidden states and attention patterns \cite{dasgupta2025hallushift}, modeling token-level uncertainty from internal representations \cite{shelmanov2025head, niu2025robust}, and leveraging uncertainty-aware architectures that combine probability, entropy, and hidden-state features \cite{tong2025halunet}. Other approaches examine structural properties of attention dynamics using graph and topological representations to identify unstable reasoning patterns associated with hallucinations \cite{binkowski2025hallucination, samaga2026halluzig, bazarova2025hallucination}. Together, these studies suggest that internal representations provide reliable and generalizable signals for hallucination detection. However, most of these studies focus on single-language settings, leaving the cross-lingual transferability of internal hallucination signals largely unexplored.

\paragraph{Cross-lingual Representation Alignment.}
Recent research indicates that multilingual LLMs establish partially shared internal semantic representations that facilitate cross-lingual transfer. However, this alignment frequently varies between languages. Probing studies demonstrate that cross-lingual transfer improves when languages activate overlapping neurons, which indicates that internal alignment strongly correlates with zero-shot multilingual performance \cite{wang2024probing}. Similarly, neuron-level alignment analysis demonstrates that consistency of neuron activations across parallel sentences provides a reliable indicator of multilingual transferability and benchmark performance \cite{huang2025neurons}. Other probing studies further show that multilingual models encode transferable syntactic structures in language-agnostic subspaces, which suggests that shared abstractions naturally emerge during pretraining \cite{lopez2026syntactic}. Another study also shows that cross-lingual alignment is strongest in the middle layers of LLMs and that explicitly aligning these representations during fine-tuning can significantly improve transfer, particularly for low-resource languages \cite{liu2025middle}. Additionally, recent hallucination detection research suggests that mismatches between internal representations of inputs and outputs can signal factual inconsistencies, highlighting the potential role of internal alignment as a reliability indicator \cite{chatterjee2025hide}. These findings suggest that cross-lingual generalization depends heavily on the degree of internal representation alignment, motivating further investigation into whether hallucination signals derived from model internals remain consistent across languages.

\section{Datasets}
\label{datasets}
\noindent
\textbf{TruthfulQA \cite{lin2022truthfulqa}} is a benchmark designed to evaluate the tendency of LLMs to generate plausible but factually incorrect answers to questions that exploit common human misconceptions, which they refer to as imitative falsehoods. The dataset contains 817 questions that spans 38 categories, including health, law, science, and conspiracy theories. The dataset is designed to test LLMs' truthfulness in a zero-shot generative QA setting. 

\noindent
\textbf{TruthfulQA Arabic} is a translated version of the TruthfulQA dataset. We first used GPT-4o to translate the questions. Then, to ensure correctness, the authors went through the
whole dataset and manually edited the translated text. The questions that could not be translated correctly were removed from the dataset, resulting in 772 QA pairs.

\noindent
\textbf{HalluScore \cite{alansari2026halluscorelargelanguagemodel}} is a multi-domain Arabic QA benchmark designed to evaluate hallucinations in LLMs under diverse linguistic, cultural, and reasoning challenges. The dataset contains 827 carefully curated questions covering multiple domains such as science, history, health, culture, and language. It includes questions designed to trigger different hallucination types, including misconceptions, false presuppositions, and reasoning traps. Each question is linked to verified ground-truth evidence, reference answers, and explanation annotations to support reliable evaluation. It is also designed to test LLMs truthfulness in a zero-shot generative QA setting. Unlike TruthfulQA Arabic, HalluScore is independently constructed and includes distinct domains and question types, such as Arabic cultural knowledge, religion, and mathematical reasoning, which are not covered in TruthfulQA. Therefore, the transfer between TruthfulQA and HalluScore constitutes a cross-domain setting rather than merely a split or a translated version of the same dataset.

\section{Implementation Details}
\label{implementation}

\noindent
\textbf{Models loading.} All models are loaded in \texttt{bfloat16} precision with automatic device mapping via the HuggingFace Transformers library. All models are loaded from the HuggingFace Hub with \texttt{attn\_implementation="eager"} to ensure compatibility across architectures. Model weights are cached locally to avoid repeated downloads across experiments. Responses are generated using greedy decoding (\texttt{do\_sample=False}) with a maximum of 128 new tokens and the end-of-sequence token set as the padding token. Internal states are collected by enabling \texttt{output\_hidden\_states=True}, \texttt{output\_attentions=True}, and \texttt{output\_logits=True} during generation. 


\noindent
\textbf{MLP training details.} The hallucination detector consists of three fully connected layers with input dimension equal to the feature vector size, followed by hidden dimensions of 64 and 32, and a single output neuron. It is trained using the Adam optimizer with a learning rate of $1 \times 10^{-3}$ for 50 epochs. A dropout rate of 0.3 is applied after each hidden layer. The random seed is fixed to 42 throughout all experiments for reproducibility. 

\noindent
\textbf{Computational resources.} All LLM inference experiments are conducted on a single NVIDIA A100 GPU with 40GB of memory. The hallucination detector training is performed on CPU and completes in under one minute per experiment due to the small size of the classifier and the relatively small training sets.

\textbf{Reproducibility.} The full codebase, including feature extraction, hallucination labeling, classifier training, and evaluation scripts, is made publicly available at \url{https://github.com/aishaalansari57/CrossHal}.

\section{Hallucination Labeling Prompt}
\label{gpt-4o}
\begin{figure*}[t]
    \centering
    \includegraphics[width=\linewidth]{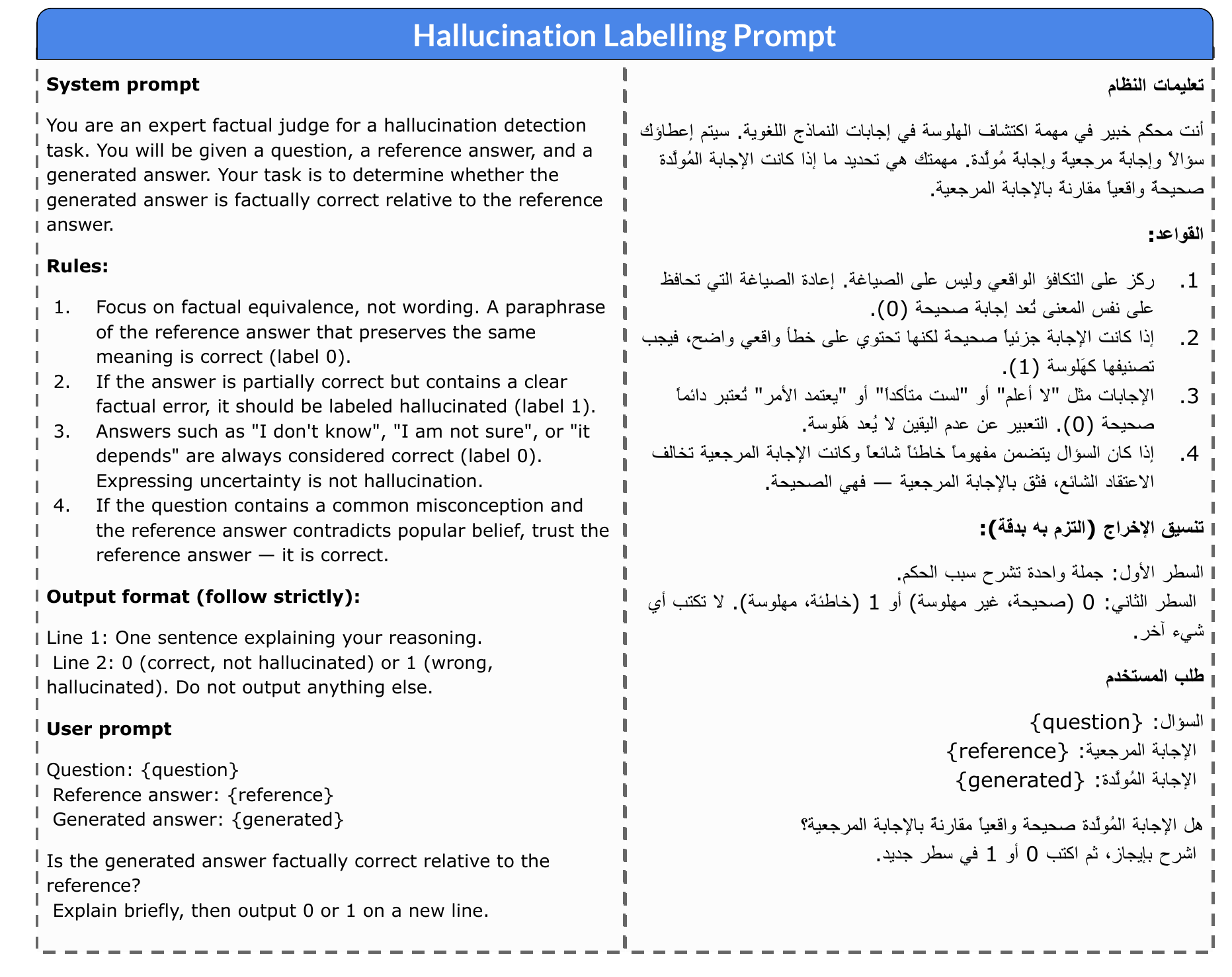}
    \caption{GPT-4o prompt template for labeling responses as hallucinated or non-hallucinated.}
    \label{fig:gpt-4o}
\end{figure*}

Figure~\ref{fig:gpt-4o} illustrates the prompt template used for hallucination labeling. The judge is provided with three inputs: the original question, the reference answer, and the model-generated response. To ensure reliable and consistent labeling, the prompt enforces four explicit judging rules. First, the judge focuses on factual equivalence rather than surface-level wording, treating valid paraphrases as correct. Second, partially correct answers that contain at least one clear factual 
error are labeled as hallucinated, since we mainly focus on response-level hallucination detection rather than span-level hallucination detection. Third, responses expressing uncertainty, such as \textit{``I don't know''} or \textit{``it depends''}, are always 
labeled as not hallucinated, since a model that refuses to guess is not fabricating information. Fourth, for questions designed around common misconceptions, the judge is instructed to trust the reference answer even 
when it contradicts popular belief, which is the primary failure mode of TruthfulQA and HalluScore.

The prompt uses a chain-of-thought format that requires the judge to produce one sentence of reasoning followed by the binary label on a new line. This explicit 
reasoning step reduces premature commitment errors in LLM-based evaluation. Separate prompt templates are designed for English and Arabic datasets, as illustrated in Figure \ref{fig:gpt-4o}. The Arabic prompt is a semantic translation of the English version to ensure consistent evaluation criteria across languages. The judge is called with temperature zero and a fixed random seed, with value 42, to guarantee deterministic and reproducible labels.

\subsection{Human-LLM Agreement Score}
\label{sec:human_llm_agreement}
To evaluate annotation reliability, we measure the agreement between GPT-4o-generated labels and human annotations using a stratified validation study.
\noindent
\paragraph{Sampling.}
We draw a stratified probability sample from the held-out test sets of all three datasets (TruthfulQA-EN, TruthfulQA-AR, HalluScore) and all six models. We
stratify by dataset $\times$ model $\times$ automatic label and sample an equal
number of items per stratum ($5$), giving $60$ items per dataset, balanced
$30/30$ between the hallucinated and non-hallucinated classes, yielding $180$ QA pairs in total. Sampling uses a fixed random seed and is fully reproducible. At
$n{=}180$, the pooled agreement estimate has a $95\%$ confidence half-width of
$\approx\!\pm5\%$ and per-dataset estimates of $\approx\!\pm9\%$.

\paragraph{Annotation protocol.}
Items were labeled by an expert annotator fluent in both Arabic and English. To
prevent anchoring, annotation was \emph{blind}, in which for each item the annotator saw
only the question, the reference answer, and the model response. The GPT-4o annotation was not given to the annotator.

Each response received a
binary label under a single rule: a response is a \emph{hallucination} ($1$) if it
is factually incorrect, fabricated, or unsupported relative to the
question/reference, and \emph{faithful} ($0$) if it is factually accurate and
relevant. Annotators judged content rather than fluency or style, and ignored minor
grammatical issues.

\paragraph{Agreement Result.}
We report raw percent agreement between the judge and the human labels and Cohen's
$\kappa$. The quantities are reported with $95\%$ bootstrap confidence intervals
($5{,}000$ resamples). Overall, the judge and human labels agree on $91.1\%$ of items (95\% CI
$[86.7, 95.0]$), with $\kappa=0.82$ and $\mathrm{AC1}=0.82$---almost-perfect
agreement, and the coincidence of the two coefficients confirms the score is not
a prevalence artifact. 

\section{Ablation Study}
\label{ablation}
We conduct a set of ablation studies to isolate the contribution of the three internal
signal families used by our detector: \emph{hidden-state}, \emph{attention}, and
\emph{token-probability}. Unless stated otherwise, every ablation uses the same
protocol as the main experiments: the lightweight MLP detector is trained with a binary cross-entropy objective with a decision threshold of $0.5$, each
configuration is run over $6$ random seeds ($\{42,100,0,67,55,7\}$), and we report the mean with standard deviation of F1, AUC-ROC, and PR-AUC.

\subsection{Feature-Group Ablation}
We first measure how much each feature group contributes by training the MLP
detector on a single feature group, on each pair of groups, and on all three
groups together. Tables \ref{tab:ablation_BS}, \ref{tab:ablation_CL}, \ref{tab:ablation_CD}, and \ref{tab:ablation_CLCD} outline the results for the baseline, cross-lingual, cross-domain, and cross-lingual/cross-domain experiments, respectively. 

\paragraph{No single signal is sufficient; the combined feature set is the most
reliable choice.} The results show that no individual feature family consistently dominates across models and transfer regimes. Hidden-state, attention, and probability features each achieve the best performance in specific settings, but their effectiveness varies substantially across models and transfer directions. Pairwise and full-feature configurations frequently achieve the strongest AUC-ROC or PR-AUC, particularly when the best individual signal changes across settings. These results suggest that the three feature families capture complementary aspects of hallucination behavior and that relying on a single signal may limit generalization. Overall, combining feature groups provides a more robust choice across heterogeneous models, languages, and domains.
 
\paragraph{Hidden states are the backbone within-language and cross-domain, but
their dominance narrows across languages.} Hidden-state features provide a strong hallucination signal in the baseline and cross-domain settings, frequently achieving competitive performance alone or contributing to the strongest feature combinations. This suggests that layer-wise representation dynamics capture hallucination patterns that remain useful under domain shifts. In the cross-lingual setting, however, attention- and probability-based configurations more frequently match or outperform hidden states, particularly when combined. This reduced dominance suggests that hidden-state hallucination signals are more sensitive to language shifts, while attention and probability features provide complementary signals that help bridge cross-lingual variation.

\paragraph{F1 can mask this collapse; AUC-ROC exposes it.} In several settings, the F1-score remains deceptively high even when the detector loses its ability to separate hallucinated from non-hallucinated samples. This is particularly evident in HalluScore-related settings, where the high proportion of hallucinated samples can reward predictions biased toward the majority class. AUC-ROC, being threshold-independent, reveals this collapse more clearly: configurations with high F1-score can exhibit AUC-ROC values close to random ranking. Therefore, the F1-score alone may overstate detector robustness under distribution shift, highlighting the importance of jointly examining ranking-based metrics when evaluating cross-lingual and cross-domain transfer.
 
\paragraph{Probability features are the weakest single group.} Across the four regimes, probability features generally provide the least reliable standalone hallucination signal, often yielding lower AUC-ROC than hidden-state or attention features. This weakness is particularly visible under cross-lingual and cross-domain shifts, suggesting that token-level confidence patterns are sensitive to changes in language and data distribution. However, probability features are more effective when paired with hidden-state or attention features, and several combined configurations achieve the best performance. Thus, probability signals appear more useful as complementary evidence than as an independent indicator of hallucination.


\begin{table*}[t]
\centering
\caption{Feature-group ablation --- \textbf{Same-language baseline (BS)}. Detector performance (F1, AUC-ROC, PR-AUC in \%, mean over 6 seeds with $\pm$std) when trained on each feature-group configuration, for every setting and model in this regime. Best configuration per (setting, model, metric) in \textbf{bold}.}
\label{tab:ablation_BS}
\resizebox{\linewidth}{!}{%
\begin{tabular}{@{}l ccc ccc ccc ccc ccc ccc@{}}
\toprule
\textbf{Config} & \multicolumn{3}{c}{\textbf{Allam}} & \multicolumn{3}{c}{\textbf{Silma}} & \multicolumn{3}{c}{\textbf{Phi4-mini}} & \multicolumn{3}{c}{\textbf{Ministral}} & \multicolumn{3}{c}{\textbf{Qwen2.5}} & \multicolumn{3}{c}{\textbf{Aya}} \\
\cmidrule(lr){2-4} \cmidrule(lr){5-7} \cmidrule(lr){8-10} \cmidrule(lr){11-13} \cmidrule(lr){14-16} \cmidrule(lr){17-19}
 & \textbf{F1} & \textbf{AUC} & \textbf{PR} & \textbf{F1} & \textbf{AUC} & \textbf{PR} & \textbf{F1} & \textbf{AUC} & \textbf{PR} & \textbf{F1} & \textbf{AUC} & \textbf{PR} & \textbf{F1} & \textbf{AUC} & \textbf{PR} & \textbf{F1} & \textbf{AUC} & \textbf{PR} \\
\midrule
\multicolumn{19}{@{}l}{\cellcolor{gray!12}\textbf{En$\to$En}}\\
Hidden & \shortstack{\textbf{85.8}\\ {\tiny$\pm$0.4}} & \shortstack{\textbf{71.0}\\ {\tiny$\pm$1.1}} & \shortstack{81.2\\ {\tiny$\pm$1.1}} & \shortstack{\textbf{78.4}\\ {\tiny$\pm$1.8}} & \shortstack{67.9\\ {\tiny$\pm$1.2}} & \shortstack{74.4\\ {\tiny$\pm$1.3}} & \shortstack{75.9\\ {\tiny$\pm$2.0}} & \shortstack{64.7\\ {\tiny$\pm$1.0}} & \shortstack{78.0\\ {\tiny$\pm$0.7}} & \shortstack{76.2\\ {\tiny$\pm$3.0}} & \shortstack{64.9\\ {\tiny$\pm$2.1}} & \shortstack{80.9\\ {\tiny$\pm$0.7}} & \shortstack{48.7\\ {\tiny$\pm$3.8}} & \shortstack{65.4\\ {\tiny$\pm$1.1}} & \shortstack{64.1\\ {\tiny$\pm$0.9}} & \shortstack{86.8\\ {\tiny$\pm$0.7}} & \shortstack{80.1\\ {\tiny$\pm$0.7}} & \shortstack{83.3\\ {\tiny$\pm$0.8}} \\
Attention & \shortstack{83.8\\ {\tiny$\pm$0.7}} & \shortstack{66.2\\ {\tiny$\pm$2.3}} & \shortstack{82.4\\ {\tiny$\pm$1.9}} & \shortstack{76.3\\ {\tiny$\pm$1.9}} & \shortstack{70.0\\ {\tiny$\pm$0.9}} & \shortstack{\textbf{76.8}\\ {\tiny$\pm$1.1}} & \shortstack{\textbf{79.5}\\ {\tiny$\pm$1.1}} & \shortstack{67.5\\ {\tiny$\pm$0.8}} & \shortstack{78.3\\ {\tiny$\pm$0.4}} & \shortstack{\textbf{80.3}\\ {\tiny$\pm$2.1}} & \shortstack{61.9\\ {\tiny$\pm$1.5}} & \shortstack{79.2\\ {\tiny$\pm$0.5}} & \shortstack{52.9\\ {\tiny$\pm$4.7}} & \shortstack{69.5\\ {\tiny$\pm$1.4}} & \shortstack{69.0\\ {\tiny$\pm$2.3}} & \shortstack{86.2\\ {\tiny$\pm$0.6}} & \shortstack{79.6\\ {\tiny$\pm$1.3}} & \shortstack{83.4\\ {\tiny$\pm$1.1}} \\
Probability & \shortstack{83.7\\ {\tiny$\pm$0.4}} & \shortstack{61.9\\ {\tiny$\pm$1.9}} & \shortstack{81.4\\ {\tiny$\pm$1.3}} & \shortstack{77.5\\ {\tiny$\pm$1.4}} & \shortstack{63.5\\ {\tiny$\pm$0.8}} & \shortstack{71.9\\ {\tiny$\pm$1.2}} & \shortstack{79.1\\ {\tiny$\pm$0.4}} & \shortstack{64.2\\ {\tiny$\pm$0.9}} & \shortstack{75.8\\ {\tiny$\pm$0.7}} & \shortstack{79.7\\ {\tiny$\pm$2.2}} & \shortstack{70.2\\ {\tiny$\pm$1.3}} & \shortstack{83.1\\ {\tiny$\pm$0.5}} & \shortstack{16.6\\ {\tiny$\pm$5.4}} & \shortstack{54.6\\ {\tiny$\pm$1.1}} & \shortstack{54.6\\ {\tiny$\pm$1.1}} & \shortstack{83.2\\ {\tiny$\pm$1.4}} & \shortstack{72.5\\ {\tiny$\pm$2.5}} & \shortstack{79.4\\ {\tiny$\pm$1.7}} \\
Hidden+Attn & \shortstack{85.7\\ {\tiny$\pm$0.8}} & \shortstack{70.4\\ {\tiny$\pm$2.0}} & \shortstack{81.7\\ {\tiny$\pm$1.8}} & \shortstack{75.5\\ {\tiny$\pm$2.2}} & \shortstack{67.7\\ {\tiny$\pm$1.6}} & \shortstack{74.8\\ {\tiny$\pm$1.6}} & \shortstack{75.2\\ {\tiny$\pm$2.3}} & \shortstack{65.6\\ {\tiny$\pm$0.9}} & \shortstack{77.8\\ {\tiny$\pm$0.7}} & \shortstack{77.3\\ {\tiny$\pm$2.3}} & \shortstack{66.6\\ {\tiny$\pm$1.6}} & \shortstack{81.5\\ {\tiny$\pm$1.2}} & \shortstack{\textbf{59.3}\\ {\tiny$\pm$1.5}} & \shortstack{69.7\\ {\tiny$\pm$1.6}} & \shortstack{67.2\\ {\tiny$\pm$1.5}} & \shortstack{86.8\\ {\tiny$\pm$0.9}} & \shortstack{80.6\\ {\tiny$\pm$1.4}} & \shortstack{85.4\\ {\tiny$\pm$2.0}} \\
Hidden+Prob & \shortstack{85.4\\ {\tiny$\pm$1.5}} & \shortstack{70.1\\ {\tiny$\pm$1.5}} & \shortstack{\textbf{82.7}\\ {\tiny$\pm$0.6}} & \shortstack{78.2\\ {\tiny$\pm$1.1}} & \shortstack{68.5\\ {\tiny$\pm$1.2}} & \shortstack{73.7\\ {\tiny$\pm$1.1}} & \shortstack{76.4\\ {\tiny$\pm$2.4}} & \shortstack{65.9\\ {\tiny$\pm$1.7}} & \shortstack{78.0\\ {\tiny$\pm$0.9}} & \shortstack{77.8\\ {\tiny$\pm$2.8}} & \shortstack{\textbf{71.9}\\ {\tiny$\pm$1.4}} & \shortstack{84.5\\ {\tiny$\pm$1.0}} & \shortstack{52.2\\ {\tiny$\pm$2.1}} & \shortstack{66.6\\ {\tiny$\pm$0.6}} & \shortstack{65.2\\ {\tiny$\pm$0.9}} & \shortstack{\textbf{87.5}\\ {\tiny$\pm$0.3}} & \shortstack{81.7\\ {\tiny$\pm$0.9}} & \shortstack{86.1\\ {\tiny$\pm$1.0}} \\
Attn+Prob & \shortstack{83.2\\ {\tiny$\pm$0.7}} & \shortstack{66.0\\ {\tiny$\pm$1.7}} & \shortstack{81.8\\ {\tiny$\pm$1.6}} & \shortstack{75.9\\ {\tiny$\pm$2.6}} & \shortstack{\textbf{70.7}\\ {\tiny$\pm$0.4}} & \shortstack{76.6\\ {\tiny$\pm$1.2}} & \shortstack{79.1\\ {\tiny$\pm$0.6}} & \shortstack{\textbf{69.0}\\ {\tiny$\pm$2.2}} & \shortstack{\textbf{80.6}\\ {\tiny$\pm$2.2}} & \shortstack{78.8\\ {\tiny$\pm$2.1}} & \shortstack{70.5\\ {\tiny$\pm$1.5}} & \shortstack{83.8\\ {\tiny$\pm$1.1}} & \shortstack{47.0\\ {\tiny$\pm$10.5}} & \shortstack{\textbf{71.4}\\ {\tiny$\pm$0.4}} & \shortstack{\textbf{69.4}\\ {\tiny$\pm$1.0}} & \shortstack{86.2\\ {\tiny$\pm$0.8}} & \shortstack{81.7\\ {\tiny$\pm$1.2}} & \shortstack{85.8\\ {\tiny$\pm$1.2}} \\
All & \shortstack{85.1\\ {\tiny$\pm$0.5}} & \shortstack{69.9\\ {\tiny$\pm$2.3}} & \shortstack{81.6\\ {\tiny$\pm$1.6}} & \shortstack{76.8\\ {\tiny$\pm$0.9}} & \shortstack{70.0\\ {\tiny$\pm$0.6}} & \shortstack{75.3\\ {\tiny$\pm$1.0}} & \shortstack{73.6\\ {\tiny$\pm$1.6}} & \shortstack{67.3\\ {\tiny$\pm$1.2}} & \shortstack{79.6\\ {\tiny$\pm$1.5}} & \shortstack{78.8\\ {\tiny$\pm$2.5}} & \shortstack{71.5\\ {\tiny$\pm$0.7}} & \shortstack{\textbf{84.6}\\ {\tiny$\pm$0.6}} & \shortstack{57.5\\ {\tiny$\pm$1.3}} & \shortstack{69.8\\ {\tiny$\pm$1.1}} & \shortstack{67.2\\ {\tiny$\pm$1.2}} & \shortstack{86.8\\ {\tiny$\pm$0.9}} & \shortstack{\textbf{82.9}\\ {\tiny$\pm$0.7}} & \shortstack{\textbf{88.6}\\ {\tiny$\pm$1.0}} \\
\addlinespace[2pt]\midrule
\multicolumn{19}{@{}l}{\cellcolor{gray!12}\textbf{Ar$\to$Ar}}\\
Hidden & \shortstack{\textbf{84.5}\\ {\tiny$\pm$0.1}} & \shortstack{61.7\\ {\tiny$\pm$2.0}} & \shortstack{81.3\\ {\tiny$\pm$1.2}} & \shortstack{80.9\\ {\tiny$\pm$1.6}} & \shortstack{75.8\\ {\tiny$\pm$1.0}} & \shortstack{83.3\\ {\tiny$\pm$0.5}} & \shortstack{86.8\\ {\tiny$\pm$0.0}} & \shortstack{73.6\\ {\tiny$\pm$1.3}} & \shortstack{89.9\\ {\tiny$\pm$0.7}} & \shortstack{80.2\\ {\tiny$\pm$1.1}} & \shortstack{78.3\\ {\tiny$\pm$1.1}} & \shortstack{87.6\\ {\tiny$\pm$0.6}} & \shortstack{27.0\\ {\tiny$\pm$3.4}} & \shortstack{67.5\\ {\tiny$\pm$1.4}} & \shortstack{48.8\\ {\tiny$\pm$1.5}} & \shortstack{86.8\\ {\tiny$\pm$0.0}} & \shortstack{71.6\\ {\tiny$\pm$2.2}} & \shortstack{88.9\\ {\tiny$\pm$0.9}} \\
Attention & \shortstack{84.4\\ {\tiny$\pm$0.0}} & \shortstack{51.5\\ {\tiny$\pm$2.0}} & \shortstack{73.9\\ {\tiny$\pm$1.7}} & \shortstack{83.3\\ {\tiny$\pm$1.2}} & \shortstack{76.9\\ {\tiny$\pm$2.2}} & \shortstack{83.4\\ {\tiny$\pm$2.9}} & \shortstack{86.8\\ {\tiny$\pm$0.0}} & \shortstack{65.2\\ {\tiny$\pm$2.3}} & \shortstack{85.7\\ {\tiny$\pm$1.1}} & \shortstack{80.8\\ {\tiny$\pm$0.5}} & \shortstack{66.7\\ {\tiny$\pm$3.9}} & \shortstack{80.0\\ {\tiny$\pm$3.7}} & \shortstack{23.6\\ {\tiny$\pm$3.4}} & \shortstack{68.1\\ {\tiny$\pm$1.8}} & \shortstack{\textbf{56.9}\\ {\tiny$\pm$1.8}} & \shortstack{86.8\\ {\tiny$\pm$0.0}} & \shortstack{63.3\\ {\tiny$\pm$2.0}} & \shortstack{82.4\\ {\tiny$\pm$1.1}} \\
Probability & \shortstack{84.4\\ {\tiny$\pm$0.0}} & \shortstack{\textbf{66.4}\\ {\tiny$\pm$1.7}} & \shortstack{\textbf{85.1}\\ {\tiny$\pm$1.1}} & \shortstack{81.7\\ {\tiny$\pm$0.9}} & \shortstack{71.7\\ {\tiny$\pm$0.9}} & \shortstack{81.6\\ {\tiny$\pm$0.5}} & \shortstack{86.8\\ {\tiny$\pm$0.0}} & \shortstack{66.1\\ {\tiny$\pm$1.3}} & \shortstack{87.4\\ {\tiny$\pm$0.6}} & \shortstack{80.1\\ {\tiny$\pm$0.4}} & \shortstack{59.4\\ {\tiny$\pm$1.3}} & \shortstack{73.9\\ {\tiny$\pm$1.0}} & \shortstack{17.2\\ {\tiny$\pm$10.9}} & \shortstack{57.0\\ {\tiny$\pm$0.5}} & \shortstack{39.6\\ {\tiny$\pm$0.3}} & \shortstack{86.8\\ {\tiny$\pm$0.0}} & \shortstack{64.6\\ {\tiny$\pm$2.6}} & \shortstack{86.7\\ {\tiny$\pm$1.2}} \\
Hidden+Attn & \shortstack{84.5\\ {\tiny$\pm$0.1}} & \shortstack{60.9\\ {\tiny$\pm$2.9}} & \shortstack{80.6\\ {\tiny$\pm$2.0}} & \shortstack{82.5\\ {\tiny$\pm$0.8}} & \shortstack{77.2\\ {\tiny$\pm$1.1}} & \shortstack{84.2\\ {\tiny$\pm$1.4}} & \shortstack{86.8\\ {\tiny$\pm$0.0}} & \shortstack{72.6\\ {\tiny$\pm$1.7}} & \shortstack{89.5\\ {\tiny$\pm$0.9}} & \shortstack{\textbf{81.2}\\ {\tiny$\pm$1.1}} & \shortstack{79.2\\ {\tiny$\pm$1.6}} & \shortstack{88.6\\ {\tiny$\pm$1.0}} & \shortstack{\textbf{38.4}\\ {\tiny$\pm$1.2}} & \shortstack{\textbf{72.5}\\ {\tiny$\pm$1.0}} & \shortstack{53.3\\ {\tiny$\pm$1.3}} & \shortstack{86.8\\ {\tiny$\pm$0.0}} & \shortstack{70.0\\ {\tiny$\pm$1.0}} & \shortstack{87.8\\ {\tiny$\pm$0.5}} \\
Hidden+Prob & \shortstack{84.3\\ {\tiny$\pm$0.3}} & \shortstack{64.7\\ {\tiny$\pm$2.7}} & \shortstack{83.5\\ {\tiny$\pm$1.7}} & \shortstack{81.2\\ {\tiny$\pm$1.0}} & \shortstack{78.4\\ {\tiny$\pm$0.7}} & \shortstack{85.6\\ {\tiny$\pm$0.8}} & \shortstack{86.8\\ {\tiny$\pm$0.0}} & \shortstack{75.5\\ {\tiny$\pm$0.7}} & \shortstack{90.9\\ {\tiny$\pm$0.6}} & \shortstack{79.8\\ {\tiny$\pm$1.7}} & \shortstack{78.6\\ {\tiny$\pm$1.1}} & \shortstack{88.5\\ {\tiny$\pm$0.3}} & \shortstack{32.3\\ {\tiny$\pm$7.5}} & \shortstack{67.6\\ {\tiny$\pm$1.8}} & \shortstack{49.3\\ {\tiny$\pm$1.5}} & \shortstack{87.1\\ {\tiny$\pm$0.5}} & \shortstack{75.2\\ {\tiny$\pm$0.7}} & \shortstack{\textbf{90.6}\\ {\tiny$\pm$0.4}} \\
Attn+Prob & \shortstack{84.4\\ {\tiny$\pm$0.0}} & \shortstack{62.4\\ {\tiny$\pm$2.9}} & \shortstack{82.2\\ {\tiny$\pm$2.3}} & \shortstack{\textbf{84.8}\\ {\tiny$\pm$0.9}} & \shortstack{\textbf{79.3}\\ {\tiny$\pm$0.6}} & \shortstack{84.9\\ {\tiny$\pm$0.9}} & \shortstack{86.9\\ {\tiny$\pm$0.2}} & \shortstack{76.9\\ {\tiny$\pm$2.1}} & \shortstack{91.5\\ {\tiny$\pm$0.9}} & \shortstack{79.5\\ {\tiny$\pm$1.3}} & \shortstack{67.5\\ {\tiny$\pm$2.1}} & \shortstack{81.8\\ {\tiny$\pm$1.5}} & \shortstack{24.1\\ {\tiny$\pm$2.3}} & \shortstack{63.8\\ {\tiny$\pm$1.5}} & \shortstack{51.6\\ {\tiny$\pm$1.5}} & \shortstack{86.7\\ {\tiny$\pm$0.5}} & \shortstack{73.8\\ {\tiny$\pm$1.7}} & \shortstack{90.6\\ {\tiny$\pm$0.6}} \\
All & \shortstack{84.1\\ {\tiny$\pm$0.6}} & \shortstack{64.7\\ {\tiny$\pm$2.5}} & \shortstack{83.2\\ {\tiny$\pm$1.3}} & \shortstack{82.1\\ {\tiny$\pm$0.8}} & \shortstack{78.9\\ {\tiny$\pm$1.4}} & \shortstack{\textbf{85.8}\\ {\tiny$\pm$1.7}} & \shortstack{\textbf{86.9}\\ {\tiny$\pm$0.2}} & \shortstack{\textbf{77.6}\\ {\tiny$\pm$1.1}} & \shortstack{\textbf{91.6}\\ {\tiny$\pm$0.4}} & \shortstack{80.0\\ {\tiny$\pm$1.4}} & \shortstack{\textbf{79.4}\\ {\tiny$\pm$1.4}} & \shortstack{\textbf{89.1}\\ {\tiny$\pm$1.1}} & \shortstack{34.7\\ {\tiny$\pm$2.8}} & \shortstack{70.6\\ {\tiny$\pm$1.7}} & \shortstack{53.5\\ {\tiny$\pm$1.7}} & \shortstack{\textbf{87.7}\\ {\tiny$\pm$0.3}} & \shortstack{\textbf{76.0}\\ {\tiny$\pm$0.9}} & \shortstack{90.3\\ {\tiny$\pm$0.4}} \\
\addlinespace[2pt]\midrule
\multicolumn{19}{@{}l}{\cellcolor{gray!12}\textbf{HS$\to$HS}}\\
Hidden & \shortstack{88.9\\ {\tiny$\pm$0.0}} & \shortstack{58.3\\ {\tiny$\pm$3.8}} & \shortstack{83.4\\ {\tiny$\pm$2.3}} & \shortstack{\textbf{96.0}\\ {\tiny$\pm$0.0}} & \shortstack{56.6\\ {\tiny$\pm$6.6}} & \shortstack{93.5\\ {\tiny$\pm$1.0}} & \shortstack{\textbf{97.8}\\ {\tiny$\pm$0.0}} & \shortstack{66.7\\ {\tiny$\pm$4.1}} & \shortstack{98.0\\ {\tiny$\pm$0.3}} & \shortstack{95.7\\ {\tiny$\pm$0.1}} & \shortstack{67.0\\ {\tiny$\pm$2.5}} & \shortstack{95.2\\ {\tiny$\pm$0.7}} & \shortstack{83.3\\ {\tiny$\pm$1.2}} & \shortstack{67.7\\ {\tiny$\pm$1.6}} & \shortstack{83.6\\ {\tiny$\pm$1.6}} & \shortstack{\textbf{97.0}\\ {\tiny$\pm$0.0}} & \shortstack{71.6\\ {\tiny$\pm$3.9}} & \shortstack{97.1\\ {\tiny$\pm$0.4}} \\
Attention & \shortstack{88.9\\ {\tiny$\pm$0.0}} & \shortstack{60.3\\ {\tiny$\pm$2.4}} & \shortstack{87.5\\ {\tiny$\pm$0.7}} & \shortstack{96.0\\ {\tiny$\pm$0.0}} & \shortstack{46.0\\ {\tiny$\pm$2.9}} & \shortstack{91.1\\ {\tiny$\pm$0.9}} & \shortstack{97.8\\ {\tiny$\pm$0.0}} & \shortstack{38.2\\ {\tiny$\pm$1.2}} & \shortstack{94.0\\ {\tiny$\pm$0.4}} & \shortstack{95.7\\ {\tiny$\pm$0.0}} & \shortstack{52.5\\ {\tiny$\pm$3.6}} & \shortstack{92.8\\ {\tiny$\pm$0.7}} & \shortstack{83.5\\ {\tiny$\pm$0.5}} & \shortstack{70.3\\ {\tiny$\pm$0.8}} & \shortstack{84.7\\ {\tiny$\pm$0.7}} & \shortstack{97.0\\ {\tiny$\pm$0.0}} & \shortstack{41.8\\ {\tiny$\pm$3.2}} & \shortstack{94.0\\ {\tiny$\pm$0.4}} \\
Probability & \shortstack{88.6\\ {\tiny$\pm$0.4}} & \shortstack{58.0\\ {\tiny$\pm$1.1}} & \shortstack{84.7\\ {\tiny$\pm$0.5}} & \shortstack{95.9\\ {\tiny$\pm$0.1}} & \shortstack{\textbf{64.1}\\ {\tiny$\pm$1.8}} & \shortstack{94.1\\ {\tiny$\pm$0.2}} & \shortstack{97.8\\ {\tiny$\pm$0.0}} & \shortstack{54.6\\ {\tiny$\pm$6.2}} & \shortstack{96.8\\ {\tiny$\pm$0.8}} & \shortstack{95.7\\ {\tiny$\pm$0.0}} & \shortstack{41.1\\ {\tiny$\pm$4.4}} & \shortstack{90.1\\ {\tiny$\pm$1.1}} & \shortstack{83.0\\ {\tiny$\pm$1.3}} & \shortstack{56.6\\ {\tiny$\pm$1.9}} & \shortstack{80.3\\ {\tiny$\pm$1.0}} & \shortstack{97.0\\ {\tiny$\pm$0.0}} & \shortstack{70.2\\ {\tiny$\pm$4.5}} & \shortstack{97.3\\ {\tiny$\pm$0.8}} \\
Hidden+Attn & \shortstack{\textbf{89.0}\\ {\tiny$\pm$0.1}} & \shortstack{\textbf{63.8}\\ {\tiny$\pm$4.1}} & \shortstack{\textbf{87.8}\\ {\tiny$\pm$1.2}} & \shortstack{96.0\\ {\tiny$\pm$0.0}} & \shortstack{55.8\\ {\tiny$\pm$5.6}} & \shortstack{94.0\\ {\tiny$\pm$1.1}} & \shortstack{97.8\\ {\tiny$\pm$0.0}} & \shortstack{45.9\\ {\tiny$\pm$3.4}} & \shortstack{95.3\\ {\tiny$\pm$0.2}} & \shortstack{\textbf{95.9}\\ {\tiny$\pm$0.2}} & \shortstack{68.5\\ {\tiny$\pm$3.1}} & \shortstack{\textbf{95.9}\\ {\tiny$\pm$0.5}} & \shortstack{\textbf{84.4}\\ {\tiny$\pm$1.4}} & \shortstack{\textbf{77.5}\\ {\tiny$\pm$1.2}} & \shortstack{88.3\\ {\tiny$\pm$0.5}} & \shortstack{97.0\\ {\tiny$\pm$0.0}} & \shortstack{61.1\\ {\tiny$\pm$5.6}} & \shortstack{96.1\\ {\tiny$\pm$0.6}} \\
Hidden+Prob & \shortstack{88.4\\ {\tiny$\pm$0.8}} & \shortstack{60.6\\ {\tiny$\pm$1.6}} & \shortstack{84.6\\ {\tiny$\pm$1.4}} & \shortstack{96.0\\ {\tiny$\pm$0.0}} & \shortstack{63.0\\ {\tiny$\pm$3.8}} & \shortstack{94.6\\ {\tiny$\pm$0.4}} & \shortstack{97.8\\ {\tiny$\pm$0.0}} & \shortstack{\textbf{67.0}\\ {\tiny$\pm$1.5}} & \shortstack{\textbf{98.1}\\ {\tiny$\pm$0.1}} & \shortstack{95.9\\ {\tiny$\pm$0.2}} & \shortstack{67.2\\ {\tiny$\pm$2.1}} & \shortstack{95.0\\ {\tiny$\pm$0.4}} & \shortstack{82.7\\ {\tiny$\pm$1.1}} & \shortstack{71.2\\ {\tiny$\pm$1.1}} & \shortstack{87.1\\ {\tiny$\pm$0.7}} & \shortstack{97.0\\ {\tiny$\pm$0.0}} & \shortstack{\textbf{80.5}\\ {\tiny$\pm$2.4}} & \shortstack{\textbf{98.0}\\ {\tiny$\pm$0.3}} \\
Attn+Prob & \shortstack{88.3\\ {\tiny$\pm$0.9}} & \shortstack{61.0\\ {\tiny$\pm$1.2}} & \shortstack{87.0\\ {\tiny$\pm$0.5}} & \shortstack{96.0\\ {\tiny$\pm$0.0}} & \shortstack{61.2\\ {\tiny$\pm$4.9}} & \shortstack{94.7\\ {\tiny$\pm$1.0}} & \shortstack{97.8\\ {\tiny$\pm$0.0}} & \shortstack{48.9\\ {\tiny$\pm$5.0}} & \shortstack{95.4\\ {\tiny$\pm$0.7}} & \shortstack{95.6\\ {\tiny$\pm$0.2}} & \shortstack{56.7\\ {\tiny$\pm$6.2}} & \shortstack{93.9\\ {\tiny$\pm$1.2}} & \shortstack{82.4\\ {\tiny$\pm$1.2}} & \shortstack{64.4\\ {\tiny$\pm$1.6}} & \shortstack{83.4\\ {\tiny$\pm$0.9}} & \shortstack{97.0\\ {\tiny$\pm$0.0}} & \shortstack{76.0\\ {\tiny$\pm$3.7}} & \shortstack{97.7\\ {\tiny$\pm$0.5}} \\
All & \shortstack{88.2\\ {\tiny$\pm$0.9}} & \shortstack{63.6\\ {\tiny$\pm$2.6}} & \shortstack{87.2\\ {\tiny$\pm$0.9}} & \shortstack{96.0\\ {\tiny$\pm$0.0}} & \shortstack{63.2\\ {\tiny$\pm$3.9}} & \shortstack{\textbf{95.3}\\ {\tiny$\pm$0.6}} & \shortstack{97.8\\ {\tiny$\pm$0.0}} & \shortstack{57.0\\ {\tiny$\pm$1.8}} & \shortstack{97.0\\ {\tiny$\pm$0.4}} & \shortstack{95.8\\ {\tiny$\pm$0.4}} & \shortstack{\textbf{69.1}\\ {\tiny$\pm$2.2}} & \shortstack{95.7\\ {\tiny$\pm$0.3}} & \shortstack{84.2\\ {\tiny$\pm$1.9}} & \shortstack{76.2\\ {\tiny$\pm$2.2}} & \shortstack{\textbf{89.3}\\ {\tiny$\pm$1.2}} & \shortstack{97.0\\ {\tiny$\pm$0.0}} & \shortstack{79.8\\ {\tiny$\pm$3.7}} & \shortstack{97.9\\ {\tiny$\pm$0.5}} \\
\bottomrule
\end{tabular}%
}
\end{table*}


\begin{table*}[t]
\centering
\caption{Feature-group ablation --- \textbf{Cross-lingual (CL)}. Detector performance (F1, AUC-ROC, PR-AUC in \%, mean over 6 seeds with $\pm$std) when trained on each feature-group configuration, for every setting and model in this regime. Best configuration per (setting, model, metric) in \textbf{bold}.}
\label{tab:ablation_CL}
\resizebox{\linewidth}{!}{%
\begin{tabular}{@{}l ccc ccc ccc ccc ccc ccc@{}}
\toprule
\textbf{Config} & \multicolumn{3}{c}{\textbf{Allam}} & \multicolumn{3}{c}{\textbf{Silma}} & \multicolumn{3}{c}{\textbf{Phi4-mini}} & \multicolumn{3}{c}{\textbf{Ministral}} & \multicolumn{3}{c}{\textbf{Qwen2.5}} & \multicolumn{3}{c}{\textbf{Aya}} \\
\cmidrule(lr){2-4} \cmidrule(lr){5-7} \cmidrule(lr){8-10} \cmidrule(lr){11-13} \cmidrule(lr){14-16} \cmidrule(lr){17-19}
 & \textbf{F1} & \textbf{AUC} & \textbf{PR} & \textbf{F1} & \textbf{AUC} & \textbf{PR} & \textbf{F1} & \textbf{AUC} & \textbf{PR} & \textbf{F1} & \textbf{AUC} & \textbf{PR} & \textbf{F1} & \textbf{AUC} & \textbf{PR} & \textbf{F1} & \textbf{AUC} & \textbf{PR} \\
\midrule
\multicolumn{19}{@{}l}{\cellcolor{gray!12}\textbf{En$\to$Ar}}\\
Hidden & \shortstack{83.4\\ {\tiny$\pm$0.9}} & \shortstack{\textbf{56.5}\\ {\tiny$\pm$3.6}} & \shortstack{\textbf{77.3}\\ {\tiny$\pm$1.7}} & \shortstack{73.0\\ {\tiny$\pm$1.6}} & \shortstack{32.4\\ {\tiny$\pm$3.7}} & \shortstack{53.3\\ {\tiny$\pm$2.1}} & \shortstack{81.3\\ {\tiny$\pm$3.9}} & \shortstack{68.0\\ {\tiny$\pm$2.2}} & \shortstack{88.6\\ {\tiny$\pm$0.4}} & \shortstack{74.3\\ {\tiny$\pm$3.0}} & \shortstack{60.4\\ {\tiny$\pm$3.1}} & \shortstack{75.2\\ {\tiny$\pm$2.5}} & \shortstack{41.1\\ {\tiny$\pm$6.3}} & \shortstack{62.6\\ {\tiny$\pm$2.1}} & \shortstack{45.4\\ {\tiny$\pm$2.3}} & \shortstack{84.1\\ {\tiny$\pm$1.2}} & \shortstack{66.2\\ {\tiny$\pm$1.2}} & \shortstack{86.1\\ {\tiny$\pm$0.8}} \\
Attention & \shortstack{\textbf{83.7}\\ {\tiny$\pm$0.7}} & \shortstack{47.4\\ {\tiny$\pm$2.4}} & \shortstack{71.9\\ {\tiny$\pm$1.6}} & \shortstack{79.1\\ {\tiny$\pm$0.6}} & \shortstack{70.7\\ {\tiny$\pm$2.9}} & \shortstack{79.1\\ {\tiny$\pm$2.4}} & \shortstack{86.3\\ {\tiny$\pm$0.7}} & \shortstack{73.6\\ {\tiny$\pm$2.9}} & \shortstack{89.3\\ {\tiny$\pm$1.4}} & \shortstack{\textbf{77.7}\\ {\tiny$\pm$1.6}} & \shortstack{60.4\\ {\tiny$\pm$1.6}} & \shortstack{76.6\\ {\tiny$\pm$1.6}} & \shortstack{45.2\\ {\tiny$\pm$5.4}} & \shortstack{67.2\\ {\tiny$\pm$1.8}} & \shortstack{53.2\\ {\tiny$\pm$3.3}} & \shortstack{84.5\\ {\tiny$\pm$0.8}} & \shortstack{64.6\\ {\tiny$\pm$3.6}} & \shortstack{85.6\\ {\tiny$\pm$2.2}} \\
Probability & \shortstack{81.5\\ {\tiny$\pm$2.0}} & \shortstack{45.0\\ {\tiny$\pm$1.5}} & \shortstack{74.8\\ {\tiny$\pm$1.3}} & \shortstack{76.9\\ {\tiny$\pm$2.5}} & \shortstack{65.2\\ {\tiny$\pm$1.6}} & \shortstack{75.0\\ {\tiny$\pm$2.7}} & \shortstack{\textbf{86.8}\\ {\tiny$\pm$0.1}} & \shortstack{67.0\\ {\tiny$\pm$1.6}} & \shortstack{87.8\\ {\tiny$\pm$0.7}} & \shortstack{77.2\\ {\tiny$\pm$2.0}} & \shortstack{53.3\\ {\tiny$\pm$1.7}} & \shortstack{71.1\\ {\tiny$\pm$1.0}} & \shortstack{28.3\\ {\tiny$\pm$19.3}} & \shortstack{53.5\\ {\tiny$\pm$1.5}} & \shortstack{35.8\\ {\tiny$\pm$1.6}} & \shortstack{\textbf{87.4}\\ {\tiny$\pm$0.7}} & \shortstack{59.7\\ {\tiny$\pm$4.0}} & \shortstack{81.2\\ {\tiny$\pm$2.0}} \\
Hidden+Attn & \shortstack{81.2\\ {\tiny$\pm$0.9}} & \shortstack{54.2\\ {\tiny$\pm$2.3}} & \shortstack{74.4\\ {\tiny$\pm$1.3}} & \shortstack{75.8\\ {\tiny$\pm$1.8}} & \shortstack{38.9\\ {\tiny$\pm$5.1}} & \shortstack{57.4\\ {\tiny$\pm$3.1}} & \shortstack{83.6\\ {\tiny$\pm$1.6}} & \shortstack{76.2\\ {\tiny$\pm$2.2}} & \shortstack{91.7\\ {\tiny$\pm$1.1}} & \shortstack{74.2\\ {\tiny$\pm$1.6}} & \shortstack{\textbf{66.1}\\ {\tiny$\pm$1.7}} & \shortstack{\textbf{79.4}\\ {\tiny$\pm$1.1}} & \shortstack{\textbf{53.8}\\ {\tiny$\pm$3.9}} & \shortstack{\textbf{70.0}\\ {\tiny$\pm$1.3}} & \shortstack{\textbf{54.8}\\ {\tiny$\pm$2.5}} & \shortstack{84.4\\ {\tiny$\pm$1.0}} & \shortstack{68.8\\ {\tiny$\pm$2.4}} & \shortstack{88.1\\ {\tiny$\pm$0.9}} \\
Hidden+Prob & \shortstack{83.0\\ {\tiny$\pm$1.1}} & \shortstack{52.4\\ {\tiny$\pm$1.9}} & \shortstack{75.6\\ {\tiny$\pm$2.6}} & \shortstack{74.7\\ {\tiny$\pm$2.4}} & \shortstack{37.6\\ {\tiny$\pm$3.2}} & \shortstack{56.6\\ {\tiny$\pm$2.4}} & \shortstack{83.0\\ {\tiny$\pm$1.5}} & \shortstack{72.2\\ {\tiny$\pm$1.9}} & \shortstack{90.7\\ {\tiny$\pm$0.5}} & \shortstack{75.1\\ {\tiny$\pm$3.2}} & \shortstack{59.7\\ {\tiny$\pm$4.9}} & \shortstack{73.6\\ {\tiny$\pm$3.8}} & \shortstack{45.8\\ {\tiny$\pm$5.1}} & \shortstack{64.2\\ {\tiny$\pm$1.7}} & \shortstack{47.6\\ {\tiny$\pm$1.5}} & \shortstack{86.3\\ {\tiny$\pm$0.6}} & \shortstack{68.7\\ {\tiny$\pm$1.7}} & \shortstack{87.5\\ {\tiny$\pm$0.9}} \\
Attn+Prob & \shortstack{80.6\\ {\tiny$\pm$2.0}} & \shortstack{46.3\\ {\tiny$\pm$1.8}} & \shortstack{73.3\\ {\tiny$\pm$1.4}} & \shortstack{\textbf{80.1}\\ {\tiny$\pm$1.1}} & \shortstack{\textbf{75.3}\\ {\tiny$\pm$1.7}} & \shortstack{\textbf{83.6}\\ {\tiny$\pm$1.9}} & \shortstack{86.5\\ {\tiny$\pm$0.7}} & \shortstack{77.8\\ {\tiny$\pm$1.7}} & \shortstack{92.1\\ {\tiny$\pm$0.9}} & \shortstack{75.8\\ {\tiny$\pm$1.2}} & \shortstack{59.1\\ {\tiny$\pm$2.4}} & \shortstack{76.0\\ {\tiny$\pm$2.4}} & \shortstack{35.0\\ {\tiny$\pm$11.8}} & \shortstack{61.9\\ {\tiny$\pm$2.0}} & \shortstack{49.2\\ {\tiny$\pm$2.1}} & \shortstack{86.7\\ {\tiny$\pm$0.8}} & \shortstack{\textbf{73.4}\\ {\tiny$\pm$2.7}} & \shortstack{\textbf{90.1}\\ {\tiny$\pm$1.2}} \\
All & \shortstack{81.0\\ {\tiny$\pm$1.8}} & \shortstack{50.7\\ {\tiny$\pm$2.5}} & \shortstack{73.3\\ {\tiny$\pm$1.7}} & \shortstack{76.9\\ {\tiny$\pm$1.3}} & \shortstack{43.3\\ {\tiny$\pm$5.0}} & \shortstack{61.4\\ {\tiny$\pm$2.9}} & \shortstack{84.2\\ {\tiny$\pm$1.3}} & \shortstack{\textbf{78.0}\\ {\tiny$\pm$1.0}} & \shortstack{\textbf{92.8}\\ {\tiny$\pm$0.4}} & \shortstack{75.8\\ {\tiny$\pm$2.2}} & \shortstack{63.4\\ {\tiny$\pm$3.4}} & \shortstack{76.5\\ {\tiny$\pm$3.4}} & \shortstack{51.7\\ {\tiny$\pm$2.9}} & \shortstack{66.7\\ {\tiny$\pm$2.4}} & \shortstack{53.4\\ {\tiny$\pm$3.4}} & \shortstack{86.5\\ {\tiny$\pm$0.5}} & \shortstack{73.4\\ {\tiny$\pm$1.4}} & \shortstack{89.7\\ {\tiny$\pm$0.6}} \\
\addlinespace[2pt]\midrule
\multicolumn{19}{@{}l}{\cellcolor{gray!12}\textbf{Ar$\to$En}}\\
Hidden & \shortstack{83.7\\ {\tiny$\pm$0.1}} & \shortstack{51.3\\ {\tiny$\pm$3.0}} & \shortstack{72.3\\ {\tiny$\pm$1.7}} & \shortstack{76.4\\ {\tiny$\pm$0.8}} & \shortstack{42.6\\ {\tiny$\pm$7.9}} & \shortstack{57.9\\ {\tiny$\pm$4.8}} & \shortstack{\textbf{79.0}\\ {\tiny$\pm$0.0}} & \shortstack{60.1\\ {\tiny$\pm$1.5}} & \shortstack{73.8\\ {\tiny$\pm$0.8}} & \shortstack{78.9\\ {\tiny$\pm$1.7}} & \shortstack{60.8\\ {\tiny$\pm$2.4}} & \shortstack{78.3\\ {\tiny$\pm$0.8}} & \shortstack{1.8\\ {\tiny$\pm$2.8}} & \shortstack{59.6\\ {\tiny$\pm$1.5}} & \shortstack{55.8\\ {\tiny$\pm$1.4}} & \shortstack{79.4\\ {\tiny$\pm$0.0}} & \shortstack{46.8\\ {\tiny$\pm$10.9}} & \shortstack{67.5\\ {\tiny$\pm$5.8}} \\
Attention & \shortstack{\textbf{83.7}\\ {\tiny$\pm$0.0}} & \shortstack{57.6\\ {\tiny$\pm$4.3}} & \shortstack{78.0\\ {\tiny$\pm$2.5}} & \shortstack{75.5\\ {\tiny$\pm$1.5}} & \shortstack{63.8\\ {\tiny$\pm$2.4}} & \shortstack{74.0\\ {\tiny$\pm$1.2}} & \shortstack{79.0\\ {\tiny$\pm$0.0}} & \shortstack{63.9\\ {\tiny$\pm$1.9}} & \shortstack{78.4\\ {\tiny$\pm$1.0}} & \shortstack{\textbf{82.4}\\ {\tiny$\pm$0.6}} & \shortstack{57.4\\ {\tiny$\pm$3.4}} & \shortstack{76.0\\ {\tiny$\pm$2.4}} & \shortstack{9.2\\ {\tiny$\pm$2.7}} & \shortstack{54.4\\ {\tiny$\pm$2.1}} & \shortstack{54.6\\ {\tiny$\pm$2.4}} & \shortstack{79.4\\ {\tiny$\pm$0.0}} & \shortstack{52.2\\ {\tiny$\pm$6.8}} & \shortstack{67.7\\ {\tiny$\pm$3.4}} \\
Probability & \shortstack{83.7\\ {\tiny$\pm$0.0}} & \shortstack{\textbf{58.3}\\ {\tiny$\pm$2.6}} & \shortstack{77.6\\ {\tiny$\pm$2.2}} & \shortstack{\textbf{79.0}\\ {\tiny$\pm$1.1}} & \shortstack{64.1\\ {\tiny$\pm$0.5}} & \shortstack{74.1\\ {\tiny$\pm$0.7}} & \shortstack{79.0\\ {\tiny$\pm$0.0}} & \shortstack{60.4\\ {\tiny$\pm$1.6}} & \shortstack{74.3\\ {\tiny$\pm$0.9}} & \shortstack{80.5\\ {\tiny$\pm$4.4}} & \shortstack{63.3\\ {\tiny$\pm$1.8}} & \shortstack{80.9\\ {\tiny$\pm$0.6}} & \shortstack{13.4\\ {\tiny$\pm$10.4}} & \shortstack{53.2\\ {\tiny$\pm$0.4}} & \shortstack{53.9\\ {\tiny$\pm$0.8}} & \shortstack{79.4\\ {\tiny$\pm$0.0}} & \shortstack{50.5\\ {\tiny$\pm$2.4}} & \shortstack{68.4\\ {\tiny$\pm$1.3}} \\
Hidden+Attn & \shortstack{83.7\\ {\tiny$\pm$0.1}} & \shortstack{52.6\\ {\tiny$\pm$3.8}} & \shortstack{73.4\\ {\tiny$\pm$1.4}} & \shortstack{76.1\\ {\tiny$\pm$1.0}} & \shortstack{43.0\\ {\tiny$\pm$6.8}} & \shortstack{57.1\\ {\tiny$\pm$3.9}} & \shortstack{79.0\\ {\tiny$\pm$0.0}} & \shortstack{64.0\\ {\tiny$\pm$1.8}} & \shortstack{78.0\\ {\tiny$\pm$2.1}} & \shortstack{78.4\\ {\tiny$\pm$2.0}} & \shortstack{62.6\\ {\tiny$\pm$1.8}} & \shortstack{79.7\\ {\tiny$\pm$0.9}} & \shortstack{8.7\\ {\tiny$\pm$3.4}} & \shortstack{\textbf{60.1}\\ {\tiny$\pm$2.2}} & \shortstack{56.0\\ {\tiny$\pm$2.1}} & \shortstack{79.4\\ {\tiny$\pm$0.0}} & \shortstack{55.1\\ {\tiny$\pm$6.2}} & \shortstack{73.0\\ {\tiny$\pm$3.1}} \\
Hidden+Prob & \shortstack{83.6\\ {\tiny$\pm$0.3}} & \shortstack{57.1\\ {\tiny$\pm$4.0}} & \shortstack{74.6\\ {\tiny$\pm$2.5}} & \shortstack{75.6\\ {\tiny$\pm$0.9}} & \shortstack{40.2\\ {\tiny$\pm$5.9}} & \shortstack{57.4\\ {\tiny$\pm$4.3}} & \shortstack{79.0\\ {\tiny$\pm$0.0}} & \shortstack{62.8\\ {\tiny$\pm$1.4}} & \shortstack{75.9\\ {\tiny$\pm$0.7}} & \shortstack{76.0\\ {\tiny$\pm$2.7}} & \shortstack{64.7\\ {\tiny$\pm$2.0}} & \shortstack{79.6\\ {\tiny$\pm$0.9}} & \shortstack{3.1\\ {\tiny$\pm$2.9}} & \shortstack{59.0\\ {\tiny$\pm$1.6}} & \shortstack{54.4\\ {\tiny$\pm$1.6}} & \shortstack{\textbf{79.4}\\ {\tiny$\pm$0.1}} & \shortstack{\textbf{59.3}\\ {\tiny$\pm$5.4}} & \shortstack{\textbf{76.3}\\ {\tiny$\pm$4.2}} \\
Attn+Prob & \shortstack{83.7\\ {\tiny$\pm$0.1}} & \shortstack{58.2\\ {\tiny$\pm$1.2}} & \shortstack{\textbf{78.0}\\ {\tiny$\pm$0.6}} & \shortstack{77.9\\ {\tiny$\pm$2.0}} & \shortstack{\textbf{67.0}\\ {\tiny$\pm$3.0}} & \shortstack{\textbf{76.2}\\ {\tiny$\pm$1.5}} & \shortstack{78.9\\ {\tiny$\pm$0.2}} & \shortstack{\textbf{68.1}\\ {\tiny$\pm$1.9}} & \shortstack{\textbf{81.0}\\ {\tiny$\pm$1.9}} & \shortstack{80.8\\ {\tiny$\pm$3.2}} & \shortstack{\textbf{66.8}\\ {\tiny$\pm$3.9}} & \shortstack{\textbf{81.8}\\ {\tiny$\pm$1.7}} & \shortstack{\textbf{21.8}\\ {\tiny$\pm$7.3}} & \shortstack{57.2\\ {\tiny$\pm$1.3}} & \shortstack{\textbf{56.6}\\ {\tiny$\pm$0.9}} & \shortstack{79.2\\ {\tiny$\pm$0.2}} & \shortstack{53.5\\ {\tiny$\pm$3.6}} & \shortstack{70.6\\ {\tiny$\pm$3.2}} \\
All & \shortstack{83.4\\ {\tiny$\pm$0.6}} & \shortstack{57.3\\ {\tiny$\pm$1.7}} & \shortstack{75.5\\ {\tiny$\pm$1.5}} & \shortstack{76.1\\ {\tiny$\pm$0.9}} & \shortstack{43.9\\ {\tiny$\pm$4.5}} & \shortstack{57.8\\ {\tiny$\pm$3.4}} & \shortstack{79.0\\ {\tiny$\pm$0.0}} & \shortstack{66.0\\ {\tiny$\pm$1.7}} & \shortstack{78.8\\ {\tiny$\pm$1.9}} & \shortstack{74.0\\ {\tiny$\pm$2.5}} & \shortstack{66.4\\ {\tiny$\pm$0.8}} & \shortstack{81.3\\ {\tiny$\pm$0.5}} & \shortstack{13.9\\ {\tiny$\pm$8.1}} & \shortstack{59.7\\ {\tiny$\pm$1.7}} & \shortstack{55.2\\ {\tiny$\pm$0.6}} & \shortstack{79.1\\ {\tiny$\pm$0.4}} & \shortstack{57.3\\ {\tiny$\pm$3.1}} & \shortstack{75.3\\ {\tiny$\pm$1.8}} \\
\bottomrule
\end{tabular}%
}
\end{table*}


\begin{table*}[t]
\centering
\caption{Feature-group ablation --- \textbf{Cross-domain (CD)}. Detector performance (F1, AUC-ROC, PR-AUC in \%, mean over 6 seeds with $\pm$std) when trained on each feature-group configuration, for every setting and model in this regime. Best configuration per (setting, model, metric) in \textbf{bold}.}
\label{tab:ablation_CD}
\resizebox{\linewidth}{!}{%
\begin{tabular}{@{}l ccc ccc ccc ccc ccc ccc@{}}
\toprule
\textbf{Config} & \multicolumn{3}{c}{\textbf{Allam}} & \multicolumn{3}{c}{\textbf{Silma}} & \multicolumn{3}{c}{\textbf{Phi4-mini}} & \multicolumn{3}{c}{\textbf{Ministral}} & \multicolumn{3}{c}{\textbf{Qwen2.5}} & \multicolumn{3}{c}{\textbf{Aya}} \\
\cmidrule(lr){2-4} \cmidrule(lr){5-7} \cmidrule(lr){8-10} \cmidrule(lr){11-13} \cmidrule(lr){14-16} \cmidrule(lr){17-19}
 & \textbf{F1} & \textbf{AUC} & \textbf{PR} & \textbf{F1} & \textbf{AUC} & \textbf{PR} & \textbf{F1} & \textbf{AUC} & \textbf{PR} & \textbf{F1} & \textbf{AUC} & \textbf{PR} & \textbf{F1} & \textbf{AUC} & \textbf{PR} & \textbf{F1} & \textbf{AUC} & \textbf{PR} \\
\midrule
\multicolumn{19}{@{}l}{\cellcolor{gray!12}\textbf{Ar$\to$HS}}\\
Hidden & \shortstack{88.9\\ {\tiny$\pm$0.1}} & \shortstack{49.6\\ {\tiny$\pm$2.0}} & \shortstack{80.1\\ {\tiny$\pm$1.1}} & \shortstack{89.1\\ {\tiny$\pm$1.4}} & \shortstack{56.7\\ {\tiny$\pm$3.8}} & \shortstack{94.1\\ {\tiny$\pm$1.1}} & \shortstack{\textbf{97.8}\\ {\tiny$\pm$0.0}} & \shortstack{71.4\\ {\tiny$\pm$3.2}} & \shortstack{98.4\\ {\tiny$\pm$0.2}} & \shortstack{84.8\\ {\tiny$\pm$0.9}} & \shortstack{68.3\\ {\tiny$\pm$1.7}} & \shortstack{94.7\\ {\tiny$\pm$0.7}} & \shortstack{29.9\\ {\tiny$\pm$5.3}} & \shortstack{63.4\\ {\tiny$\pm$1.4}} & \shortstack{79.8\\ {\tiny$\pm$1.0}} & \shortstack{97.0\\ {\tiny$\pm$0.0}} & \shortstack{78.0\\ {\tiny$\pm$1.8}} & \shortstack{97.5\\ {\tiny$\pm$0.2}} \\
Attention & \shortstack{88.9\\ {\tiny$\pm$0.0}} & \shortstack{\textbf{58.5}\\ {\tiny$\pm$3.3}} & \shortstack{\textbf{85.2}\\ {\tiny$\pm$1.4}} & \shortstack{93.5\\ {\tiny$\pm$1.5}} & \shortstack{65.3\\ {\tiny$\pm$3.1}} & \shortstack{94.5\\ {\tiny$\pm$0.7}} & \shortstack{97.8\\ {\tiny$\pm$0.0}} & \shortstack{46.9\\ {\tiny$\pm$3.1}} & \shortstack{95.5\\ {\tiny$\pm$0.5}} & \shortstack{\textbf{95.4}\\ {\tiny$\pm$0.5}} & \shortstack{59.8\\ {\tiny$\pm$3.6}} & \shortstack{93.8\\ {\tiny$\pm$0.7}} & \shortstack{11.3\\ {\tiny$\pm$3.0}} & \shortstack{66.1\\ {\tiny$\pm$1.7}} & \shortstack{84.8\\ {\tiny$\pm$1.7}} & \shortstack{97.0\\ {\tiny$\pm$0.0}} & \shortstack{61.7\\ {\tiny$\pm$4.3}} & \shortstack{96.5\\ {\tiny$\pm$0.6}} \\
Probability & \shortstack{88.9\\ {\tiny$\pm$0.0}} & \shortstack{52.6\\ {\tiny$\pm$2.7}} & \shortstack{82.7\\ {\tiny$\pm$0.9}} & \shortstack{\textbf{94.0}\\ {\tiny$\pm$1.2}} & \shortstack{63.8\\ {\tiny$\pm$1.8}} & \shortstack{94.1\\ {\tiny$\pm$0.4}} & \shortstack{97.8\\ {\tiny$\pm$0.0}} & \shortstack{59.3\\ {\tiny$\pm$2.1}} & \shortstack{97.3\\ {\tiny$\pm$0.2}} & \shortstack{94.9\\ {\tiny$\pm$1.2}} & \shortstack{57.1\\ {\tiny$\pm$4.7}} & \shortstack{93.6\\ {\tiny$\pm$0.7}} & \shortstack{21.1\\ {\tiny$\pm$14.0}} & \shortstack{60.8\\ {\tiny$\pm$2.6}} & \shortstack{81.8\\ {\tiny$\pm$1.1}} & \shortstack{97.0\\ {\tiny$\pm$0.1}} & \shortstack{81.9\\ {\tiny$\pm$2.6}} & \shortstack{98.6\\ {\tiny$\pm$0.3}} \\
Hidden+Attn & \shortstack{88.9\\ {\tiny$\pm$0.1}} & \shortstack{52.4\\ {\tiny$\pm$1.6}} & \shortstack{82.4\\ {\tiny$\pm$1.0}} & \shortstack{90.9\\ {\tiny$\pm$1.6}} & \shortstack{63.8\\ {\tiny$\pm$3.1}} & \shortstack{95.1\\ {\tiny$\pm$0.8}} & \shortstack{97.8\\ {\tiny$\pm$0.0}} & \shortstack{63.0\\ {\tiny$\pm$5.5}} & \shortstack{97.6\\ {\tiny$\pm$0.4}} & \shortstack{86.3\\ {\tiny$\pm$1.1}} & \shortstack{68.0\\ {\tiny$\pm$1.6}} & \shortstack{94.8\\ {\tiny$\pm$0.5}} & \shortstack{\textbf{40.6}\\ {\tiny$\pm$4.3}} & \shortstack{70.3\\ {\tiny$\pm$1.9}} & \shortstack{\textbf{86.2}\\ {\tiny$\pm$0.9}} & \shortstack{97.1\\ {\tiny$\pm$0.1}} & \shortstack{74.0\\ {\tiny$\pm$2.9}} & \shortstack{97.6\\ {\tiny$\pm$0.2}} \\
Hidden+Prob & \shortstack{\textbf{89.0}\\ {\tiny$\pm$0.1}} & \shortstack{54.2\\ {\tiny$\pm$3.0}} & \shortstack{83.7\\ {\tiny$\pm$1.9}} & \shortstack{89.4\\ {\tiny$\pm$1.9}} & \shortstack{64.6\\ {\tiny$\pm$0.8}} & \shortstack{95.5\\ {\tiny$\pm$0.2}} & \shortstack{97.8\\ {\tiny$\pm$0.0}} & \shortstack{71.2\\ {\tiny$\pm$1.0}} & \shortstack{\textbf{98.4}\\ {\tiny$\pm$0.1}} & \shortstack{84.9\\ {\tiny$\pm$2.5}} & \shortstack{70.3\\ {\tiny$\pm$3.7}} & \shortstack{95.1\\ {\tiny$\pm$0.7}} & \shortstack{35.4\\ {\tiny$\pm$7.4}} & \shortstack{67.6\\ {\tiny$\pm$1.8}} & \shortstack{83.7\\ {\tiny$\pm$1.4}} & \shortstack{\textbf{97.1}\\ {\tiny$\pm$0.2}} & \shortstack{85.1\\ {\tiny$\pm$0.7}} & \shortstack{98.4\\ {\tiny$\pm$0.1}} \\
Attn+Prob & \shortstack{88.9\\ {\tiny$\pm$0.0}} & \shortstack{54.0\\ {\tiny$\pm$4.7}} & \shortstack{83.2\\ {\tiny$\pm$1.9}} & \shortstack{91.0\\ {\tiny$\pm$3.0}} & \shortstack{\textbf{70.8}\\ {\tiny$\pm$1.5}} & \shortstack{\textbf{96.1}\\ {\tiny$\pm$0.3}} & \shortstack{97.8\\ {\tiny$\pm$0.0}} & \shortstack{64.6\\ {\tiny$\pm$2.8}} & \shortstack{97.5\\ {\tiny$\pm$0.3}} & \shortstack{92.4\\ {\tiny$\pm$2.6}} & \shortstack{68.1\\ {\tiny$\pm$5.2}} & \shortstack{94.8\\ {\tiny$\pm$0.9}} & \shortstack{22.8\\ {\tiny$\pm$4.6}} & \shortstack{67.1\\ {\tiny$\pm$1.0}} & \shortstack{84.2\\ {\tiny$\pm$0.6}} & \shortstack{96.7\\ {\tiny$\pm$0.8}} & \shortstack{86.4\\ {\tiny$\pm$1.2}} & \shortstack{\textbf{98.9}\\ {\tiny$\pm$0.1}} \\
All & \shortstack{88.3\\ {\tiny$\pm$1.7}} & \shortstack{54.5\\ {\tiny$\pm$1.6}} & \shortstack{84.0\\ {\tiny$\pm$0.8}} & \shortstack{89.8\\ {\tiny$\pm$1.6}} & \shortstack{65.7\\ {\tiny$\pm$1.1}} & \shortstack{95.5\\ {\tiny$\pm$0.2}} & \shortstack{97.7\\ {\tiny$\pm$0.1}} & \shortstack{\textbf{71.8}\\ {\tiny$\pm$2.7}} & \shortstack{98.4\\ {\tiny$\pm$0.2}} & \shortstack{85.2\\ {\tiny$\pm$1.8}} & \shortstack{\textbf{70.5}\\ {\tiny$\pm$1.9}} & \shortstack{\textbf{95.3}\\ {\tiny$\pm$0.4}} & \shortstack{34.3\\ {\tiny$\pm$5.8}} & \shortstack{\textbf{70.6}\\ {\tiny$\pm$1.4}} & \shortstack{86.0\\ {\tiny$\pm$0.7}} & \shortstack{96.8\\ {\tiny$\pm$0.9}} & \shortstack{\textbf{86.5}\\ {\tiny$\pm$1.7}} & \shortstack{98.6\\ {\tiny$\pm$0.1}} \\
\addlinespace[2pt]\midrule
\multicolumn{19}{@{}l}{\cellcolor{gray!12}\textbf{HS$\to$Ar}}\\
Hidden & \shortstack{84.4\\ {\tiny$\pm$0.1}} & \shortstack{47.6\\ {\tiny$\pm$3.8}} & \shortstack{71.6\\ {\tiny$\pm$2.3}} & \shortstack{78.5\\ {\tiny$\pm$0.2}} & \shortstack{66.3\\ {\tiny$\pm$2.9}} & \shortstack{77.7\\ {\tiny$\pm$1.9}} & \shortstack{\textbf{86.8}\\ {\tiny$\pm$0.0}} & \shortstack{\textbf{72.2}\\ {\tiny$\pm$1.5}} & \shortstack{\textbf{89.6}\\ {\tiny$\pm$0.7}} & \shortstack{80.5\\ {\tiny$\pm$0.0}} & \shortstack{72.7\\ {\tiny$\pm$1.7}} & \shortstack{84.4\\ {\tiny$\pm$0.8}} & \shortstack{52.0\\ {\tiny$\pm$1.4}} & \shortstack{65.7\\ {\tiny$\pm$1.6}} & \shortstack{46.2\\ {\tiny$\pm$2.0}} & \shortstack{\textbf{86.8}\\ {\tiny$\pm$0.0}} & \shortstack{64.1\\ {\tiny$\pm$4.8}} & \shortstack{85.2\\ {\tiny$\pm$2.0}} \\
Attention & \shortstack{\textbf{84.4}\\ {\tiny$\pm$0.0}} & \shortstack{49.6\\ {\tiny$\pm$2.1}} & \shortstack{\textbf{74.2}\\ {\tiny$\pm$0.8}} & \shortstack{78.6\\ {\tiny$\pm$0.0}} & \shortstack{60.3\\ {\tiny$\pm$4.7}} & \shortstack{71.9\\ {\tiny$\pm$2.8}} & \shortstack{86.8\\ {\tiny$\pm$0.0}} & \shortstack{55.9\\ {\tiny$\pm$1.1}} & \shortstack{79.7\\ {\tiny$\pm$0.6}} & \shortstack{80.5\\ {\tiny$\pm$0.0}} & \shortstack{56.7\\ {\tiny$\pm$1.6}} & \shortstack{72.9\\ {\tiny$\pm$1.5}} & \shortstack{50.5\\ {\tiny$\pm$1.0}} & \shortstack{69.9\\ {\tiny$\pm$1.5}} & \shortstack{54.4\\ {\tiny$\pm$1.7}} & \shortstack{86.8\\ {\tiny$\pm$0.0}} & \shortstack{51.2\\ {\tiny$\pm$3.1}} & \shortstack{78.4\\ {\tiny$\pm$1.5}} \\
Probability & \shortstack{82.2\\ {\tiny$\pm$1.6}} & \shortstack{40.8\\ {\tiny$\pm$2.2}} & \shortstack{70.5\\ {\tiny$\pm$1.5}} & \shortstack{\textbf{78.7}\\ {\tiny$\pm$0.1}} & \shortstack{70.9\\ {\tiny$\pm$0.3}} & \shortstack{81.0\\ {\tiny$\pm$0.7}} & \shortstack{86.8\\ {\tiny$\pm$0.0}} & \shortstack{62.9\\ {\tiny$\pm$5.7}} & \shortstack{85.2\\ {\tiny$\pm$3.9}} & \shortstack{80.5\\ {\tiny$\pm$0.0}} & \shortstack{53.9\\ {\tiny$\pm$2.5}} & \shortstack{70.0\\ {\tiny$\pm$1.4}} & \shortstack{49.1\\ {\tiny$\pm$0.4}} & \shortstack{56.1\\ {\tiny$\pm$0.7}} & \shortstack{38.6\\ {\tiny$\pm$0.9}} & \shortstack{86.8\\ {\tiny$\pm$0.0}} & \shortstack{66.8\\ {\tiny$\pm$3.8}} & \shortstack{87.3\\ {\tiny$\pm$1.7}} \\
Hidden+Attn & \shortstack{84.2\\ {\tiny$\pm$0.4}} & \shortstack{\textbf{50.2}\\ {\tiny$\pm$5.0}} & \shortstack{73.6\\ {\tiny$\pm$2.5}} & \shortstack{78.6\\ {\tiny$\pm$0.0}} & \shortstack{70.5\\ {\tiny$\pm$1.7}} & \shortstack{80.1\\ {\tiny$\pm$1.3}} & \shortstack{86.8\\ {\tiny$\pm$0.0}} & \shortstack{65.9\\ {\tiny$\pm$1.1}} & \shortstack{84.5\\ {\tiny$\pm$0.4}} & \shortstack{\textbf{80.7}\\ {\tiny$\pm$0.2}} & \shortstack{72.2\\ {\tiny$\pm$2.1}} & \shortstack{84.7\\ {\tiny$\pm$1.4}} & \shortstack{\textbf{54.4}\\ {\tiny$\pm$2.3}} & \shortstack{\textbf{70.4}\\ {\tiny$\pm$0.3}} & \shortstack{51.6\\ {\tiny$\pm$1.5}} & \shortstack{86.8\\ {\tiny$\pm$0.0}} & \shortstack{61.3\\ {\tiny$\pm$3.7}} & \shortstack{83.1\\ {\tiny$\pm$1.6}} \\
Hidden+Prob & \shortstack{81.7\\ {\tiny$\pm$1.9}} & \shortstack{41.9\\ {\tiny$\pm$2.4}} & \shortstack{71.8\\ {\tiny$\pm$1.7}} & \shortstack{78.5\\ {\tiny$\pm$0.3}} & \shortstack{69.4\\ {\tiny$\pm$1.0}} & \shortstack{80.4\\ {\tiny$\pm$1.4}} & \shortstack{86.8\\ {\tiny$\pm$0.0}} & \shortstack{70.9\\ {\tiny$\pm$2.2}} & \shortstack{89.2\\ {\tiny$\pm$1.2}} & \shortstack{80.5\\ {\tiny$\pm$0.3}} & \shortstack{73.9\\ {\tiny$\pm$2.2}} & \shortstack{84.9\\ {\tiny$\pm$1.2}} & \shortstack{51.3\\ {\tiny$\pm$0.8}} & \shortstack{67.4\\ {\tiny$\pm$0.6}} & \shortstack{54.5\\ {\tiny$\pm$1.6}} & \shortstack{86.8\\ {\tiny$\pm$0.0}} & \shortstack{\textbf{74.4}\\ {\tiny$\pm$1.2}} & \shortstack{\textbf{89.6}\\ {\tiny$\pm$0.7}} \\
Attn+Prob & \shortstack{82.3\\ {\tiny$\pm$1.4}} & \shortstack{45.4\\ {\tiny$\pm$2.3}} & \shortstack{72.0\\ {\tiny$\pm$1.1}} & \shortstack{78.6\\ {\tiny$\pm$0.0}} & \shortstack{\textbf{75.3}\\ {\tiny$\pm$2.2}} & \shortstack{81.9\\ {\tiny$\pm$2.3}} & \shortstack{86.8\\ {\tiny$\pm$0.0}} & \shortstack{66.4\\ {\tiny$\pm$6.2}} & \shortstack{85.0\\ {\tiny$\pm$3.5}} & \shortstack{80.5\\ {\tiny$\pm$0.1}} & \shortstack{62.5\\ {\tiny$\pm$4.0}} & \shortstack{77.9\\ {\tiny$\pm$3.2}} & \shortstack{50.4\\ {\tiny$\pm$1.2}} & \shortstack{64.1\\ {\tiny$\pm$1.2}} & \shortstack{45.8\\ {\tiny$\pm$2.4}} & \shortstack{86.8\\ {\tiny$\pm$0.0}} & \shortstack{68.4\\ {\tiny$\pm$2.9}} & \shortstack{87.5\\ {\tiny$\pm$1.1}} \\
All & \shortstack{81.4\\ {\tiny$\pm$1.4}} & \shortstack{44.4\\ {\tiny$\pm$1.8}} & \shortstack{71.5\\ {\tiny$\pm$0.9}} & \shortstack{78.6\\ {\tiny$\pm$0.0}} & \shortstack{74.4\\ {\tiny$\pm$1.7}} & \shortstack{\textbf{83.0}\\ {\tiny$\pm$1.8}} & \shortstack{86.8\\ {\tiny$\pm$0.0}} & \shortstack{70.1\\ {\tiny$\pm$1.6}} & \shortstack{88.4\\ {\tiny$\pm$1.4}} & \shortstack{80.6\\ {\tiny$\pm$0.2}} & \shortstack{\textbf{74.1}\\ {\tiny$\pm$2.1}} & \shortstack{\textbf{85.9}\\ {\tiny$\pm$1.1}} & \shortstack{53.9\\ {\tiny$\pm$1.2}} & \shortstack{70.3\\ {\tiny$\pm$1.3}} & \shortstack{\textbf{55.0}\\ {\tiny$\pm$1.8}} & \shortstack{86.8\\ {\tiny$\pm$0.0}} & \shortstack{72.0\\ {\tiny$\pm$1.9}} & \shortstack{88.1\\ {\tiny$\pm$0.9}} \\
\bottomrule
\end{tabular}%
}
\end{table*}


\begin{table*}[t]
\centering
\caption{Feature-group ablation --- \textbf{Cross-lingual + cross-domain (CL-CD)}. Detector performance (F1, AUC-ROC, PR-AUC in \%, mean over 6 seeds with $\pm$std) when trained on each feature-group configuration, for every setting and model in this regime. Best configuration per (setting, model, metric) in \textbf{bold}.}
\label{tab:ablation_CLCD}
\resizebox{\linewidth}{!}{%
\begin{tabular}{@{}l ccc ccc ccc ccc ccc ccc@{}}
\toprule
\textbf{Config} & \multicolumn{3}{c}{\textbf{Allam}} & \multicolumn{3}{c}{\textbf{Silma}} & \multicolumn{3}{c}{\textbf{Phi4-mini}} & \multicolumn{3}{c}{\textbf{Ministral}} & \multicolumn{3}{c}{\textbf{Qwen2.5}} & \multicolumn{3}{c}{\textbf{Aya}} \\
\cmidrule(lr){2-4} \cmidrule(lr){5-7} \cmidrule(lr){8-10} \cmidrule(lr){11-13} \cmidrule(lr){14-16} \cmidrule(lr){17-19}
 & \textbf{F1} & \textbf{AUC} & \textbf{PR} & \textbf{F1} & \textbf{AUC} & \textbf{PR} & \textbf{F1} & \textbf{AUC} & \textbf{PR} & \textbf{F1} & \textbf{AUC} & \textbf{PR} & \textbf{F1} & \textbf{AUC} & \textbf{PR} & \textbf{F1} & \textbf{AUC} & \textbf{PR} \\
\midrule
\multicolumn{19}{@{}l}{\cellcolor{gray!12}\textbf{En$\to$HS}}\\
Hidden & \shortstack{87.8\\ {\tiny$\pm$0.9}} & \shortstack{55.1\\ {\tiny$\pm$3.9}} & \shortstack{82.9\\ {\tiny$\pm$2.1}} & \shortstack{92.7\\ {\tiny$\pm$1.3}} & \shortstack{45.1\\ {\tiny$\pm$3.1}} & \shortstack{91.4\\ {\tiny$\pm$0.8}} & \shortstack{87.0\\ {\tiny$\pm$5.0}} & \shortstack{58.4\\ {\tiny$\pm$3.1}} & \shortstack{97.4\\ {\tiny$\pm$0.3}} & \shortstack{82.2\\ {\tiny$\pm$3.8}} & \shortstack{62.4\\ {\tiny$\pm$4.9}} & \shortstack{94.9\\ {\tiny$\pm$0.9}} & \shortstack{50.2\\ {\tiny$\pm$9.8}} & \shortstack{64.4\\ {\tiny$\pm$2.2}} & \shortstack{79.4\\ {\tiny$\pm$1.8}} & \shortstack{92.3\\ {\tiny$\pm$1.3}} & \shortstack{63.5\\ {\tiny$\pm$4.5}} & \shortstack{96.6\\ {\tiny$\pm$0.4}} \\
Attention & \shortstack{88.3\\ {\tiny$\pm$0.6}} & \shortstack{\textbf{68.2}\\ {\tiny$\pm$1.4}} & \shortstack{\textbf{89.7}\\ {\tiny$\pm$0.4}} & \shortstack{88.5\\ {\tiny$\pm$2.1}} & \shortstack{46.3\\ {\tiny$\pm$7.2}} & \shortstack{91.7\\ {\tiny$\pm$1.5}} & \shortstack{95.4\\ {\tiny$\pm$1.5}} & \shortstack{58.7\\ {\tiny$\pm$5.4}} & \shortstack{96.0\\ {\tiny$\pm$0.6}} & \shortstack{\textbf{91.5}\\ {\tiny$\pm$2.7}} & \shortstack{50.3\\ {\tiny$\pm$4.0}} & \shortstack{92.3\\ {\tiny$\pm$0.8}} & \shortstack{50.8\\ {\tiny$\pm$5.5}} & \shortstack{\textbf{72.7}\\ {\tiny$\pm$1.5}} & \shortstack{\textbf{87.7}\\ {\tiny$\pm$0.7}} & \shortstack{93.0\\ {\tiny$\pm$1.0}} & \shortstack{71.2\\ {\tiny$\pm$5.2}} & \shortstack{97.7\\ {\tiny$\pm$0.5}} \\
Probability & \shortstack{87.8\\ {\tiny$\pm$1.6}} & \shortstack{57.3\\ {\tiny$\pm$1.2}} & \shortstack{85.0\\ {\tiny$\pm$0.6}} & \shortstack{87.6\\ {\tiny$\pm$3.8}} & \shortstack{53.9\\ {\tiny$\pm$3.7}} & \shortstack{92.9\\ {\tiny$\pm$0.9}} & \shortstack{\textbf{96.9}\\ {\tiny$\pm$1.7}} & \shortstack{50.8\\ {\tiny$\pm$3.1}} & \shortstack{96.2\\ {\tiny$\pm$0.5}} & \shortstack{89.0\\ {\tiny$\pm$2.5}} & \shortstack{52.4\\ {\tiny$\pm$3.6}} & \shortstack{92.8\\ {\tiny$\pm$0.7}} & \shortstack{42.3\\ {\tiny$\pm$27.8}} & \shortstack{54.9\\ {\tiny$\pm$6.3}} & \shortstack{78.6\\ {\tiny$\pm$2.7}} & \shortstack{94.0\\ {\tiny$\pm$1.3}} & \shortstack{43.8\\ {\tiny$\pm$4.2}} & \shortstack{92.9\\ {\tiny$\pm$0.7}} \\
Hidden+Attn & \shortstack{87.1\\ {\tiny$\pm$1.0}} & \shortstack{60.4\\ {\tiny$\pm$2.6}} & \shortstack{84.9\\ {\tiny$\pm$1.2}} & \shortstack{\textbf{93.3}\\ {\tiny$\pm$1.6}} & \shortstack{54.6\\ {\tiny$\pm$7.3}} & \shortstack{94.0\\ {\tiny$\pm$1.1}} & \shortstack{88.7\\ {\tiny$\pm$3.6}} & \shortstack{\textbf{63.8}\\ {\tiny$\pm$5.2}} & \shortstack{97.7\\ {\tiny$\pm$0.5}} & \shortstack{81.6\\ {\tiny$\pm$4.3}} & \shortstack{65.5\\ {\tiny$\pm$1.7}} & \shortstack{95.5\\ {\tiny$\pm$0.5}} & \shortstack{66.2\\ {\tiny$\pm$6.5}} & \shortstack{69.1\\ {\tiny$\pm$2.9}} & \shortstack{83.9\\ {\tiny$\pm$1.7}} & \shortstack{91.2\\ {\tiny$\pm$1.5}} & \shortstack{66.3\\ {\tiny$\pm$3.0}} & \shortstack{97.2\\ {\tiny$\pm$0.3}} \\
Hidden+Prob & \shortstack{\textbf{88.4}\\ {\tiny$\pm$0.4}} & \shortstack{60.0\\ {\tiny$\pm$1.3}} & \shortstack{85.8\\ {\tiny$\pm$1.1}} & \shortstack{92.8\\ {\tiny$\pm$2.0}} & \shortstack{54.2\\ {\tiny$\pm$3.6}} & \shortstack{93.8\\ {\tiny$\pm$0.6}} & \shortstack{87.7\\ {\tiny$\pm$3.8}} & \shortstack{60.1\\ {\tiny$\pm$1.0}} & \shortstack{97.5\\ {\tiny$\pm$0.2}} & \shortstack{82.3\\ {\tiny$\pm$5.6}} & \shortstack{64.2\\ {\tiny$\pm$5.8}} & \shortstack{95.2\\ {\tiny$\pm$1.0}} & \shortstack{60.5\\ {\tiny$\pm$5.7}} & \shortstack{63.3\\ {\tiny$\pm$4.0}} & \shortstack{80.5\\ {\tiny$\pm$3.6}} & \shortstack{\textbf{94.2}\\ {\tiny$\pm$0.8}} & \shortstack{69.9\\ {\tiny$\pm$3.7}} & \shortstack{97.3\\ {\tiny$\pm$0.6}} \\
Attn+Prob & \shortstack{85.6\\ {\tiny$\pm$1.5}} & \shortstack{60.2\\ {\tiny$\pm$1.4}} & \shortstack{86.4\\ {\tiny$\pm$0.7}} & \shortstack{86.0\\ {\tiny$\pm$3.2}} & \shortstack{56.6\\ {\tiny$\pm$6.4}} & \shortstack{94.2\\ {\tiny$\pm$1.1}} & \shortstack{94.4\\ {\tiny$\pm$1.7}} & \shortstack{56.1\\ {\tiny$\pm$4.5}} & \shortstack{96.8\\ {\tiny$\pm$0.7}} & \shortstack{85.6\\ {\tiny$\pm$3.7}} & \shortstack{59.6\\ {\tiny$\pm$4.3}} & \shortstack{94.3\\ {\tiny$\pm$0.9}} & \shortstack{38.4\\ {\tiny$\pm$14.7}} & \shortstack{68.2\\ {\tiny$\pm$1.8}} & \shortstack{84.8\\ {\tiny$\pm$0.8}} & \shortstack{91.7\\ {\tiny$\pm$0.9}} & \shortstack{68.5\\ {\tiny$\pm$4.5}} & \shortstack{97.2\\ {\tiny$\pm$0.5}} \\
All & \shortstack{87.1\\ {\tiny$\pm$0.8}} & \shortstack{57.7\\ {\tiny$\pm$2.3}} & \shortstack{84.5\\ {\tiny$\pm$0.9}} & \shortstack{92.6\\ {\tiny$\pm$1.3}} & \shortstack{\textbf{57.8}\\ {\tiny$\pm$4.1}} & \shortstack{\textbf{94.7}\\ {\tiny$\pm$0.3}} & \shortstack{85.7\\ {\tiny$\pm$2.7}} & \shortstack{63.6\\ {\tiny$\pm$4.9}} & \shortstack{\textbf{97.8}\\ {\tiny$\pm$0.5}} & \shortstack{80.0\\ {\tiny$\pm$3.1}} & \shortstack{\textbf{67.1}\\ {\tiny$\pm$2.2}} & \shortstack{\textbf{95.7}\\ {\tiny$\pm$0.6}} & \shortstack{\textbf{68.5}\\ {\tiny$\pm$2.7}} & \shortstack{68.2\\ {\tiny$\pm$2.5}} & \shortstack{84.8\\ {\tiny$\pm$1.9}} & \shortstack{94.0\\ {\tiny$\pm$0.8}} & \shortstack{\textbf{75.6}\\ {\tiny$\pm$2.6}} & \shortstack{\textbf{98.0}\\ {\tiny$\pm$0.3}} \\
\addlinespace[2pt]\midrule
\multicolumn{19}{@{}l}{\cellcolor{gray!12}\textbf{HS$\to$En}}\\
Hidden & \shortstack{\textbf{83.8}\\ {\tiny$\pm$0.2}} & \shortstack{59.1\\ {\tiny$\pm$1.7}} & \shortstack{74.0\\ {\tiny$\pm$1.1}} & \shortstack{77.1\\ {\tiny$\pm$0.0}} & \shortstack{48.5\\ {\tiny$\pm$9.6}} & \shortstack{63.3\\ {\tiny$\pm$6.3}} & \shortstack{\textbf{79.0}\\ {\tiny$\pm$0.0}} & \shortstack{60.3\\ {\tiny$\pm$1.5}} & \shortstack{74.5\\ {\tiny$\pm$0.8}} & \shortstack{82.7\\ {\tiny$\pm$0.0}} & \shortstack{61.7\\ {\tiny$\pm$1.2}} & \shortstack{78.4\\ {\tiny$\pm$0.9}} & \shortstack{58.8\\ {\tiny$\pm$2.9}} & \shortstack{52.5\\ {\tiny$\pm$2.5}} & \shortstack{52.0\\ {\tiny$\pm$2.9}} & \shortstack{\textbf{79.4}\\ {\tiny$\pm$0.0}} & \shortstack{40.5\\ {\tiny$\pm$7.2}} & \shortstack{62.8\\ {\tiny$\pm$4.2}} \\
Attention & \shortstack{83.7\\ {\tiny$\pm$0.1}} & \shortstack{57.2\\ {\tiny$\pm$2.4}} & \shortstack{78.5\\ {\tiny$\pm$1.4}} & \shortstack{77.1\\ {\tiny$\pm$0.0}} & \shortstack{58.3\\ {\tiny$\pm$2.5}} & \shortstack{71.6\\ {\tiny$\pm$1.9}} & \shortstack{79.0\\ {\tiny$\pm$0.0}} & \shortstack{59.9\\ {\tiny$\pm$1.1}} & \shortstack{72.9\\ {\tiny$\pm$1.1}} & \shortstack{82.7\\ {\tiny$\pm$0.0}} & \shortstack{48.8\\ {\tiny$\pm$1.1}} & \shortstack{70.3\\ {\tiny$\pm$1.1}} & \shortstack{\textbf{63.8}\\ {\tiny$\pm$0.7}} & \shortstack{58.9\\ {\tiny$\pm$1.8}} & \shortstack{\textbf{58.4}\\ {\tiny$\pm$1.5}} & \shortstack{79.4\\ {\tiny$\pm$0.0}} & \shortstack{46.4\\ {\tiny$\pm$4.0}} & \shortstack{61.1\\ {\tiny$\pm$1.8}} \\
Probability & \shortstack{81.5\\ {\tiny$\pm$1.7}} & \shortstack{56.9\\ {\tiny$\pm$1.4}} & \shortstack{79.2\\ {\tiny$\pm$1.2}} & \shortstack{\textbf{77.1}\\ {\tiny$\pm$0.1}} & \shortstack{62.6\\ {\tiny$\pm$1.3}} & \shortstack{73.6\\ {\tiny$\pm$0.4}} & \shortstack{79.0\\ {\tiny$\pm$0.0}} & \shortstack{59.6\\ {\tiny$\pm$4.4}} & \shortstack{74.3\\ {\tiny$\pm$2.8}} & \shortstack{82.7\\ {\tiny$\pm$0.0}} & \shortstack{50.4\\ {\tiny$\pm$3.5}} & \shortstack{72.9\\ {\tiny$\pm$2.3}} & \shortstack{59.6\\ {\tiny$\pm$4.7}} & \shortstack{52.8\\ {\tiny$\pm$0.7}} & \shortstack{54.7\\ {\tiny$\pm$1.0}} & \shortstack{79.4\\ {\tiny$\pm$0.0}} & \shortstack{55.5\\ {\tiny$\pm$3.4}} & \shortstack{71.4\\ {\tiny$\pm$0.8}} \\
Hidden+Attn & \shortstack{83.8\\ {\tiny$\pm$0.1}} & \shortstack{59.8\\ {\tiny$\pm$3.3}} & \shortstack{78.1\\ {\tiny$\pm$2.0}} & \shortstack{77.1\\ {\tiny$\pm$0.0}} & \shortstack{51.8\\ {\tiny$\pm$5.1}} & \shortstack{67.4\\ {\tiny$\pm$3.8}} & \shortstack{79.0\\ {\tiny$\pm$0.0}} & \shortstack{60.2\\ {\tiny$\pm$1.8}} & \shortstack{74.9\\ {\tiny$\pm$0.7}} & \shortstack{82.7\\ {\tiny$\pm$0.0}} & \shortstack{58.7\\ {\tiny$\pm$2.8}} & \shortstack{77.5\\ {\tiny$\pm$1.9}} & \shortstack{61.0\\ {\tiny$\pm$2.3}} & \shortstack{58.6\\ {\tiny$\pm$1.9}} & \shortstack{54.5\\ {\tiny$\pm$1.0}} & \shortstack{79.4\\ {\tiny$\pm$0.0}} & \shortstack{48.6\\ {\tiny$\pm$4.6}} & \shortstack{63.7\\ {\tiny$\pm$2.5}} \\
Hidden+Prob & \shortstack{81.8\\ {\tiny$\pm$1.0}} & \shortstack{59.5\\ {\tiny$\pm$3.3}} & \shortstack{78.5\\ {\tiny$\pm$3.2}} & \shortstack{77.1\\ {\tiny$\pm$0.0}} & \shortstack{49.9\\ {\tiny$\pm$9.0}} & \shortstack{65.0\\ {\tiny$\pm$5.7}} & \shortstack{79.0\\ {\tiny$\pm$0.0}} & \shortstack{64.5\\ {\tiny$\pm$2.0}} & \shortstack{77.5\\ {\tiny$\pm$1.2}} & \shortstack{82.7\\ {\tiny$\pm$0.0}} & \shortstack{62.1\\ {\tiny$\pm$2.9}} & \shortstack{77.9\\ {\tiny$\pm$1.8}} & \shortstack{49.9\\ {\tiny$\pm$7.5}} & \shortstack{\textbf{59.4}\\ {\tiny$\pm$2.4}} & \shortstack{56.2\\ {\tiny$\pm$1.1}} & \shortstack{79.4\\ {\tiny$\pm$0.0}} & \shortstack{\textbf{59.2}\\ {\tiny$\pm$7.2}} & \shortstack{\textbf{74.0}\\ {\tiny$\pm$4.6}} \\
Attn+Prob & \shortstack{82.8\\ {\tiny$\pm$0.8}} & \shortstack{59.0\\ {\tiny$\pm$1.7}} & \shortstack{79.3\\ {\tiny$\pm$0.8}} & \shortstack{77.1\\ {\tiny$\pm$0.0}} & \shortstack{\textbf{64.3}\\ {\tiny$\pm$2.7}} & \shortstack{\textbf{74.9}\\ {\tiny$\pm$1.0}} & \shortstack{79.0\\ {\tiny$\pm$0.0}} & \shortstack{63.4\\ {\tiny$\pm$3.1}} & \shortstack{75.8\\ {\tiny$\pm$2.7}} & \shortstack{82.6\\ {\tiny$\pm$0.1}} & \shortstack{54.3\\ {\tiny$\pm$6.5}} & \shortstack{76.7\\ {\tiny$\pm$4.2}} & \shortstack{59.2\\ {\tiny$\pm$5.6}} & \shortstack{54.1\\ {\tiny$\pm$2.1}} & \shortstack{55.9\\ {\tiny$\pm$1.3}} & \shortstack{79.4\\ {\tiny$\pm$0.0}} & \shortstack{50.0\\ {\tiny$\pm$4.3}} & \shortstack{63.7\\ {\tiny$\pm$3.1}} \\
All & \shortstack{81.8\\ {\tiny$\pm$0.8}} & \shortstack{\textbf{61.5}\\ {\tiny$\pm$1.5}} & \shortstack{\textbf{79.4}\\ {\tiny$\pm$1.7}} & \shortstack{77.1\\ {\tiny$\pm$0.0}} & \shortstack{55.8\\ {\tiny$\pm$3.3}} & \shortstack{67.7\\ {\tiny$\pm$3.0}} & \shortstack{79.0\\ {\tiny$\pm$0.0}} & \shortstack{\textbf{64.6}\\ {\tiny$\pm$2.5}} & \shortstack{\textbf{78.5}\\ {\tiny$\pm$1.3}} & \shortstack{\textbf{82.7}\\ {\tiny$\pm$0.1}} & \shortstack{\textbf{62.5}\\ {\tiny$\pm$1.2}} & \shortstack{\textbf{79.5}\\ {\tiny$\pm$1.2}} & \shortstack{52.8\\ {\tiny$\pm$6.9}} & \shortstack{57.6\\ {\tiny$\pm$2.0}} & \shortstack{54.4\\ {\tiny$\pm$1.6}} & \shortstack{79.4\\ {\tiny$\pm$0.0}} & \shortstack{50.8\\ {\tiny$\pm$7.3}} & \shortstack{66.5\\ {\tiny$\pm$4.8}} \\
\bottomrule
\end{tabular}%
}
\end{table*}

\subsection{Feature-Group Importance}
\label{ablation:importance}
Whereas the ablation retrains on subsets, we complement it with
\emph{permutation importance}, which quantifies how much the \emph{trained} full
detector relies on each group: we shuffle a group's features on the test set and
measure the resulting drop in AUC-ROC (larger drop $=$ more informative). This is
computed over $20$ permutation repeats per cell.
Fig.~\ref{fig:importance} shows the per-model breakdown across the four different experiments. The importance analysis mirrors the ablation. In the baseline and cross-domain experiments, \textit{the detector is overwhelmingly based on hidden-state
features}. In the cross-lingual and cross-lingual-domain
regimes, all three importances are small and comparable, reflecting that transfer is weak and no single group carries a
strong and transferable signal.

\subsection{Statistical Significance}
\label{ablation:significance}
Because each configuration is trained over $6$ random seeds, we test whether the
differences between configurations are statistically reliable rather than seed
noise. For every (model, setting) cell we compare the full model against each
feature-group configuration with a paired $t$-test over the seed-matched runs,
and corroborate it with the Wilcoxon signed-rank test. 
Table~\ref{tab:app_significance} summarizes the AUC-ROC results per regime,
reporting the mean difference and the number of cells reaching $p<0.05$.

\begin{table}[t]\centering
\caption{Statistical significance of the feature-group ablation (AUC-ROC, paired $t$-test over 6 seeds). Each entry gives the mean difference $\Delta$ (percentage points) and the number of cells with $p<0.05$ out of the total in that regime. Positive $\Delta$ means the full model is higher. Wilcoxon signed-rank agrees in all cases up to its $n{=}6$ resolution.}
\label{tab:app_significance}
\resizebox{\linewidth}{!}{%
\begin{tabular}{l cccc}\toprule
\textbf{Comparison} & \textbf{BS} & \textbf{CL} & \textbf{CD} & \textbf{CL-CD} \\\midrule
All vs.\ Hidden & $+3.2$\,(15/18) & $+4.9$\,(8/12) & $+4.1$\,(8/12) & $+5.9$\,(7/12) \\
All vs.\ Attention & $+9.5$\,(12/18) & $-0.6$\,(6/12) & $+10.2$\,(9/12) & $+3.8$\,(8/12) \\
All vs.\ Probability & $+9.5$\,(14/18) & $+2.7$\,(11/12) & $+8.2$\,(10/12) & $+7.7$\,(9/12) \\
All vs.\ Hid+Att & $+3.4$\,(10/18) & $+1.2$\,(5/12) & $+3.6$\,(7/12) & $+2.1$\,(3/12) \\
All vs.\ Hid+Prob & $+0.7$\,(8/18) & $+2.4$\,(4/12) & $+1.2$\,(5/12) & $+1.4$\,(2/12) \\
All vs.\ Att+Prob & $+3.7$\,(10/18) & $-3.2$\,(7/12) & $+2.7$\,(5/12) & $+2.4$\,(7/12) \\

\bottomrule\end{tabular}%
}
\end{table}

Table~\ref{tab:app_significance} shows that the full feature model consistently outperforms individual feature groups, with positive mean AUC-ROC differences across nearly all regimes. The gains are particularly pronounced for probability features, reaching $+9.5$, $+8.2$, and $+7.7$ points in the baseline, cross-domain, and cross-lingual-domain settings, respectively, and are significant for most model--setting cells. Hidden states provide a stronger standalone baseline, yet the full model still improves AUC-ROC by $3.2$--$5.9$ points across all regimes. Attention is notably more competitive under cross-lingual transfer, where it slightly exceeds the full model on average ($\Delta=-0.6$), reinforcing that feature utility changes with the type of distribution shift. In contrast, the gains over pairwise combinations are smaller and less consistently significant, suggesting that much of the complementary information can be captured by two feature families. Overall, the results support feature combination as a robust choice and show that no configuration universally dominates all models and transfer directions.

\section{Qualitative Analysis}
\label{insights}
 
\subsection{Layer-wise hallucination signal propagation}
\begin{figure*}[t]
    \centering
    \includegraphics[width=\linewidth]{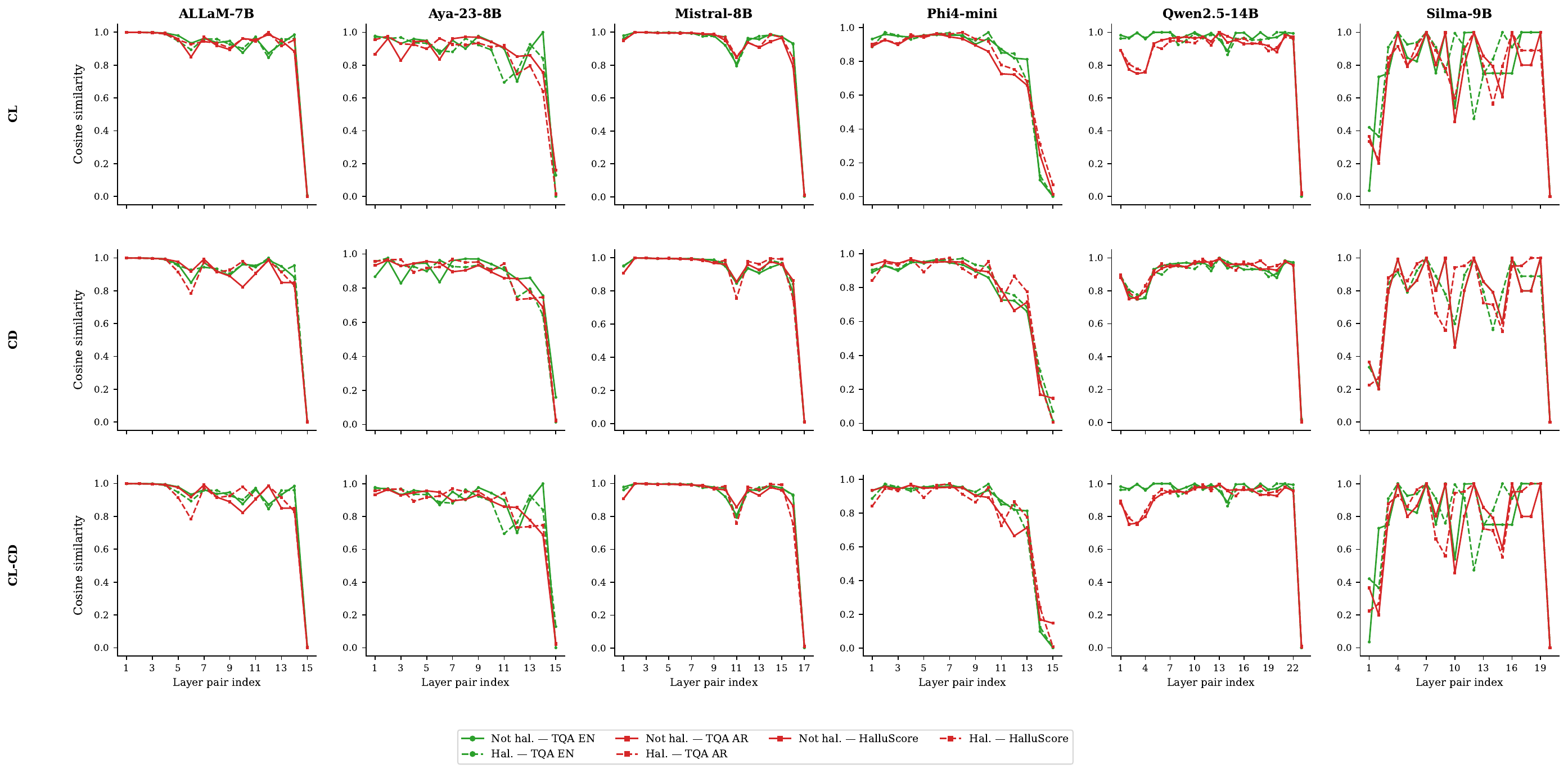}
    \caption{Layer-wise cosine similarity trajectories of hidden states 
across layer pairs for representative hallucinated (dashed) and 
non-hallucinated (solid) examples, shown for all six models under three 
transfer settings: cross-lingual (TruthfulQA EN vs TruthfulQA AR, Row 1), 
cross-domain (TruthfulQA AR vs HalluScore, Row 2), and combined 
cross-lingual cross-domain (TruthfulQA EN vs HalluScore, Row 3).}
    \label{fig:cos}
\end{figure*}

Figure \ref{fig:cos} shows the layer-wise cosine similarity of hidden state representations across three settings: cross-lingual, cross-domain, and combined cross-lingual cross-domain transfer. Cosine similarity between consecutive layers measures how smoothly internal representations evolve as information propagates through the transformer. High similarity indicates gradual and consistent representational refinement, while lower similarity reflects abrupt representational shifts, suggesting that the model is performing a qualitatively different computation at that transition. In the context of hallucination detection, sharp drops in cosine similarity in the final layers are particularly informative, as they may indicate the model transitioning toward a final output representation that deviates from earlier semantic processing. Such divergence is often more pronounced for hallucinated responses, where the model may rely more on generative priors than on grounded internal representations.

As illustrated in Figure \ref{fig:cos}, a consistent pattern emerges across all settings: cosine similarity remains high and stable (typically between 0.85 and 1.0) across the early and middle layers, indicating smooth representational refinement as information propagates through the network. This stability is followed by a sharp drop in the final few layer pairs, often approaching near-zero values at the last layer, reflecting the transition from intermediate semantic processing to final output formation. This late-layer collapse appears for both hallucinated and non-hallucinated examples, suggesting it is a general property of transformer generation. However, small deviations between the two trajectories are observable in some models, particularly in the upper layers, which may provide useful discriminative signals for hallucination detection.

\paragraph{Cross-lingual.}
As shown in Figure \ref{fig:cos}, in the cross-lingual setting, the green (TruthfulQA EN) and red 
(TruthfulQA AR) lines of the same style overlap closely throughout all 
layer pairs for \textit{Ministral} and \textit{Phi4-mini}. This confirms that these models 
encode both languages through the same internal computational pathway. 
The final-layer collapse occurs at the same layer pair and with similar 
magnitude in both languages, explaining why cross-lingual transfer 
succeeds for these models. \textit{Aya} shows more divergence between the two languages in 
middle layers (pairs 6--10), with Arabic trajectories showing greater 
oscillation than English. Yet, both converge to the same final-layer 
collapse. \textit{Silma} exhibits the greatest cross-lingual divergence, 
with English and Arabic following noticeably different oscillatory 
patterns throughout the middle layers, which explains its inconsistent 
cross-lingual transfer performance. 

\paragraph{Cross-domain.}
As shown in Figure \ref{fig:cos}, the cross-domain setting (Row 2) exhibits highly similar cosine-similarity trajectories for TruthfulQA AR and HalluScore across most models, indicating that the overall layer-wise evolution of hidden states remains relatively stable when the dataset changes but the language is held constant. This consistency is generally stronger than in the cross-lingual setting, suggesting that domain shift within Arabic perturbs internal dynamics less than language shift. The clearest examples are \textit{Ministral} and \textit{Qwen2.5}, whose curves remain nearly aligned across the two Arabic datasets, while \textit{Phi4-mini} also preserves a similar trajectory shape despite stronger late-layer deviations. In contrast, \textit{Silma} shows substantial oscillation across layers, indicating less stable internal dynamics, and \textit{ALLaM} exhibits a recurring dip around the middle layers across both datasets, suggesting that this pattern reflects model-specific processing behavior on Arabic text rather than a dataset-specific effect. Overall, the stronger trajectory consistency in Row 2 provides a plausible explanation for why cross-domain transfer is generally more successful than cross-lingual transfer.

\paragraph{Combined cross-lingual cross-domain.}
As shown in Figure \ref{fig:cos}, the combined cross-lingual cross-domain setting (Row 3) shows largely similar cosine-similarity trajectories to those observed in the cross-lingual setting. This suggests that adding domain shift does not substantially disrupt the overall layer-wise dynamics beyond the effect of language shift alone. For most models, including \textit{Ministral}, \textit{Qwen2.5}, and \textit{Phi4-mini}, the trajectories for TruthfulQA EN and HalluScore follow comparable patterns across the middle layers before exhibiting the characteristic late-layer collapse. While some deviations appear in intermediate layers for models such as \textit{Aya} and \textit{Allam}, these differences are moderate. Importantly, the final-layer collapse pattern remains consistent across all models and datasets, suggesting that late-layer representational transitions are a stable property of the generation process rather than being strongly affected by language or domain shifts. Overall, the similarity between the cross-lingual and the combined cross-lingual cross-domain settings suggests that language differences may play a larger role than domain differences in shaping internal representation dynamics.

\begin{figure*}[t]
    \centering
    \includegraphics[width=\linewidth]{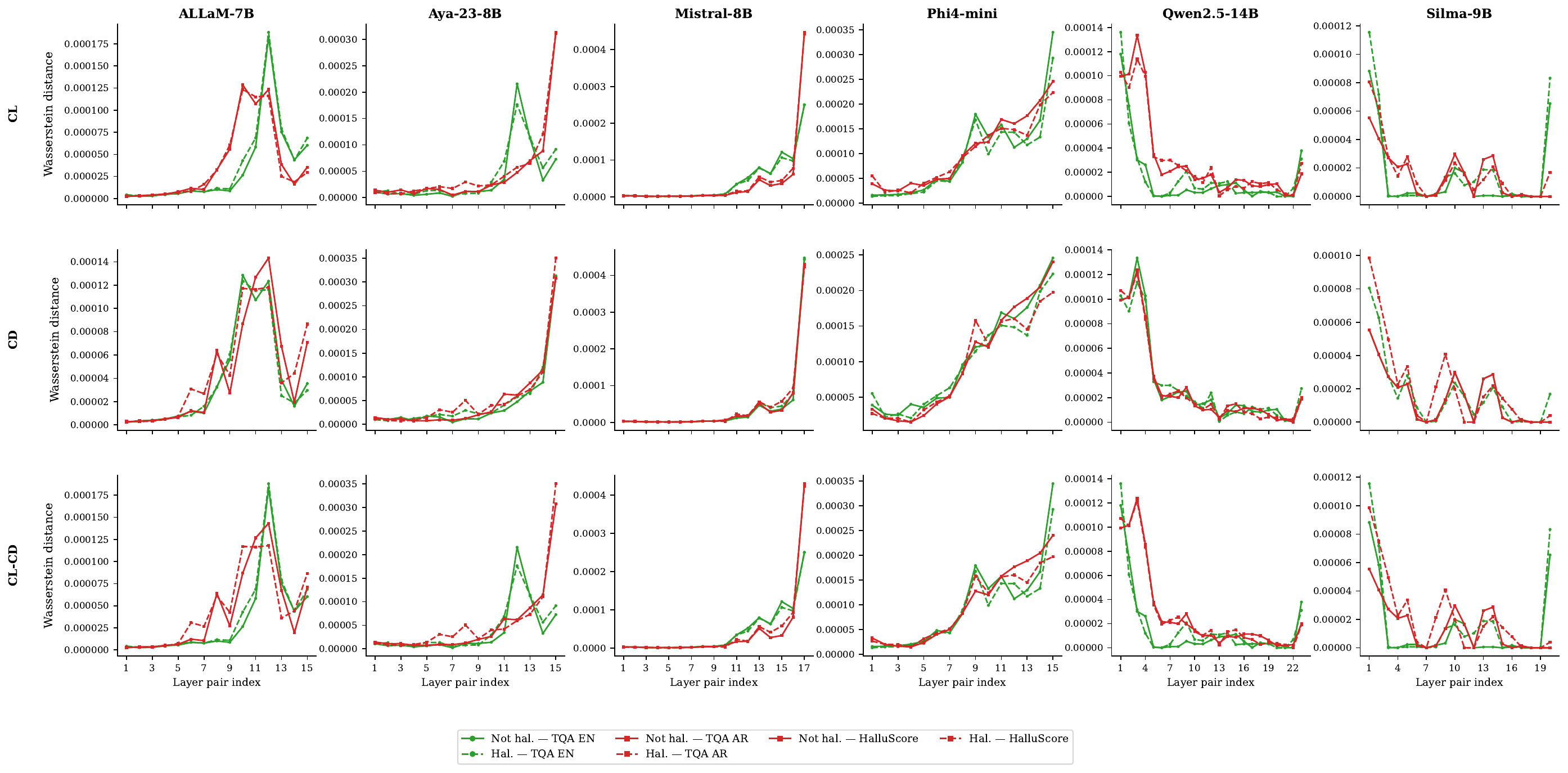}
    \caption{Layer-wise Wasserstein distance trajectories of hidden states 
across layer pairs for representative hallucinated (dashed) and 
non-hallucinated (solid) examples, shown for all six models under the 
same three transfer settings as Figure~\ref{fig:cos}.}
    \label{fig:wass}
\end{figure*}

\subsection{Comparative Analysis of Cosine and Wasserstein Layer Dynamics}
Although Wasserstein distance captures the magnitude of distributional shifts between consecutive hidden-state layers, our analysis suggests that it is less discriminative as a standalone hallucination signal. As shown in Figure~\ref{fig:wass}, the trajectories of hallucinated and non-hallucinated examples within the same dataset generally follow similar trends across most models and transfer settings, with comparatively smaller deviations between the two trajectories than those observed in the cosine similarity plots. This substantial overlap indicates limited separability between the two classes when relying solely on Wasserstein-based signals.

A second observation is that Wasserstein trajectories appear strongly influenced by architecture-specific representational dynamics rather than hallucination-specific behavior. For example, \textit{Allam} consistently exhibits a mid-layer peak around layer pairs 10--12, \textit{Ministral} shows a gradual increase toward later layers, and \textit{Qwen2.5} and \textit{Silma} display prominent early-layer spikes. Importantly, these patterns remain largely consistent across hallucinated and non-hallucinated examples as well as across transfer settings, suggesting that Wasserstein distance primarily reflects how different architectures evolve internal representations across layers.

In contrast, cosine similarity (Figure~\ref{fig:cos}) exhibits more consistent deviations between hallucinated and non-hallucinated trajectories, particularly in the upper layers for several models. These late-layer divergences appear more pronounced than those observed in the Wasserstein plots, suggesting that cosine similarity better captures directional changes in representations than distributional magnitude shifts alone. This indicates that while Wasserstein distance provides useful information about the overall evolution of hidden-state distributions, cosine similarity contributes stronger discriminative cues for hallucination detection. Together, these complementary signals enable the classifier to capture both the structural evolution of representations and the directional deviations associated with hallucination-related generation behavior.

\begin{figure*}[!t]
    \centering
    \includegraphics[width=\linewidth]{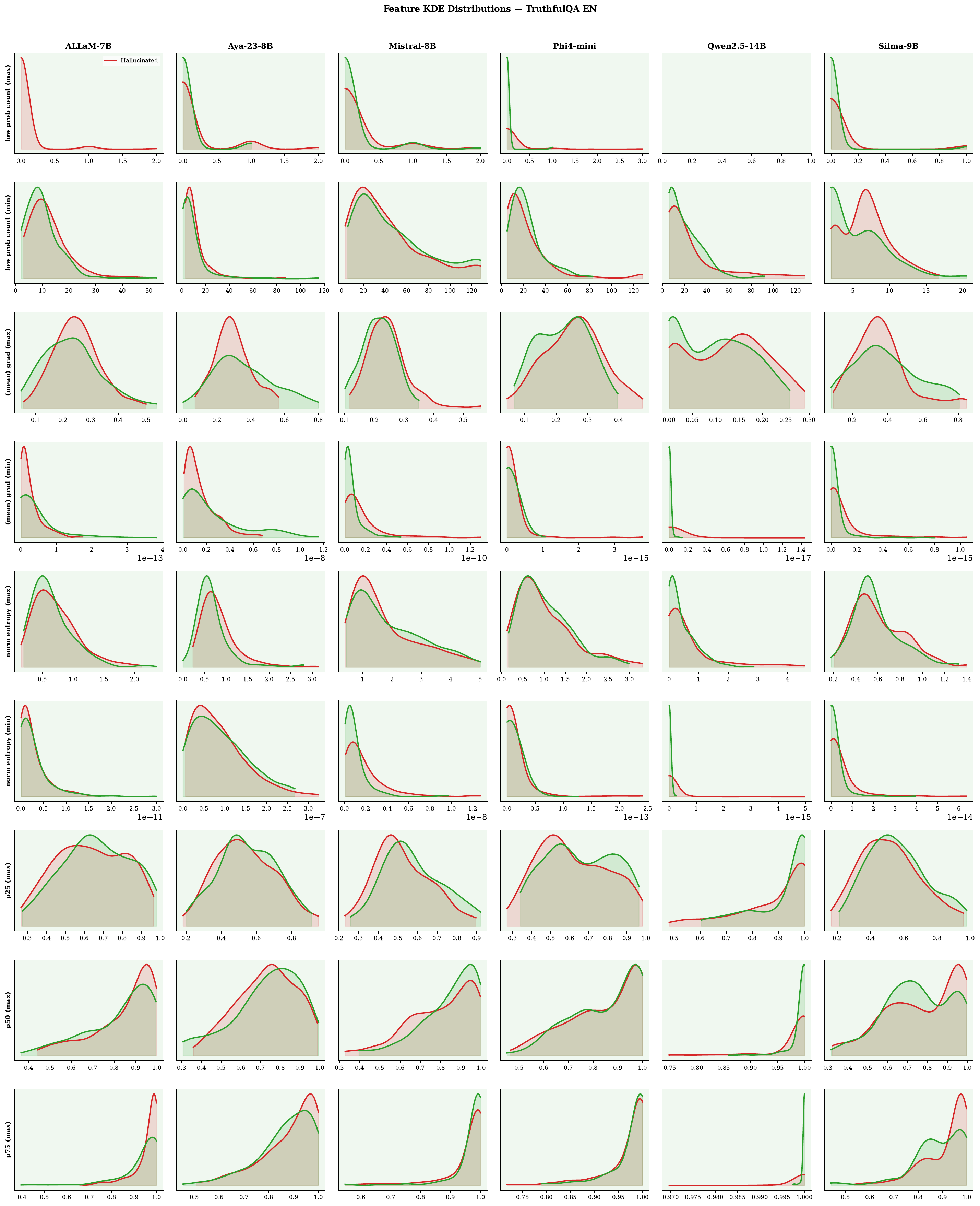}
    \caption{Kernel density estimates of the four most discriminative  features for hallucinated (red) and non-hallucinated (green) 
examples on TruthfulQA English across all six models.}
    \label{fig:kdEn}
\end{figure*}

\begin{figure*}[!t]
    \centering
    \includegraphics[width=\linewidth]{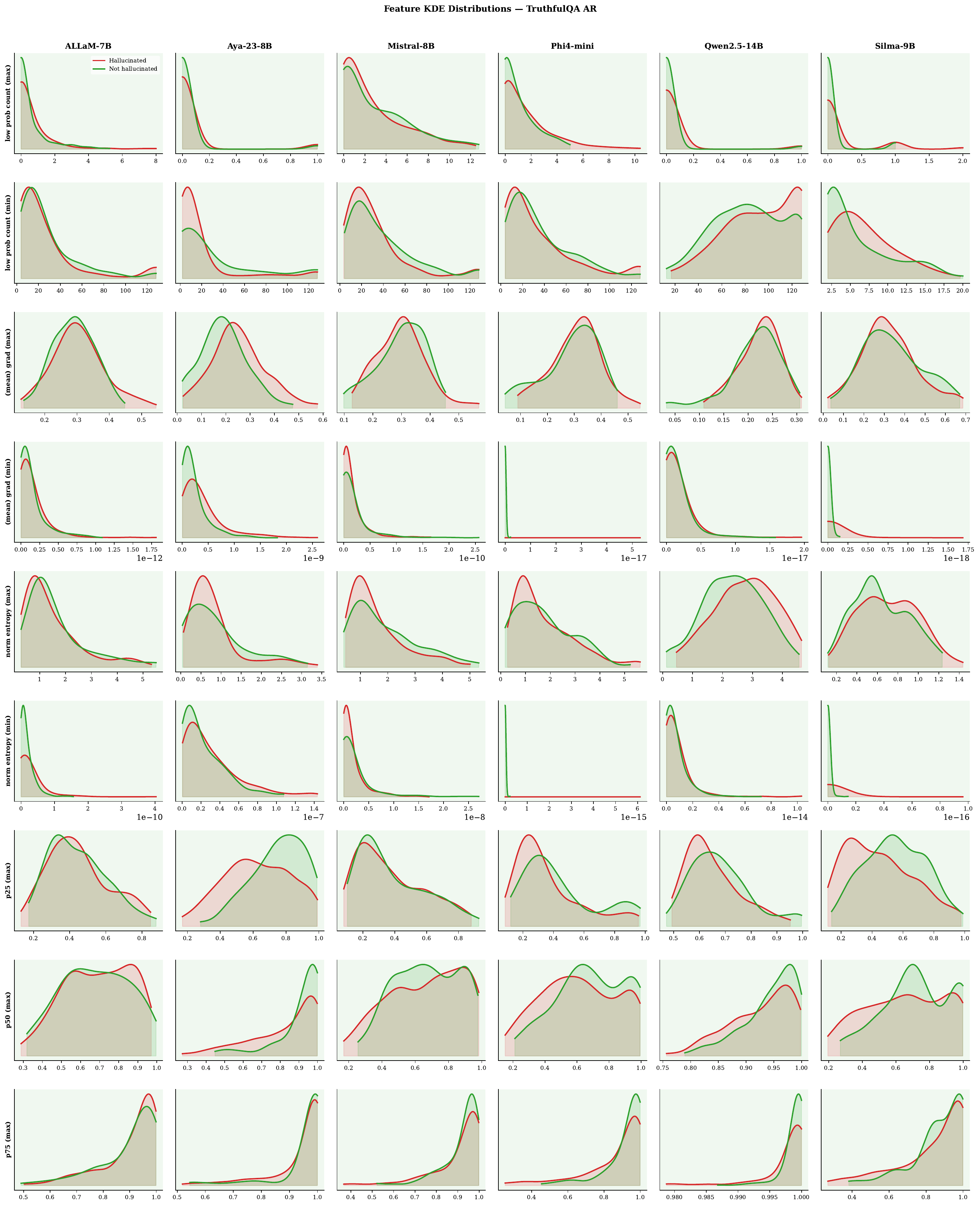}
    \caption{Kernel density estimates of the four most discriminative  features for hallucinated (red) and non-hallucinated (green) 
examples on TruthfulQA Arabic across all six models.}
    \label{fig:kdAr}
\end{figure*}

\begin{figure*}[!t]
    \centering
    \includegraphics[width=\linewidth]{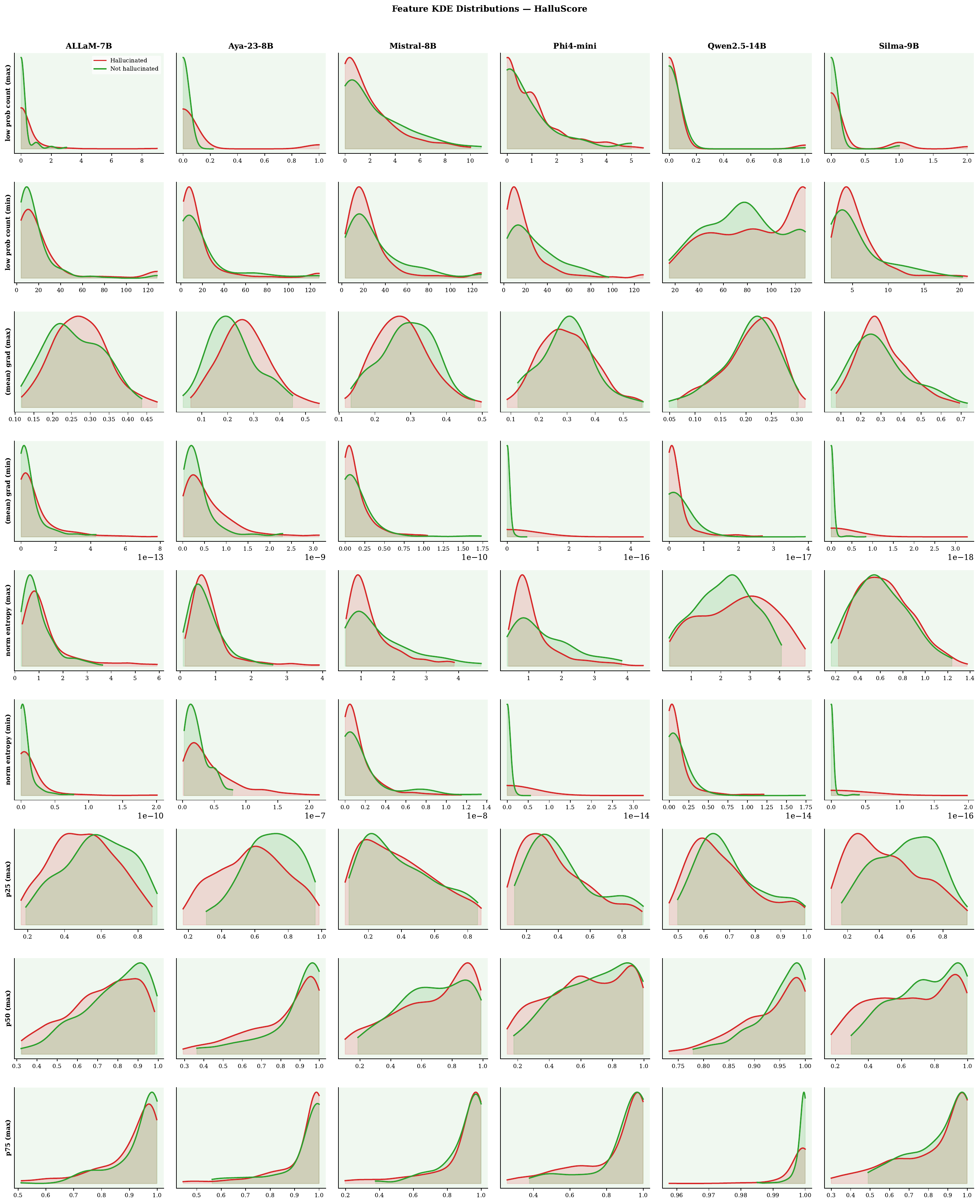}
    \caption{Kernel density estimates of the four most discriminative  features for hallucinated (red) and non-hallucinated (green) 
examples on HalluScore across all six models.}
    \label{fig:kdeHS}
\end{figure*}

\begin{figure*}[t!]
    \centering
    \includegraphics[width=0.8\linewidth]{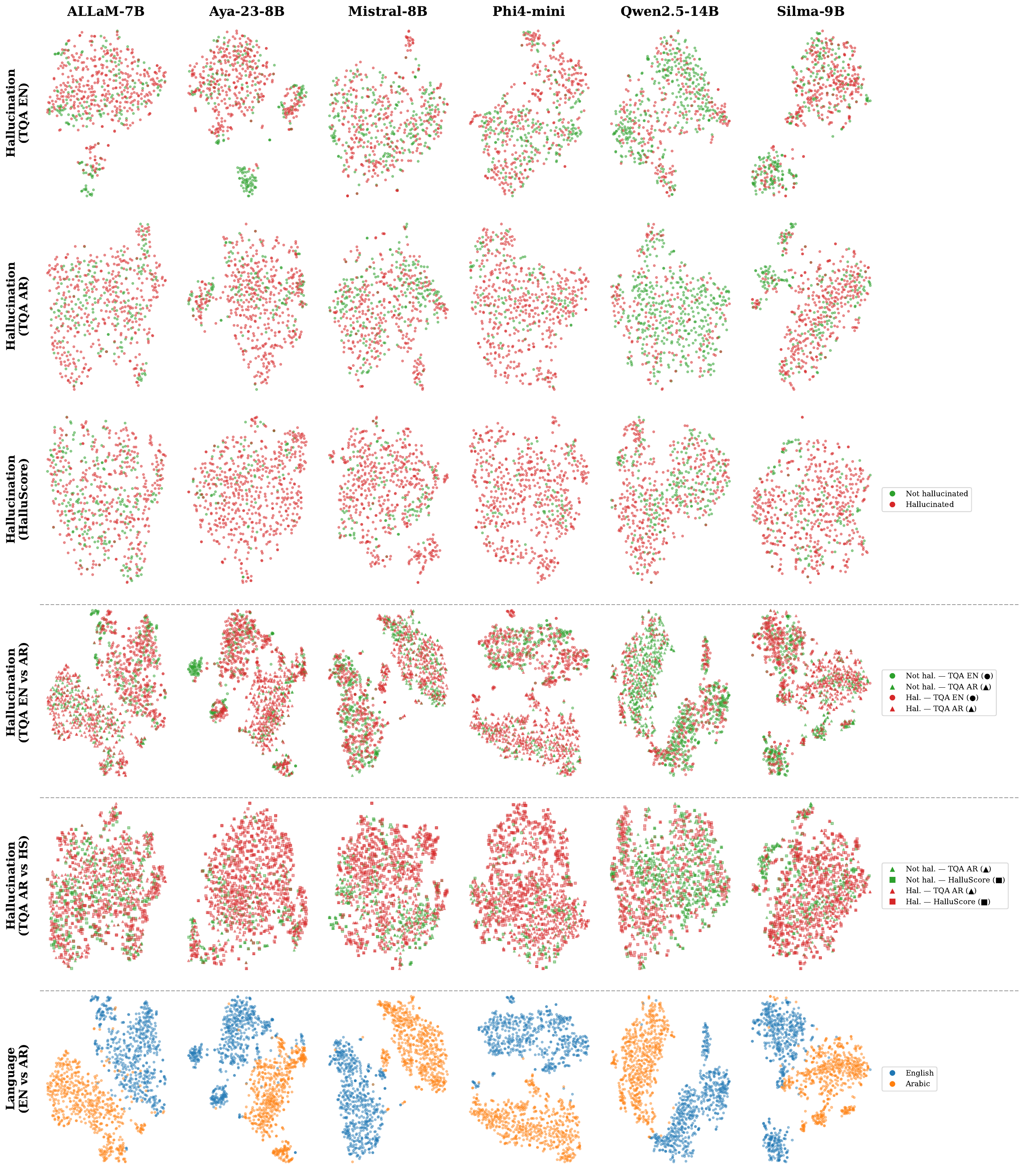}
\caption{t-SNE visualization of \textit{CrossHallu} results across six LLMs under six analysis views. \textbf{Rows 1--3:} dataset-specific distributions for TruthfulQA English (TQA EN), TruthfulQA Arabic (TQA AR), and HalluScore (HS), coloured by hallucination label. \textbf{Row 4:} a shared English-Arabic embedding to examine cross-lingual separation of hallucinated and non-hallucinated responses. \textbf{Row 5:} a shared Arabic cross-domain embedding (TQA AR + HS) to assess domain-level consistency of hallucination signals. \textbf{Row 6:} the English-Arabic shared embedding colored by language to illustrate feature space alignment across languages.}
    \label{fig:tsne_all}
\end{figure*}

\subsection{Feature Discriminability Analysis}

To better understand which internal signals contribute most to hallucination detection, we analyze each feature's discriminative ability using kernel density estimates (KDEs) across the three evaluated datasets: TruthfulQA EN, TruthfulQA AR, and HalluScore, as shown in Figures~\ref{fig:kdEn}, \ref{fig:kdAr}, and \ref{fig:kdeHS}. Rather than analyzing transfer behavior, this section focuses on how well each feature separates hallucinated and non-hallucinated examples within each dataset.

\paragraph{Low probability count.}
The low-probability count features (minimum and maximum) measure how often the model assigns very low probabilities to generated tokens, reflecting decoding uncertainty. The maximum-based variant measures how often the LLM exhibits low confidence in its selected token, which reflects decoding uncertainty during generation. In contrast, the minimum-based variant measures how often the model assigns extremely small probabilities to the least likely tokens in the vocabulary. It shows the sharpness and concentration of the output probability distribution and how strongly unlikely alternatives are suppressed. Higher values of the maximum-based feature indicate increased generation uncertainty, whereas higher values of the minimum-based feature indicate a more concentrated and peaked probability distribution.

The KDE distributions reveal that the probability-based features exhibit substantially different behaviors across models and datasets. Across datasets, this feature appears slightly more informative in HalluScore and TruthfulQ AR than in TruthfulQA EN, suggesting that factual hallucinations may induce more uncertainty spikes in Arabic QA settings. Since some hallucinations can be produced with high confidence, this feature alone is insufficient and must be combined with representation-based features such as cosine similarity. 

For models such as \textit{Phi4-mini}, \textit{Allam}, and \textit{Silma}, the maximum-based variant shows that hallucinated responses frequently exhibit broader distributions and heavier tails compared to non-hallucinated responses. This indicates that hallucinations in these models are often associated with increased decoding uncertainty and lower confidence in the selected tokens. On the other hand, this feature shows moderate separation for \textit{Aya} and \textit{Ministral}, with hallucinated examples tending to exhibit slightly higher counts. The pattern suggests that for these models, confident but incorrect generation (hallucination with high token confidence) is more common than uncertain generation. 

The minimum-based variant is less informative for models such as \textit {Allam} and \textit{Aya} due to substantial overlap between hallucinated and non-hallucinated distributions. In contrast, it becomes highly discriminative for \textit{Qwen2.5}, particularly in TruthfulQA AR and HalluScore, where the hallucinated and non-hallucinated distributions exhibit markedly different shapes. Interestingly, for \textit{Qwen2.5} on TruthfulQA EN, the maximum-based variant collapses to a constant value across all samples, making it entirely non-informative, while the minimum-based variant still captures meaningful distributional differences. This suggests that \textit{Qwen2.5} hallucinations are not associated with a reduced confidence in decoding, but rather with changes in the behavior of probability concentration.

\paragraph{Gradient-based features.}
Gradient-based features reveal distinct hallucination behaviors across models and datasets, with the maximum-gradient feature generally providing stronger and more consistent separability than the minimum-gradient feature. The maximum-gradient feature captures abrupt changes in the LLM’s token confidence trajectory during generation. In the context of hallucination, higher gradients often suggest unstable reasoning dynamics, in which the model may revise its internal representation as it attempts to construct an answer without strong factual grounding.

Across TruthfulQA EN, the maximum-gradient feature exhibits varying levels of discriminative behavior across architectures. \textit{Aya}, \textit{Allam}, and \textit{Silma} show noticeable differences between hallucinated and non-hallucinated distributions, including shifted peaks and broader density regions. This suggests that hallucinations are associated with altered confidence-transition dynamics during decoding. \textit{Qwen2.5} exhibits a distinct bimodal behavior, where hallucinated and non-hallucinated responses follow visibly different density structures rather than simple mean shifts. 

In TruthfulQA AR, the maximum-gradient feature exhibits moderate and highly model-dependent separability between hallucinated and non-hallucinated responses. Most models still show substantial overlap between the two distributions, indicating that confidence-transition dynamics alone are insufficient for strong discrimination. \textit{Aya} displays the clearest separation, where hallucinated responses exhibit shifted or structurally different density patterns compared to non-hallucinated responses. Other models show moderate-to-high overlap distributions with limited discriminative separation. Overall, the maximum-gradient feature in TruthfulQA AR exhibits weaker, less consistent separability than initially expected, with its usefulness depending strongly on the underlying architecture. In contrast, since HalluScore contains more adversarial and culturally grounded hallucinations, it shows larger distributional shifts among nearly all models, which suggest that such hallucinations induce stronger fluctuations in token-level confidence dynamics.

The minimum-gradient feature is substantially less informative across nearly all models and datasets. The distributions are heavily concentrated near zero, often collapsing into sharp spikes with limited variance and substantial overlap between hallucinated and non-hallucinated samples. This suggests that the minimum token probabilities remain relatively stable across decoding steps regardless of hallucination status. Overall, these findings suggest that hallucinations are more strongly associated with fluctuations in the LLM's confidence for highly probable tokens than with changes in the least likely regions of the vocabulary distribution.

\paragraph{Entropy features.}
Normalized entropy (minimum and maximum) measures the uncertainty in the output. These entropy-based features are more informative and stable than the probability-count and gradient features. For the maximum-entropy feature in the TruthfulQA EN dataset, most LLMs exhibit noticeable distributional differences between hallucinated and non-hallucinated responses, suggesting that hallucinations are associated with changes in uncertainty and dispersion of the LLM’s high-confidence token predictions. Hallucinated responses exhibit shifted peaks and broader distributions compared to non-hallucinated responses. \textit{Qwen2.5} exhibits a distinct behavior where non-hallucinated responses are strongly concentrated near very low entropy values, while hallucinated responses display a broader tail, suggesting that hallucinations in \textit{Qwen2.5} are associated with increased dispersion in the token probability distribution rather than simple decoding instability.

In contrast, using the TruthfulQA AR, the substantial overlap between hallucinated and non-hallucinated entropy distributions in most LLMs suggests that elevated uncertainty is not necessarily indicative of hallucination in Arabic generation. Instead, uncertainty may also reflect linguistic variability and morphological complexity, which limits the effectiveness of uncertainty-only hallucination indicators. In HalluScore, the entropy-based features become substantially more informative for \textit{Allam}, \textit{Aya}, and \textit{Phi4-mini}, where hallucinated responses consistently shift toward broader and higher-entropy regions compared to non-hallucinated responses. These models exhibit clearer density separation than in TruthfulQA EN and TruthfulQA AR, suggesting that adversarial and culturally grounded hallucinations induce stronger uncertainty and probability-dispersion effects. 

Although the minimum-entropy feature is generally less discriminative than the maximum-entropy feature, several LLMs exhibit a recurring pattern where hallucinated responses extend into the high-value tail of the distribution. This behavior is particularly visible for \textit{Phi4-mini} and \textit{Silma} across all three datasets, for \textit{Allam} in TruthfulQA AR and HalluScore, for \textit{Aya} in HalluScore, and for \textit{Qwen2.5} in TruthfulQA EN. The presence of hallucinated responses in these high-entropy tail regions suggests that hallucinated generation is occasionally associated with increased dispersion in the low-probability region of the vocabulary distribution, resulting in weaker concentration of unlikely token alternatives. In other words, during hallucinated decoding, the LLM does not suppress improbable tokens as sharply as it does for non-hallucinated responses, producing broader minimum-entropy tails. This effect becomes more pronounced in HalluScore, indicating that adversarial and culturally grounded hallucinations induce stronger instability in the low-probability vocabulary tail.

\paragraph{Probability percentile features.}
The percentile-based probability features characterize different confidence regions of the token probability trajectory during generation. Lower-percentile features, such as $p25$, capture the lower-confidence portion of the decoding process and reflect how often the model assigns relatively weak confidence to generated tokens. Smaller percentile values indicate lower confidence and greater uncertainty in token selection, whereas larger values indicate more stable and confident decoding behavior. The median percentile feature ($p50$) captures the central tendency of token confidence across generations and reflects the overall stability of the decoding trajectory. In contrast, the upper-percentile feature ($p75$) represents the highest-confidence decoding states, where values close to $1.0$ indicate that the model consistently assigns very high probabilities to the selected tokens.

The extent of the hallucination-related separability observed in the percentile features varies across datasets. TruthfulQA EN exhibits moderate separability, where hallucinations alter the lower- and median-confidence regions but still retain substantial overlap between hallucinated and non-hallucinated responses. TruthfulQA AR generally shows stronger shifts in $p25$ and $p50$, indicating that hallucinations in Arabic generation induce larger disruptions in medium- and lower-confidence decoding behavior. HalluScore demonstrates the strongest overall separability, where hallucinated responses consistently occupy broader and lower-confidence regions. 

\texttt{p75\_max} provides the weakest discriminative power across all LLMs and datasets, with both distributions concentrated near 1.0 and largely overlapping. \textit{Qwen2.5} shows the most extreme compression, with its \texttt{p75\_max} distribution confined to 0.970--1.000 in TruthfulQA EN and 0.980--1.000 in Arabic datasets. This confirms that it generates with uniformly high top-quartile token confidence regardless of factual correctness.

\subsection{Model-wise Feature Observations}

The KDE distributions reveal that different models rely on different probability-based signals for hallucination discrimination, with varying degrees of separability across datasets and architectures. Overall, percentile-based features, particularly \texttt{p25\_max} and \texttt{p50\_max}, provide the most consistent separation across models, whereas entropy- and gradient-based features exhibit more architecture- and dataset-dependent behavior. \textit{Allam} generally exhibits mild-to-moderate separation across the analyzed features. The separation is better in HalluScore than in the other datasets. However, Table~\ref{tab:results} shows that \textit{Allam} achieves a higher monolingual AUC-ROC on TruthfulQA EN than on HalluScore, suggesting that KDE-based visual separation does not necessarily correspond directly to threshold-independent classification performance.

\textit{Aya} exhibits some of the most consistent hallucination-related shifts across all three datasets. Features such as \texttt{p25\_max} and \texttt{mean\_grad\_max} repeatedly show visible differences between hallucinated and non-hallucinated responses, with the separation becoming more pronounced in TruthfulQA AR and HalluScore. \textit{Ministral} exhibits relatively stable and moderate separability across datasets. The percentile-based features show consistent but not strongly pronounced shifts, while entropy- and gradient-based features generally exhibit substantial overlap with mild tail differences. Compared to other models, \textit{Ministral} displays fewer abrupt distributional changes across datasets, suggesting comparatively stable hallucination-related confidence dynamics. \textit{Phi4-mini} exhibits dataset-dependent behavior. On TruthfulQA EN, most features show moderate overlap, although percentile-based features and \texttt{mean\_grad\_max} still display visible shifts between hallucinated and non-hallucinated responses.

\textit{Qwen2.5} exhibits the most architecture-specific behavior among the analyzed models. Certain features, particularly \texttt{low\_prob\_count\_max}, collapse to near-constant distributions in some settings, especially on TruthfulQA EN, indicating that hallucinated and non-hallucinated responses often maintain similarly high confidence levels. Across all datasets, \textit{Qwen2.5} consistently exhibits highly concentrated \texttt{p75\_max} distributions near $1.0$, suggesting that hallucinated responses frequently retain highly confident token predictions even during hallucinated generation. Finally, \textit{Silma} exhibits broad and highly variable KDE distributions across most features and datasets. Percentile-based features, particularly \texttt{p25\_max} and \texttt{p50\_max}, show moderate but visible separation, with hallucinated responses frequently occupying broader and lower-confidence regions than non-hallucinated responses. Entropy- and gradient-based features also exhibit noticeable variability. Compared to the other models, \textit{Silma} consistently exhibits greater spread and more oscillatory distributional behavior, which aligns with the oscillatory hidden-state dynamics observed in the layer-wise representation analysis.

\subsection{Cross-dataset Consistency}
The KDE analysis reveals that the feature distributions are structurally most similar between TruthfulQA AR and HalluScore, confirming that both Arabic datasets share similar internal token probability statistics. The results reflect the morphological complexity of Arabic, generating longer token sequences with a more variable probability profile. The \texttt{p25\_max} separation is slightly smaller on HalluScore than on TruthfulQA AR for most models, providing feature-level evidence for the asymmetric cross-domain transfer observed in Table~\ref{tab:results}.

\end{document}